\documentclass{article}

\usepackage{microtype}
\usepackage{graphicx}
\usepackage{subfigure}
\usepackage[subfigure]{tocloft}
\usepackage{booktabs} 

\usepackage{hyperref}  
\usepackage{url}
\usepackage{booktabs}       
\usepackage{amsfonts}       
\usepackage{nicefrac}       
\usepackage{microtype}      
\usepackage[dvipsnames]{xcolor}    
\usepackage{xcolor}         
\usepackage{paralist}
\usepackage{amsmath}
\usepackage{booktabs}
\usepackage{multirow,makecell}
\usepackage{color, colortbl}
\definecolor{Highlight}{rgb}{0.89,0.89,0.94}
\newcommand{\chl}{\cellcolor{Highlight}}
\usepackage{wrapfig}
\usepackage{tocloft}   
\usepackage{tikz}
\usepackage{printlen}

\hypersetup{
    colorlinks=true,
    linkcolor=RoyalBlue,
    urlcolor=RoyalBlue,
    citecolor=RoyalBlue
}

\usepackage[accepted]{icml2025}

\icmltitlerunning{Diffusion on Language Model Encodings for Protein Sequence Generation}

\begin{document}
\setcounter{tocdepth}{-5} 

\twocolumn[
\icmltitle{Diffusion on Language Model Encodings for Protein Sequence Generation}

\icmlsetsymbol{equal}{*}

\begin{icmlauthorlist}
\icmlauthor{Viacheslav Meshchaninov}{equal,cons}
\icmlauthor{Pavel Strashnov}{equal,airi}
\icmlauthor{Andrey Shevtsov}{equal,airi}
\icmlauthor{Fedor Nikolaev}{airi} \\
\icmlauthor{Nikita Ivanisenko}{airi}
\icmlauthor{Olga Kardymon}{airi}
\icmlauthor{Dmitry Vetrov}{cons}
\end{icmlauthorlist}

\icmlaffiliation{airi}{AIRI, Moscow, Russia}
\icmlaffiliation{cons}{Constructor University, Bremen, Germany}

\icmlcorrespondingauthor{Viacheslav Meshchaninov}{meshchaninov.viacheslav@gmail.com}
\icmlcorrespondingauthor{Pavel Strashnov}{strashnov@airi.net}
\icmlcorrespondingauthor{Andrey Shevtsov}{shevtsov@airi.net}

\icmlkeywords{Machine Learning, ICML, protein design, diffusion}

\vskip 0.3in
]



\printAffiliationsAndNotice{\icmlEqualContribution} 

\begin{abstract}

Protein sequence design has seen significant advances through discrete diffusion and autoregressive approaches, yet the potential of continuous diffusion remains underexplored. Here, we present DiMA, a latent diffusion framework that operates on protein language model representations. Through systematic exploration of architectural choices and diffusion components, we develop a robust methodology that generalizes across multiple protein encoders ranging from 8M to 3B parameters. We demonstrate that our framework achieves consistently high performance across sequence-only (ESM-2, ESMc), dual-decodable (CHEAP), and multimodal (SaProt) representations using the same architecture and training approach. We extensively evaluate existing methods alongside DiMA using multiple metrics across two protein modalities, covering quality, diversity, novelty, and distribution matching of generated proteins. DiMA consistently produces novel, high-quality and diverse protein sequences and achieves strong results compared to baselines such as autoregressive, discrete diffusion and flow matching language models. The model demonstrates versatile functionality, supporting conditional generation tasks including protein family-generation, motif scaffolding and infilling, and fold-specific sequence design. This work provides a universal continuous diffusion framework for protein sequence generation, offering both architectural insights and practical applicability across various protein design scenarios. Code is released at \href{https://github.com/MeshchaninovViacheslav/DiMA}{GitHub}.

\end{abstract}

\begin{figure*}[t]
    \centering
    \includegraphics[width=0.99\textwidth]{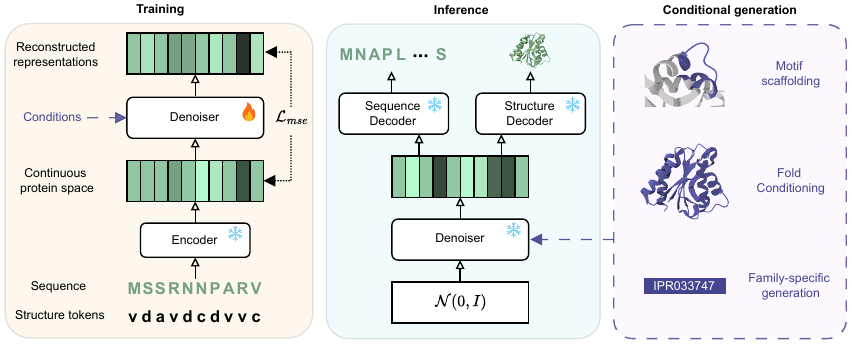}
    \caption{ \small \textbf{DiMA.} The framework consists of three main components: (1) a pre-trained protein language model encoder that maps amino acid sequences to continuous latent representations, (2) a diffusion denoiser that generates latent vectors from Gaussian noise, and (3) sequence and structure decoders that reconstruct amino acid sequences and protein structures from the generated latent representations. During training, the model learns to denoise corrupted protein representations. During inference, the framework supports both unconditional generation and conditional generation tasks including motif scaffolding, fold conditioning, and family-specific generation. The approach enables joint sequence-structure generation while operating entirely in continuous latent space.}
    \label{main}
\end{figure*}

\section{Introduction}
\label{introduction}

Protein generation is emerging as a key area in academic research, with applications spanning bioinformatics, synthetic biology, and protein-based therapeutics \citep{gen_seq, gen_struct}. Recent progress in this field has been driven by three main approaches: autoregressive models \citep{ProGen_Nature, protgpt2, SeqDesign, lv2024prollama}, which have demonstrated effectiveness in capturing sequential dependencies, diffusion models \citep{EvoDiff, wang2024diffusion, DPLM2}, which have shown promise in both sequence and structure generation tasks and flow- based models \cite{foldflow2, campbell2024generative, scaffold_frameflow, lin2024out}, which have achieved impressive results on protein conditional structure generation.

While discrete diffusion models have been successfully adapted for amino acid sequence generation \citep{EvoDiff, wang2024diffusion}, and significant advances have been made in three-dimensional protein diffusion \citep{RFDiffusion, FoldingDiff, AlQuraishi_diffusion, fu2024latent}, continuous diffusion on protein sequence representations remains underexplored. Previous attempts \citep{lee_score-based_2023, zhang2023pro} have been limited to specific protein representations or focused primarily on conditional tasks, leaving the potential of continuous diffusion for general protein generation largely untapped.

Recent advances in protein language models (pLMs) have produced increasingly sophisticated continuous representations of proteins \citep{ESM-2, lu2024tokenized, su2023saprot}. These representations capture both sequence and structural information, providing a natural foundation for latent diffusion approaches. Notably, studies have demonstrated that reconstructing proteins from continuous representations yields greater accuracy than from discrete ones \citep{lu2024tokenized, gaujac2024learning}, suggesting potential advantages for continuous diffusion models in capturing protein properties.

In this study, we develop DiMA, a new latent diffusion model that operates on pLM representations. We demonstrate that continuous diffusion on protein embeddings enables effective sequence and structure generation across multiple tasks and encoder architectures. DiMA addresses the limitations of previous continuous diffusion approaches that have been limited to specific representations by establishing a unified framework that generalizes across diverse protein encoders. By operating in the continuous latent space of these pre-trained encoders, our approach circumvents the challenges associated with discrete sequence modeling while maintaining the expressiveness needed for complex protein design tasks. The framework is designed to be encoder-agnostic, allowing it to benefit from advances in protein representation learning without requiring architectural modifications. Our key contributions are as follows:
\begin{compactitem}
   \item We develop DiMA, a continuous latent diffusion model that operates effectively across different pLM representations, ranging from sequence-only ESM-2  to dual-decodable CHEAP and structure-aware SaProt encoders.
   \item Through systematic exploration of architectural choices and diffusion components, we establish a robust methodology that generalizes across encoders ranging from 8M to 3B parameters.
   \item Despite having only 35M parameters, DiMA demonstrates strong performance on multiple benchmarks, matching or exceeding specialized models in unconditional generation, family-specific design, motif scaffolding and sequence infilling, and fold-conditioned generation.
   \item Our empirical evaluation across protein sequence and structure metrics demonstrates that DiMA maintains high structural quality while generating diverse and novel proteins, as validated on both SwissProt and AFDBv4-90 datasets.
   \item We show that a single architecture optimized for unconditional generation can effectively adapt to conditional tasks and backbone generation through lightweight modifications, suggesting a path toward unified protein sequence generation.
   
\end{compactitem}

\section{Continuous diffusion on LM representations of protein sequences}

DiMA is a latent diffusion model that operates on continuous protein representations. DiMA consists of three components: a pre-trained encoder ($\mathcal{E}$) that provides a meaningful latent space representation, a diffusion model ($\mathcal{F}$) that generates latent vectors from Gaussian noise, and a decoder ($\mathcal{D}$) that maps generated latents back to amino acid sequences.

\vspace{-1em}
\paragraph{Continuous vs. Discrete Diffusion for Proteins.} While discrete diffusion may appear more intuitive for sequence generation, continuous representations offer compelling advantages for protein modeling. Continuous encodings from protein language models capture rich semantic and structural information that has proven effective across diverse protein tasks, from representation learning to structure prediction. Recent studies have shown that continuous representations yield superior reconstruction accuracy compared to discrete alternatives \citep{lu2024tokenized}, suggesting their potential for generative modeling.

Continuous diffusion offers several theoretical and practical advantages over discrete approaches such as direct application of established score-based techniques like classifier and classifier-free guidance without requiring discrete approximations. It also provides seamless integration with multimodal representations that jointly encode sequence and structure (e.g. CHEAP, SaProt). Also, continuous diffusion is more stable and efficient in training compared to discrete ones. However, the sequential and discrete nature of protein sequences presents unique challenges for continuous diffusion that necessitate careful adaptation of diffusion components to effectively capture the complexity of protein space.

\vspace{-1em}
\paragraph{Latent Space Adaptation.}
We utilize a pre-trained transformer-based pLM as an encoder with ESM-2 \citep{ESM-2} being the default choice unless otherwise specified. The encoder maps the sequence of discrete amino acids $y = [y_1, ..., y_s]$ of length $s$ to the latent vectors $x = [x_1, ..., x_s] \in \mathbb{R}^{s \times d}$, $x = \mathcal{E}(y)$. We apply normalization $z_0 = Normalize(x)$ over the hidden dimension $d$: for each component $d_i$, we precompute the mean and variance over the training data and then apply normalization using these statistics to achieve zero mean and unit variance. This transformation allows us to adapt the discrete protein input to a standard Gaussian diffusion framework.

\vspace{-1em}
\paragraph{Noise Schedule Optimization.}
We have found that the linear and cosine noise schedulers widely employed in the image domain \citep{song2020score, ho2020denoising, nichol2021improved} are sub-optimal for the protein domain. We conjecture that this happens due to the sequential and discrete nature of the protein representations.

\begin{figure}
    \centering
    \includegraphics[width=0.48\textwidth]{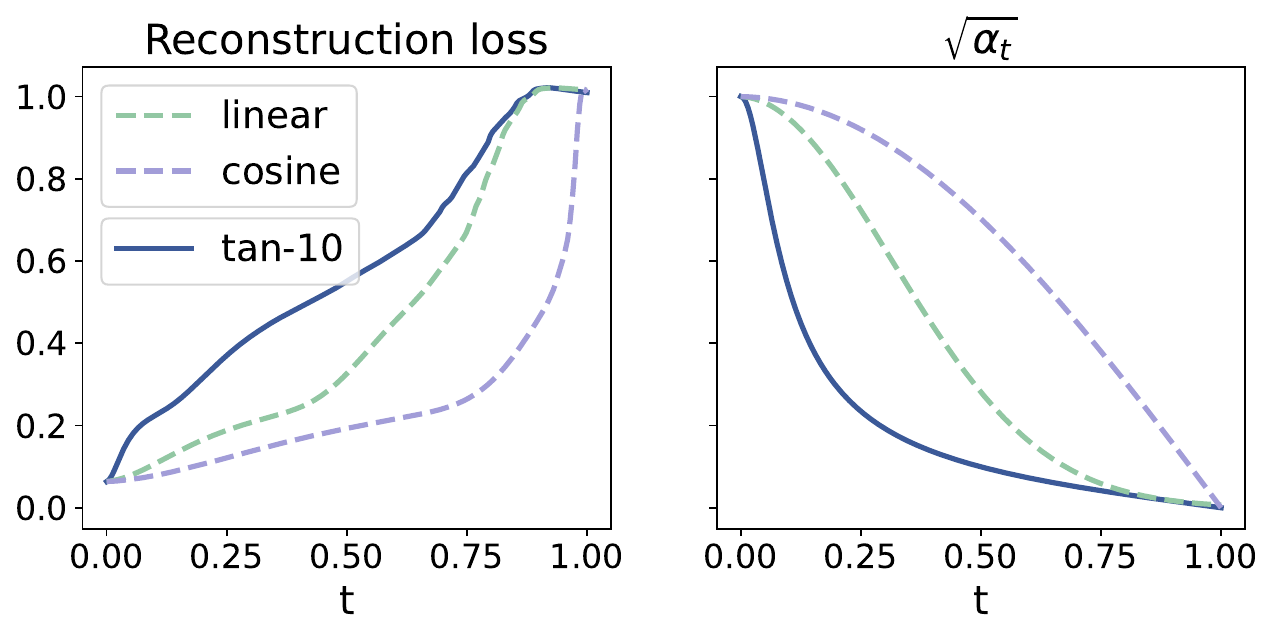}
    \vspace{-5mm}
    \caption{ \small Left: the diffusion reconstruction loss of $z_0$ from $z_t$ with different noise schedules: $||z_0~-~\hat{z}_{\theta}(z_t, t)||^2$. Right: $\sqrt{\alpha_t} \eqref{alpha}$.}
    \vspace{-5mm}
    \label{fig:schedulers}
\end{figure}

The reconstruction loss of diffusion models trained with such schedulers remains minimal at small noise scales. Consequently, the reconstruction of $z_0$ from $z_t = \sqrt{\alpha_t}z_0 + \sqrt{1 - \alpha_t}\varepsilon$ becomes quite trivial for the model for a long period of time, leading to inefficient training. We adopted the noise schedule from \citep{hoogeboom2023simple}:

\vspace{-1em}
\begin{equation}
\alpha_t = \frac{1}{1 + d^2\tan^2(\frac{\pi t}{2})}
\label{alpha}
\end{equation}
\vspace{-1em}

where $d$ is a hyperparameter reflecting the schedule's rate. The larger the value of $d$, the greater the data corruption rate. In this work we use $d=$~10, hence the schedule is named \textit{tan-10}. We utilize a heuristic approach based on the observation that the reconstruction loss should exhibit an approximately linear increase over diffusion time (Figure~\ref{fig:schedulers}). The rationale behind the tan-10 schedule is discussed in Apeendix~\ref{appendix:schedule}.

\vspace{-1em}
\paragraph{Self-Conditioning.}
Following recent advances in sequence generation, we apply the self-conditioning technique \citep{chen2022analog}. The denoising network predicts $\hat{z}_0$ using the latent variable $z_t$ and timestep $t$ as input. Self-conditioning additionally utilizes predicted $\hat{z}_{0,s}$ from the previous timestep $s$ for estimation $\hat{z}_{0,t} = \hat{z}_{\theta}(z_t, t, \hat{z}_{0,s})$, $t < s$.

During training, we sample timestep $t \sim U[0; 1]$. In half of the cases, we provide no additional input to the model, setting $\hat{z}_{0,t} = \emptyset$, where $\emptyset$ is a zero vector. In the remaining cases, we estimate $\hat{z}_{0,t} = \hat{z}_{\theta}(z_t, t, \emptyset)$. The loss is computed as:

\vspace{-1em}
\begin{equation}
\scalebox{0.95}{$
    \mathcal{L}(\theta) = 
    \mathop{\mathbb{E}}_{\varepsilon \sim \mathcal{N}(0, \mathbf{I}), t \sim U[0; 1]} 
    \left[
    ||z_0 - \hat{z}_{\theta}(z_t, t, \text{SG}[\hat{z}_{0, t}])||^2
    \right]
$}
\end{equation}

where $\text{SG}[\cdot]$ denotes the stop-gradient operation. Unlike \citep{chen2022analog}, we apply a linear transformation to $\hat{z}_{0,t}$ and incorporate it into each transformer block's input. The effect of self-conditioning rate on the quality-diversity trade-off is depicted on Figure \ref{fig:self-cond_rate}

\vspace{-1em}
\paragraph{Token Reconstruction.}

Our architecture utilizes the ESM-2 decoder, which was pre-trained alongside the encoder on masked language modeling objectives. We found that additional fine-tuning of the decoder, specifically for amino acid reconstruction, improves sequence generation accuracy from latent representations $x$ during inference. The decoder maintains a simple architecture with a single linear layer. For CHEAP and SaProt representations, we fine-tune their corresponding pretrained decoders alongside the diffusion model, similar to our approach with ESM-2. We found that low-dimensional embeddings (e.g., ESM-2 8M, $d=$~320) are less robust to small perturbations than higher-dimensional ones (ESM2/SaProt 650M, $d=$~1280, CHEAP, $d=$~1024). Fine-tuning the decoders helps minimize these effects during diffusion generation.

\vspace{-1em}
\paragraph{Length Determination.}

Determining sequence length is a key challenge in the inference phase. While many discrete diffusion models \citep{gong2022diffuseq, yuan2022seqdiffuseq, li2022diffusion, mahabadi2023tess} generate padding tokens alongside semantic tokens, we found this approach suboptimal for protein generation. Instead, we employ two distinct strategies for training and inference.

During training, we use an attention mask to focus the model exclusively on semantic tokens. This masking strategy is crucial as encodings of special tokens (like padding) often contain irrelevant or potentially detrimental information to the diffusion model. By excluding these tokens from reconstruction loss computation, we improve both training stability and generation quality.

During inference, we first sample the target sequence length from the training data distribution to ensure realistic protein lengths. We then sample a random Gaussian vector of the appropriate dimension and apply $T$ steps of iterative refinement to generate $\hat{z}_0$. Finally, we denormalize the latent representation and decode it into an amino acid sequence. Additional details about length distribution modeling are provided in Appendix \ref{appendix:length}.

\vspace{-1em}
\paragraph{Model Architecture.}

Our diffusion model employs a transformer architecture with 12 layers, 16 attention heads, and a hidden size of 320. To adapt this architecture for protein sequence generation, we implement two key modifications (detailed architecture specifications are provided in Appendix \ref{appendix:architecture}).

To incorporate time-dependent information, we integrate time embeddings into the transformer blocks via a linear projection, followed by summation at the input of each block.
We extend this mechanism to enable conditional generation tasks, including protein family-specific generation, motif scaffolding, and fold-specific sequence design.
For scaffolding and fold-specific sequence design, we further enhance conditioning by introducing cross-attention blocks that process condition encodings, allowing the model to leverage structural constraints effectively.

Inspired by~\citep{bao2023all}, we incorporate long skip connections throughout the transformer architecture.
Our experiments demonstrate that these connections significantly accelerate model convergence and enhance training efficiency, facilitating the generation of high-quality protein sequences with improved stability.

\section{Experiments}

In this section, we first identify which architectural components significantly affect generation quality, diversity, and distributional properties (§\ref{subsec:ablation}). We then show that our architecture maintains strong performance across protein encoders of different sizes and dataset scales, without requiring modifications (§\ref{subsec:encoders}). We conclude by demonstrating practical applications in family-specific design (§\ref{section:family}), motif scaffolding (§\ref{section:scaffolding}), and fold-conditioned generation (\ref{section:fold_cond}), using quantitative metrics to assess success in each task.

\subsection{Evaluation Metrics}

We evaluate protein generation using four key metrics that capture essential aspects of the task: sequence Fr\'echet distance (FD-seq) measures how well generated sequences match the distribution of natural proteins on the held-out test set, predicted Local Distance Difference Test (pLDDT) assesses the structural plausibility of generated proteins, clustering density at 50\% sequence identity (CD$_{0.5}$) quantifies sequence diversity by measuring the fraction of distinct protein clusters in generated samples, and novelty scores evaluate similarity to training data to detect potential memorization. These metrics complement each other, allowing us to evaluate both the quality and diversity of generated proteins across sequence and structure modalities.

Our comprehensive evaluation framework also includes structural metrics (TM-score, scPerplexity), additional distributional similarity measures (maximum mean discrepancy, 1-Wasserstein optimal transport) computed on both sequence and structure representations, perplexity scores from state-of-the-art pLMs, and diversity assessment through multiple sequence identity thresholds. To ensure robustness, we evaluate all metrics on large sample sizes of 2,048 sequences and validate against independent test sets. While we present key findings in the main text for clarity, each experimental result is accompanied by a complete evaluation using the full metric suite in the corresponding appendix subsection. Detailed descriptions and implementation details of the evaluation metrics are provided in Appendix~\ref{appendix:metrics}.


\subsection{Denoiser Component Analysis}
\label{subsec:ablation}

\vspace{-5pt}
Existing protein generation methods based on Gaussian diffusion \citep{lee_score-based_2023, zhang2023pro} insufficiently address the selection of optimal methodologies, largely relying on techniques adapted from image diffusion models.
In this study, we recognize the need to carefully select the diffusion components, in order to develop a Gaussian diffusion model that can effectively capture the complex patterns of protein space.

In this part of our study, we utilize the ESM-8M encoder and DiMA-35M model for our experiments.
To assess the contribution of the proposed design choices to the performance of DiMA, we train several models from scratch with the following modifications: removing the long \textbf{skip-connections} between the shallow and the deep transformer blocks; using \textbf{time conditioning} through admixing the time embeddings to the corrupted latent vectors of amino acids instead of employing a dedicated time layer before each transformer block; omitting the transformer \textbf{encoder} (ESM-2), retaining only its embedding matrix; training the model without \textbf{self-conditioning}; training models with \textbf{linear and cosine noise schedule}; training models with padding reconstruction and without prior \textbf{length sampling}; omitting \textbf{finetuning the decoder}; and using \textbf{flow matching} paradigm in our latent generative model.

\begin{table}[t]
   \centering
   \small
   \caption{\sl Ablation study of key components of DiMA trained on SwissProt dataset using ESM-8M encoder. The complete results are presented in Table \ref{tab:ablation:total}.}
   \label{tab:main:ablation}

  \resizebox{\linewidth}{!}{ %
   \begin{tabular}{lcccc}
   \toprule
    \bf Model 
   & \bf{FD-seq} ($\downarrow$) 
   & \bf{pLDDT} ($\uparrow$) 
   & \bf{CD$_{0.5}$} ($\uparrow$)
   & \bf{Novelty} ($\uparrow$)
   \\
   \midrule

   Dataset                                & 0.13                & 80.7               & 1.000    &    25.3      \\
   Random sequences                       & 3.97                & 24.8               & 1.000    &    85.1       \\
   \cmidrule[0.5pt](lr){1-5}
   \chl DiMA                              & \chl \textbf{0.34}  & \chl \textbf{83.3} & \chl 0.611 & \chl 35.7      \\
   \hspace{3pt} w/o skip connections      & 0.45                & 77.3               & 0.619    &     43.1        \\
   \hspace{3pt} w/o time layers           & 0.41                & 79.4               & 0.550    &     38.4         \\
   \hspace{3pt} w/o ESM encoder           & 1.07                & 62.7               & 0.619    &     50.0    \\
   \hspace{3pt} w/o self-conditioning     & 0.55                & 68.2               & 0.929    &     56.6    \\
   \hspace{3pt} w/o finetuned decoder	 & 0.54                & 80.1	            & 0.589      &    38.4  \\     
   \hspace{3pt} w/o length sampling       & 0.67                & 65.0               & 0.880    &     58.4    \\
   \hspace{3pt} w linear schedule         & 0.47                & 77.0               & 0.611    &     45.8    \\
   \hspace{3pt} w cosine schedule         & 0.94                & 54.1               & 0.878    &     67.0    \\
   \hspace{3pt} w flow-matching	         & 0.71                & 63.4               & \textbf{0.960}  & \textbf{72.2}  \\

  
  
   \bottomrule
   \end{tabular}
   }
   \vspace{-5mm}
\end{table}

Table~\ref{tab:main:ablation} demonstrates that each proposed feature contributes significantly to the model's performance individually. 
The most substantial decrease in both the quality and distribution similarity of the generated sequences occurs in the ablated models without the ESM-2 encoder, without length sampling, and when trained without self-conditioning. 
Removing skip-connections and time layers results in a less pronounced impact, but still significant decrease in repetitions of generated sequences and a slight improvement in overall quality.

To ablate the impact of the tan-$10$ noise schedule, we train our diffusion model with standard linear and cosine schedules, leaving other parameters intact.
We find that tan-$10$ significantly outperforms the cosine schedule in both quality and distribution similarity.
It also achieves less expressed but better results than the linear schedule.
The detailed results are provided in the Appendix~\ref{appendix:ablation}.

\subsection{Comparison Across Generative Paradigms}
\label{subsec:baselines}
\vspace{-5pt}
We evaluate DiMA, a latent Gaussian diffusion model trained on ESM-8M encodings, against various generative models for protein sequence generation in a small-scale setup.
To ensure a fair comparison, all models are trained from scratch with 35M parameters on SwissProt. 
For methods requiring predefined sequence lengths, we sample lengths from the training set distribution. 
This comparison focuses on sequence-based models with publicly available code.

We consider five groups of baselines:  
 autoregressive models (\textbf{RITA} \citep{hesslow2022rita}, \textbf{SeqDesign} \citep{SeqDesign}, \textbf{nanoGPT} \citep{nanoGPT}), score-based models (\textbf{Walk-Jump} \citep{frey2023protein}), generative adversarial networks (\textbf{ProteinGAN} \citep{ProteinGAN}), discrete diffusion models (\textbf{EvoDiff-OADM} \citep{EvoDiff}, \textbf{DPLM} \citep{wang2024diffusion}, \textbf{D3PM} \citep{austin2021structured}), and flow-based models (\textbf{DFM} \citep{campbell2024generative}).

Table ~\ref{tab:main:comparison} presents the result of the comparison of existing methods and DiMA.
The evaluation demonstrates that the proposed latent Gaussian diffusion approach performs exceptionally well in unconditional protein generation compared to alternative generative paradigms. 
DiMA achieves the best proximity to the distribution of real proteins on the held-out test set, while also showcasing high quality in the generated proteins.
NanoGPT, an autoregressive model, shows promising results but fails to match dataset-level metrics, struggling with quality and distribution alignment in protein space.
DPLM, a discrete diffusion model, generates structurally plausible proteins but suffers from excessive amino acid repetition and whole-sequence duplication, leading to low diversity and novelty. 
Compared to DiMA, DPLM exhibits threefold lower novelty and twice the repetition rate, despite close pLDDT scores (see Table~\ref{tab:main:comparison} and Figure \ref{fig:comparison}). This suggests that while DPLM captures structural features, it lacks the diversity essential for realistic protein generation.

In comparison, other baselines exhibit notably poorer performance. 
SeqDesign and ProteinGAN, initially designed for narrow classes of proteins, may not be suitable for training on diverse datasets.
While EvoDiff outperforms SeqDesign and ProteinGAN, it still demonstrates metric values closer to a random sample than to the dataset, consistent with observations in the original EvoDiff paper (Table S3 of \citep{EvoDiff}).

Detailed results with additional metrics are presented in the Appendix~\ref{appendix:comparison}.

\begin{table}[t]
   \centering
   \small
   \caption{\sl Performance comparison between DiMA and alternative generative models for the protein generation of the same parameter count trained on SwissProt dataset. The complete results are presented in Table \ref{tab:comparison:total}.}
   \label{tab:main:comparison}

\resizebox{\linewidth}{!}{ %
   \begin{tabular}{lcccc}
   \toprule
   \bf Model 
   & \bf{FD-seq} ($\downarrow$) 
   & \bf{pLDDT} ($\uparrow$) 
   & \bf{CD$_{0.5}$} ($\uparrow$)
   & \bf{Novelty} ($\uparrow$)
   \\
\midrule

   Dataset                                & 0.13 & 80.7               & 1.000 & 25.3         \\
   Random sequences                       & 3.97 & 24.8               & 1.000 & 85.1         \\

   \cmidrule[0.5pt](lr){1-5}
   Walk-Jump                              & 2.63 & 32.4               & \textbf{1.000}  & 82.2        \\
   RITA                                   & 1.19 & 43.9               & 0.988  & 60.4        \\
   proteinGAN                             & 2.94 & 30.4               & 0.955 & \textbf{83.5}         \\
   SeqDesign                              & 3.53 & 43.1               & 0.929  & 81.2        \\
   EvoDiff-OADM                           & 1.49 & 37.1               & 0.986  & 77.6        \\
   D3PM                              	  & 1.50 & 36.7               & 0.994  & 78.4        \\
   DFM                                    & 1.46 & 37.8               & 0.996  & 77.2        \\
   DPLM                                   & 0.50 &  \textbf{84.0}     & 0.494  & 11.5       \\
   nanoGPT                                & 1.24 & 61.0               & 0.900  & 53.7         \\
   \chl DiMA                              & \chl \textbf{0.34} & \chl 83.3    &\chl 0.611  & \chl 35.7 \\
   
   \bottomrule

   \end{tabular}
}
\vspace{-5mm}
\end{table}

\subsubsection{Biological relevance}
\label{subsec:biorelevance}

To explore the biological relevance of the generated sequences, we employ the established protein annotation tool InterProScan \citep{paysan2023interpro,jones2014interproscan}. We use three different Swissprot-trained models: DPLM, DiMA, and nanoGPT. Our analysis shows that DiMA and DPLM, models exhibiting high-quality metrics, consistently generate sequences with a high degree of annotation compared to the lower-performing nanoGPT (Figure \ref{fig:fig_annotation_InterPro_distrib}A). This pattern is further reflected through the annotation intersections, where DiMA and DPLM demonstrate more significant overlap in their annotations (Figure \ref{fig:fig_annotation_InterPro_distrib}B).

While both approaches achieve similar levels of annotated proteins, their domain length characteristics differ. DiMA accurately reproduces dataset domain lengths and tends to generate small domains (50-75 amino acids). In contrast, DPLM frequently produces more extended domains (approaching 254 amino acids in length) (Figure \ref{fig:fig_annotation_InterPro_distrib}C). We hypothesize that the prevalence of long domains in DPLM correlates with its lower generation diversity, as evidenced by our diversity and distribution similarity metrics (Table~\ref{tab:comparison:total}).

\begin{table}[t]
   \centering
   \small
   \caption{\sl \small Performance of protein sequence generation using DiMA and different encoders on AFDBv4-90 dataset. The complete results are presented in Table~\ref{tab:encoders:total}}
   \label{tab:main:encoders_afdb}

   \resizebox{\linewidth}{!}{ %
  \begin{tabular}{lcccc}
   \toprule
    \bf Encoder 
   & \bf{FD-seq} ($\downarrow$) 
   & \bf{pLDDT} ($\uparrow$) 
   & \bf{CD$_{0.5}$} ($\uparrow$)
   & \bf{Novelty} ($\uparrow$)
   \\
  
\midrule      
     Dataset           & 0.11              & 83.9                  & 0.994            & 57.6 \\
     Random           & 2.55              & 22.16                  & 1.000            & 84.7 \\
\midrule     
     ESM-2 8M           & 0.560              & 74.25                  & 0.981            & 68.0 \\
     ESM-2 35M	      & 0.340              & 75.71                  & 0.986             & \textbf{69.1} \\
     ESM-2 150M         & 0.323              & 80.07                  & \textbf{0.988}   & 65.6 \\
     ESM-2 650M         & 0.318              & 82.48                  & 0.986            & 64.1 \\
     ESM-2 3B           & \textbf{0.314}     & \textbf{83.40}         & 0.969            & 63.0 \\
     ESMc 300M        & 0.326              & 82.70                  & 0.963              & 64.2 \\
\midrule                                                               
     CHEAP-shorten-1   & 0.346              & 81.92                  & 0.951             & 64.6 \\
     CHEAP-shorten-2   & 0.340              & 78.81                  & 0.946             & 66.2 \\
    \midrule 
     SaProt 35M        & 0.366              & 82.23                  & 0.976             & 65.5 \\
     SaProt 650M       & 0.411              & 83.01                  & 0.980             & 65.7 \\
   \bottomrule
   \end{tabular}
   }

   \vspace{-5mm}
\end{table}
\subsection{Representation Space Scaling}
\label{subsec:encoders}

In the previous sections, we established a robust diffusion model through comprehensive ablation studies and demonstrated its strong performance against baseline generative approaches. Building on this foundation, we now explore the model's adaptability across diverse protein representation spaces using the large-scale AFDBv4-90 dataset.

The AFDBv4-90 dataset represents a carefully curated subset of UniRef50. Comprising 2.2 million protein sequences with $>90\%$ AlphaFold2-predicted structural confidence, this dataset by design excludes intrinsically disordered proteins and low-entropy sequences, ensuring a high-quality protein corpus. As a consequence, this curation allows us to use pLDDT as a reliable, consistent measure of protein structural quality. 

We maintain the core denoising model architecture developed in our ablation studies (§\ref{subsec:ablation}), using only simple linear projection techniques to adapt to different encoder dimensions. Here, we examine three representative encoder architectures:

\begin{compactitem}
    \item The ESM-2 sequence-only encoder family, spanning models from 8M to 3B parameters, allows us to investigate performance scaling with model complexity (§\ref{subsec:enc_esm2}).
    
    \item CHEAP representations, which uniquely enable decoding into both sequence and 3D-structure from its latent spaces (§\ref{subsec:enc_cheap}).
    
    \item The SaProt encoder, which integrates structural tokens from FoldSeek's 3Di vocabulary, introducing a hybrid approach to sequence representation with structural awareness (§\ref{subsec:enc_saprot}).
\end{compactitem}

By preserving the core diffusion methodology, we aim to understand how different representation spaces influence protein sequence generation, providing insights into the generative capabilities across diverse protein embedding approaches.

\subsubsection{Performance Scaling with ESM-2 Encoders}
\label{subsec:enc_esm2}

We analyze the latent spaces of the ESM encoder family (ESM-2 8M through ESM-2 3B), investigating how performance scales with model capacity (Table \ref{tab:main:encoders_afdb}). DiMA successfully generates designable proteins across the entire encoder spectrum, with generation quality (pLDDT) consistently improving from ESM-2 8M (74.3) to ESM-2 3B (83.4). 

The scaling analysis reveals a key trade-off between quality and diversity. While larger encoders demonstrate enhanced precision in capturing protein patterns and improved distributional matching (FD-seq decreases from 0.560 to 0.314), they show reduced sequence diversity ($\text{CD}_{0.5}$ declines from 0.981 to 0.969). This pattern manifests in the repetition metric -- smaller encoders generate more repetitive patterns, while larger ones produce more refined but potentially less novel sequences (Table \ref{tab:encoders:total}).

The recently introduced ESM-C 300M \cite{ESMC} matches ESM-2 650M in generation quality (82.7 vs 82.5 pLDDT) but shows slightly reduced coverage of protein space (FD-seq 0.326 vs 0.318, $\text{CD}_{0.5}$ 0.963 vs 0.986). The marginal quality improvement from ESM-2 650M to ESM-2 3B suggests that mid-range architectures may offer an optimal balance between performance and computational efficiency.

This systematic evaluation demonstrates that DiMA can effectively leverage encoder representations of varying complexity while maintaining a lightweight deployment footprint, as the encoder becomes dispensable during inference.

\subsubsection{Adaptation to CHEAP Representations}
\label{subsec:enc_cheap}

CHEAP is an encoder that enables efficient dual-modal representations of proteins, providing access to both sequence and structure information from sequence input alone. Based on ESMFold, it aggregates information from all layers of ESM-2 3B encoder and compresses the continuous space, achieving significant dimensionality reduction while preserving high-fidelity structural and sequence reconstruction. We explore two compression variants: CHEAP\_shorten\_1\_dim\_1024 and CHEAP\_shorten\_2\_dim\_1024, both reducing the channel dimension while the latter additionally compresses the sequence length dimension.

\vspace{-1em}
\paragraph{Sequence Generation.}
DiMA trained with CHEAP encoder demonstrates strong performance in sequence space (Table \ref{tab:main:encoders_afdb}). Both compression variants maintain high generation quality: CHEAP\_shorten\_1\_dim\_1024 achieves pLDDT of 81.9 while CHEAP\_shorten\_2\_dim\_1024 reaches 78.8, approaching the dataset quality benchmark of 83.9. The variant without sequence length reduction shows superior performance, suggesting that preserving the full sequence dimension benefits generation quality.

\vspace{-1em}
\paragraph{Structure Generation.}
We assess structural consistency through two evaluation protocols: co-design, comparing generated backbones against structures reconstructed from predicted sequences, and structure-only evaluation using standart protocol desribed in \citet{FrameDiff} (Table \ref{tab:structure_generation}). The structure-only approach achieves 92.3\% success rate with mean scRMSD of 1.091, while co-design demonstrates strong performance with 88.8\% success rate and mean scRMSD of 1.043\AA. These results indicate that DiMA effectively leverages CHEAP's compressed yet information-rich latent space for both sequence and structure generation. Notably, the sequence length reduction in CHEAP\_shorten\_2 enables using half-length transformer context for the denoising model, offering substantial computational advantages while maintaining generation quality.

Additional results for the DiMA model using CHEAP encoders for structure generation are available in Appendix~\ref{app:structure}.



\subsubsection{Adaptation to SaProt Encoder}
\label{subsec:enc_saprot}
Multimodal SaProt encoder \cite{su2023saprot} presents an alternative approach to protein representation, integrating both sequence and local structural information. Unlike sequence-only models, SaProt enriches its representations using structural tokens from FoldSeek's \cite{FoldSeek} 3Di vocabulary, providing compact yet informative structural descriptors.

We evaluate SaProt encoders of two sizes (35M and 650M parameters) using the same hyperparameters for the denoiser model that we established through our ablation studies. The quality metrics demonstrate that structural awareness in SaProt's representations translates to improved generation capabilities - models achieve higher pLDDT scores compared to same-sized ESM-2 variants (82.23 vs 75.71 for 35M and 83.01 vs 82.48 for 650M architectures, Table \ref{tab:main:encoders_afdb}). This improvement is particularly notable with the smaller 35M parameter encoder, suggesting that structural tokens provide an efficient way to encode protein properties.

The integration with SaProt demonstrates DiMA's ability to leverage different types of protein embeddings without architectural modifications. This adaptability, combined with strong performance in structure-aware tasks like motif-scaffolding (detailed in §\ref{section:scaffolding}), indicates that our diffusion framework can effectively capitalize on both sequence and structural information encoded in the latent space.

\subsection{Comparison with Large Pretrained Models}

In this section we compare DiMA-35M trained over ESM-3B encodings with existing large protein models, we demonstrate that DiMA achieves performance comparable to existing pre-trained large protein models. 

 \begin{figure}[t]
    \centering
    \includegraphics[width=0.48\textwidth]{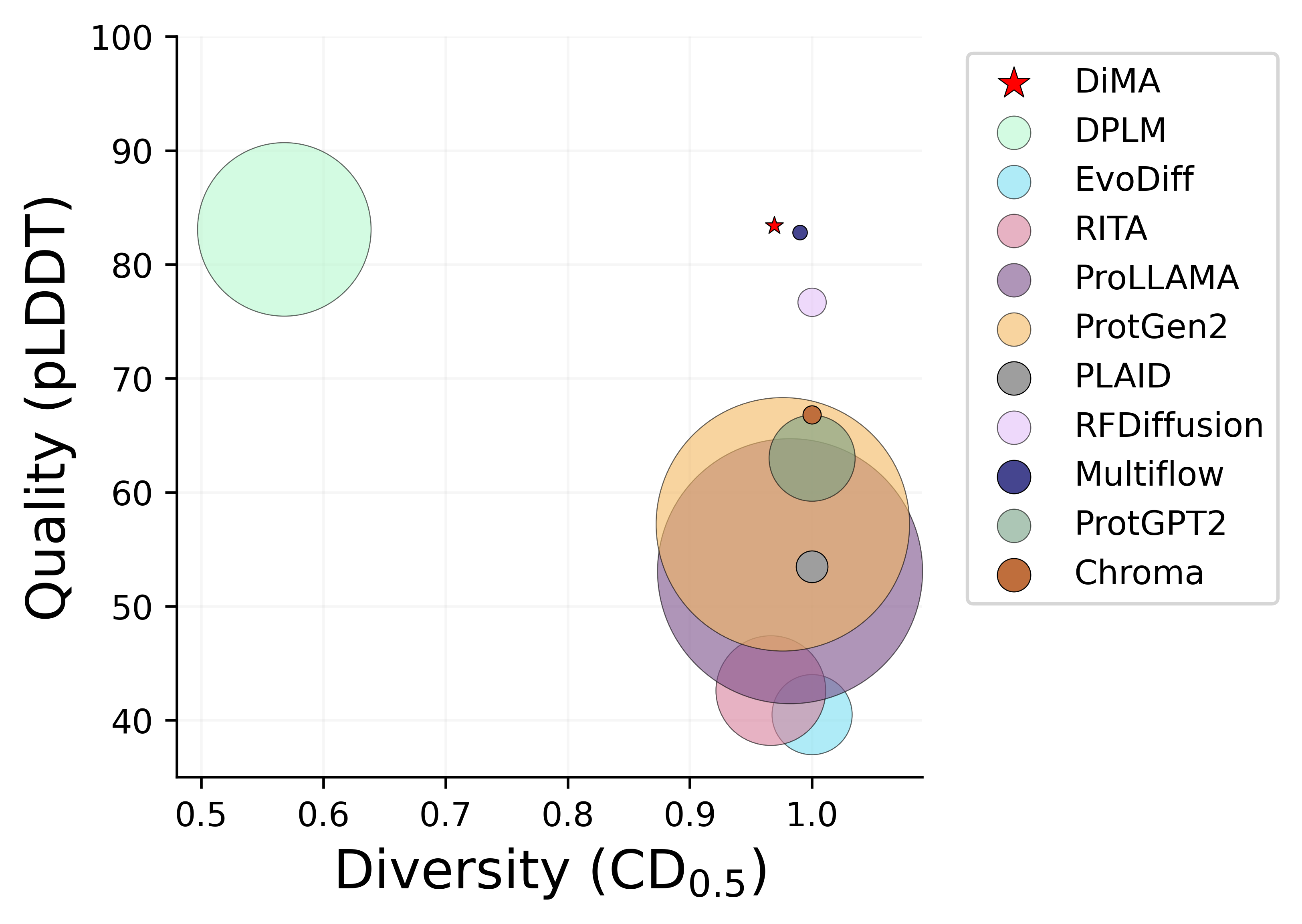}
    \vspace{-5mm}
    \caption{\small Comparison of DiMA with large pretrained protein generative models in quality and diversity. Circle size represents model scale. The complete results are presented in Table~\ref{tab:appendix:pretrain}.}
    \vspace{-5mm}
    \label{fig:main:pretrained_models}
\end{figure}

We evaluate DiMA-35M, trained on ESM-3B encodings, against state-of-the-art pre-trained protein models, namely, 
RITA~\citep{hesslow2022rita}, 
ProtGPT2~\citep{protgpt2}, 
ProGen2~\citep{ProGen_Nature}, 
EvoDiff~\citep{EvoDiff}, 
ProLLAMA~\citep{lv2024prollama}, 
DPLM~\citep{wang2024diffusion}, 
Chroma~\citep{ingraham2023illuminating}, 
Multiflow~\citep{campbell2024generative}, 
RFDiffusion~\citep{RFDiffusion}, and PLAID-100M~\citep{PLAID}.
For all models, we use the authors' recommended sampling parameters to ensure a fair comparison. However, for autoregressive models ProGen2 and ProLLAMA, which show suboptimal quality and collapse to highly repetitive sequences on default settings, we performed grid searches to identify optimal temperature and top-p values.
We consider only models with publicly available pre-trained weights, ensuring transparency and reproducibility.
Figure~\ref{fig:main:pretrained_models} illustrates the relationship between quality, diversity, and model size.

DiMA demonstrates high-quality and diverse protein generation, achieving performance on par with much larger models despite using two orders of magnitude fewer parameters. While DPLM and other diffusion-based models excel in structural plausibility, they lag in diversity, where DiMA outperforms them significantly.
Notably, Multiflow, trained with structural data, achieves similar performance to DiMA despite operating at a comparable parameter scale. 
An important advantage of DiMA is that it achieves performance comparable to models trained using structural information, while being trained exclusively on amino acid sequences. 
This demonstrates the model's efficiency and ability to extract meaningful representations from sequence data alone, making it a highly versatile and resource-efficient solution for protein generation tasks.

These findings underscore DiMA's efficiency and scalability, establishing it as a compelling approach for protein sequence generation, even in comparison to large-scale pre-trained models. Full results across model sizes and evaluation metrics are provided in Appendix~\ref{appendix:pretrain} and Table~\ref{tab:appendix:pretrain}.

\subsection{Advanced generation tasks}

\subsubsection{Functional-motif Scaffolding}
\label{section:scaffolding}

We evaluate the conditional generation capabilities of DiMA on a challenging task of functional-motif scaffolding using the established RFDiffusion benchmark \cite{RFDiffusion}. This task requires designing entirely new protein structures that incorporate and preserve specific functional motifs.

We evaluate DiMA on 24 benchmark problems, where each problem requires generating protein sequences that maintain precise spatial positioning of functionally important residues. For conditioning, we augment DiMA with an encoder that provides motif information to each transformer layer. Following established protocols \cite{EvoDiff}, we sample 100 designs per problem and consider a problem solved if at least one design achieves both structural quality (pLDDT $\geq$ 70) and motif preservation (RMSD $\leq$ 1\AA{} for motif residues).

\begin{figure}{}
    \centering
    \includegraphics[width=0.48\textwidth]{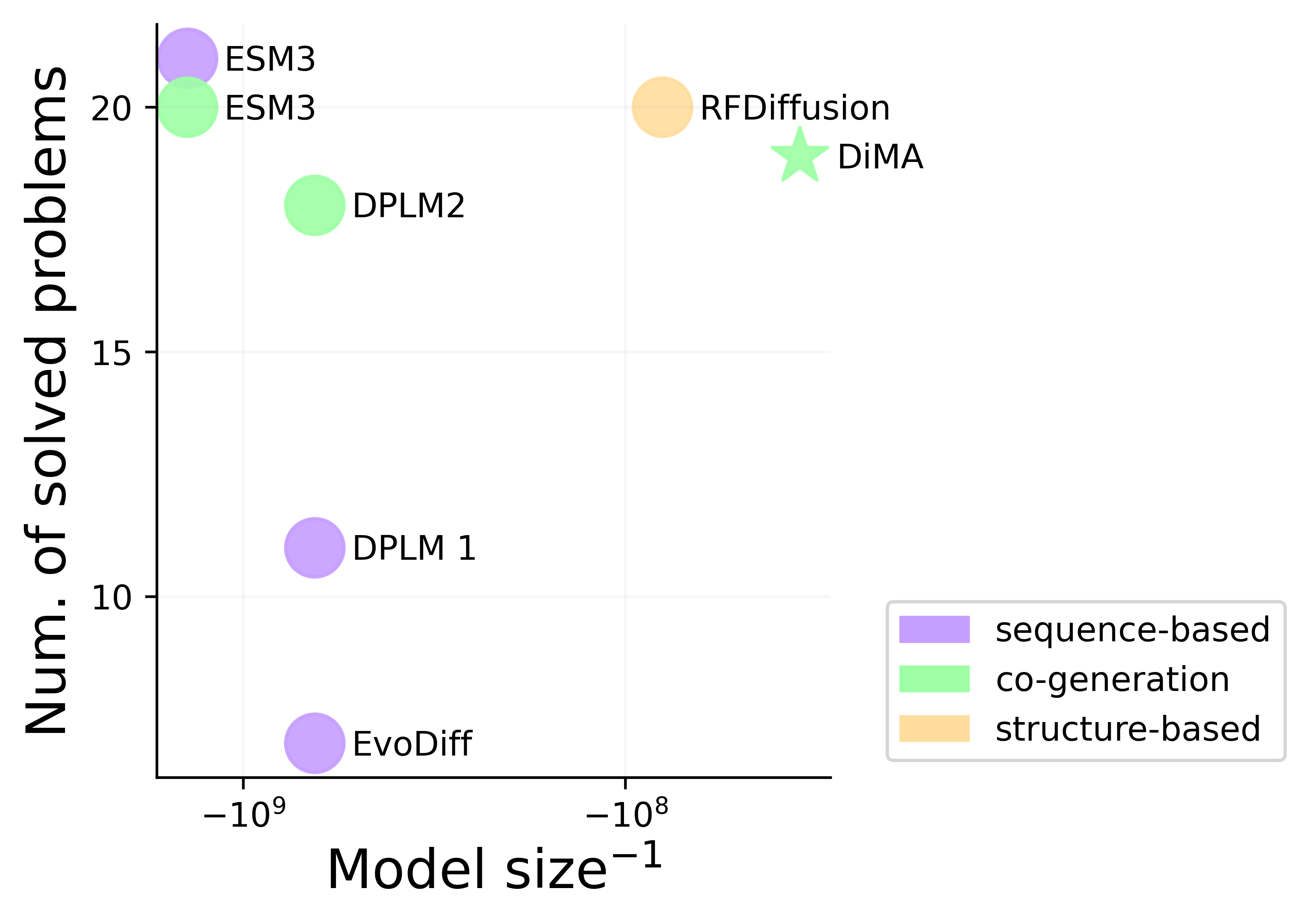}
    \vspace{-5mm}
    \caption{\small Motif-scaffolding: performance comparison across different model sizes. Methods are colored by input modality: sequence-based (purple), sequence-structure co-generation (green), and structure-based (orange). DiMA solves 19/24 benchmark problems while being significantly more compact than other high-performing models. The complete results are presented in Table~\ref{tab:condition:scaffolding_sr}}
    \vspace{-5mm}
    \label{fig:scaffolding_main}
\end{figure}

Using the SaProt-650M encoder, DiMA successfully solves 19 out of 24 problems, outperforming other sequence-based methods, including EvoDiff, DPLM, and DPLM2 (Figure~\ref{fig:scaffolding_main}). While structure-based methods like RFDiffusion achieve higher overall success rates, DiMA generates more diverse successful scaffolds as measured by unique success rate (0.1 vs 0.06 for RFDiffusion) matching the diversity of ESM-3 while using two orders of magnitude fewer parameters. Interestingly, DiMA and ESM-3 show complementary strengths across different scaffold types, with DiMA achieving the highest unique success rates on 6E6R-type problems. Complete results and experimental details are provided in Appendix~\ref{appendix:scaffolding}.

\subsubsection{Family-specific Generation}
\label{section:family}
Generating proteins that belong to specific functional families is a key task in protein engineering, enabling targeted exploration of sequence spaces with desired characteristics. We investigate two approaches for training DiMA (35M parameters) to perform family-specific generation: classifier guidance and conditional fine-tuning.

Using ESM2-650M encodings and the AFDBv4\_90 dataset, we train and evaluate our model on eight diverse protein families, including CRISPR-associated proteins, calmodulins, and glycosyl hydrolases. For classifier guidance, we train a lightweight classifier (3 transformer blocks) on noisy protein encodings to predict family membership. For conditional fine-tuning, we augment DiMA with family class label embeddings and fine-tune on all families simultaneously. We compare against significantly larger baselines, including ProLLAMA (7B parameters), ProGen2 (151M parameters), and EvoDiff (640M parameters).

We evaluate generations using multiple complementary metrics: InterProScan for family membership verification (Fidelity), pLDDT for structural quality assessment, and cluster diversity at 50\% sequence identity (CD\textsubscript{0.5}) to measure generated sequence diversity. Both DiMA variants achieve high fidelity to target families while maintaining structural quality. While autoregressive models demonstrate high sequence diversity but struggle with fidelity and quality (pLDDT $\approx$ 60), discrete diffusion achieves comparable fidelity and quality but generates fewer novel proteins. Notably, the classifier-guided variant achieves competitive performance without requiring model fine-tuning, offering a practical advantage for targeted protein design. Complete results and experimental details are provided in Appendix~\ref{appendix:family}.


\begin{figure}[!htb]
    \centering
    \scalebox{0.7}{
    \begin{tikzpicture}[remember picture]
        \node[anchor=south west,inner sep=0] (image1) at (0,11)
            {\includegraphics[width=0.35\textwidth]{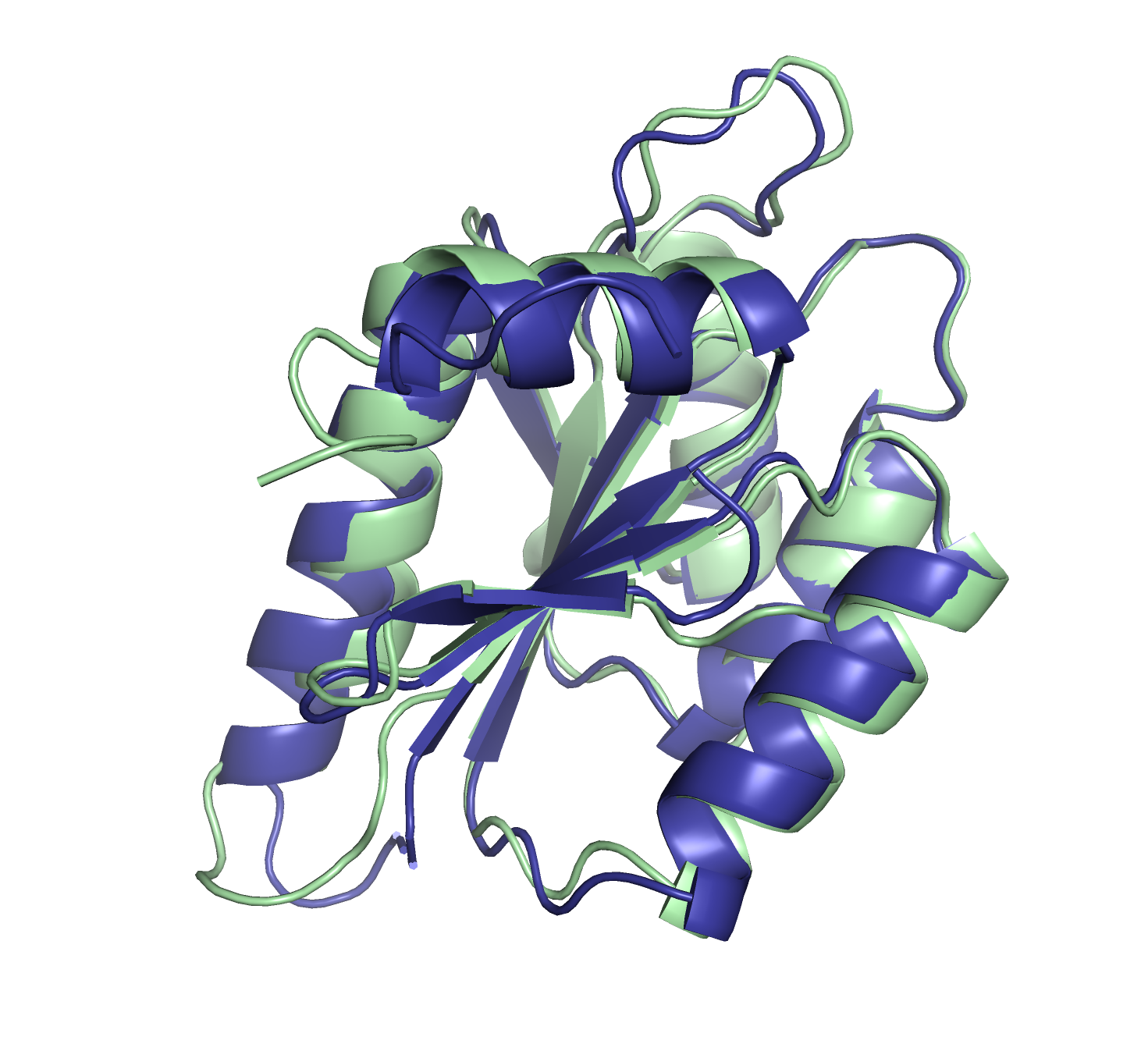}};
        \node[anchor=north,fill=white,font=\large] at (image1.north) {Seq. identity: 46.30\%};
        
        \node[anchor=south west,inner sep=0] (image2) at (0.35\textwidth,11)
            {\includegraphics[width=0.35\textwidth]{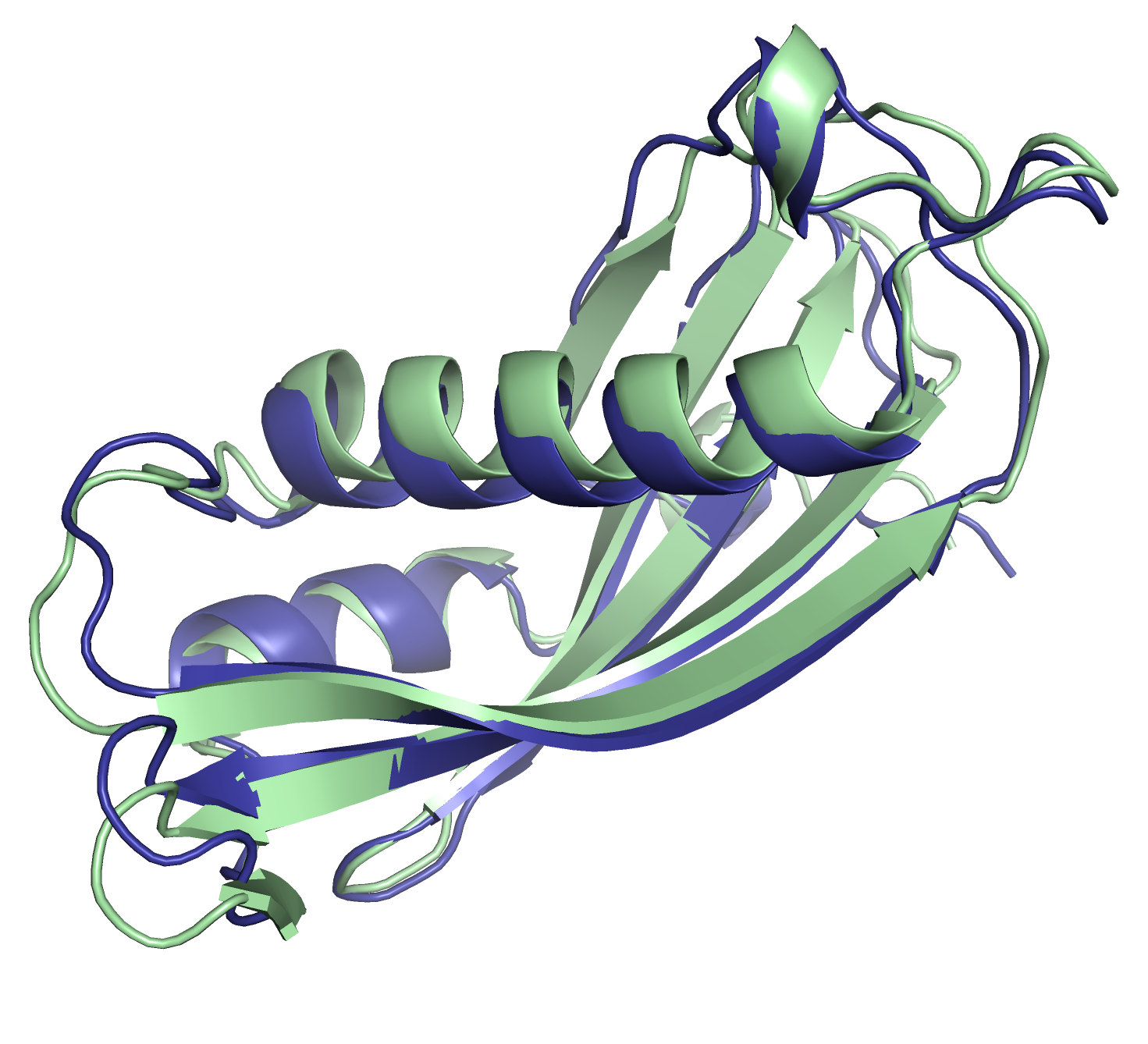}};
        \node[anchor=north,fill=white,font=\large] at (image2.north) {Seq. identity: 36.08\%};
        
        \node[anchor=south west,inner sep=0] (image3) at (0,5.5)
            {\includegraphics[width=0.35\textwidth]{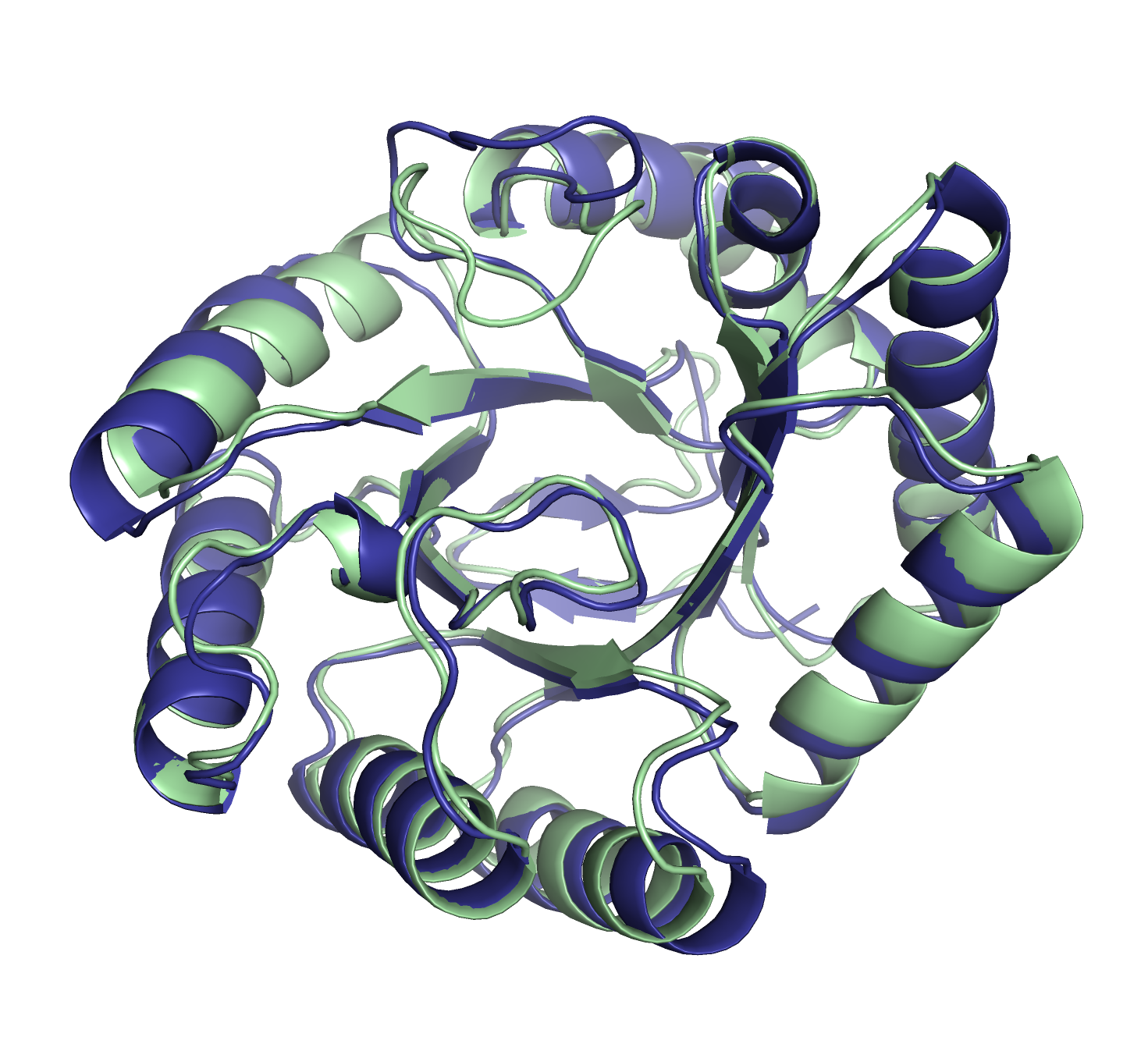}};
        \node[anchor=north,fill=white,font=\large] at (image3.north) {Seq. identity: 34.41\%};
        
        \node[anchor=south west,inner sep=0] (image4) at (0.35\textwidth,5.5)
            {\includegraphics[width=0.35\textwidth]{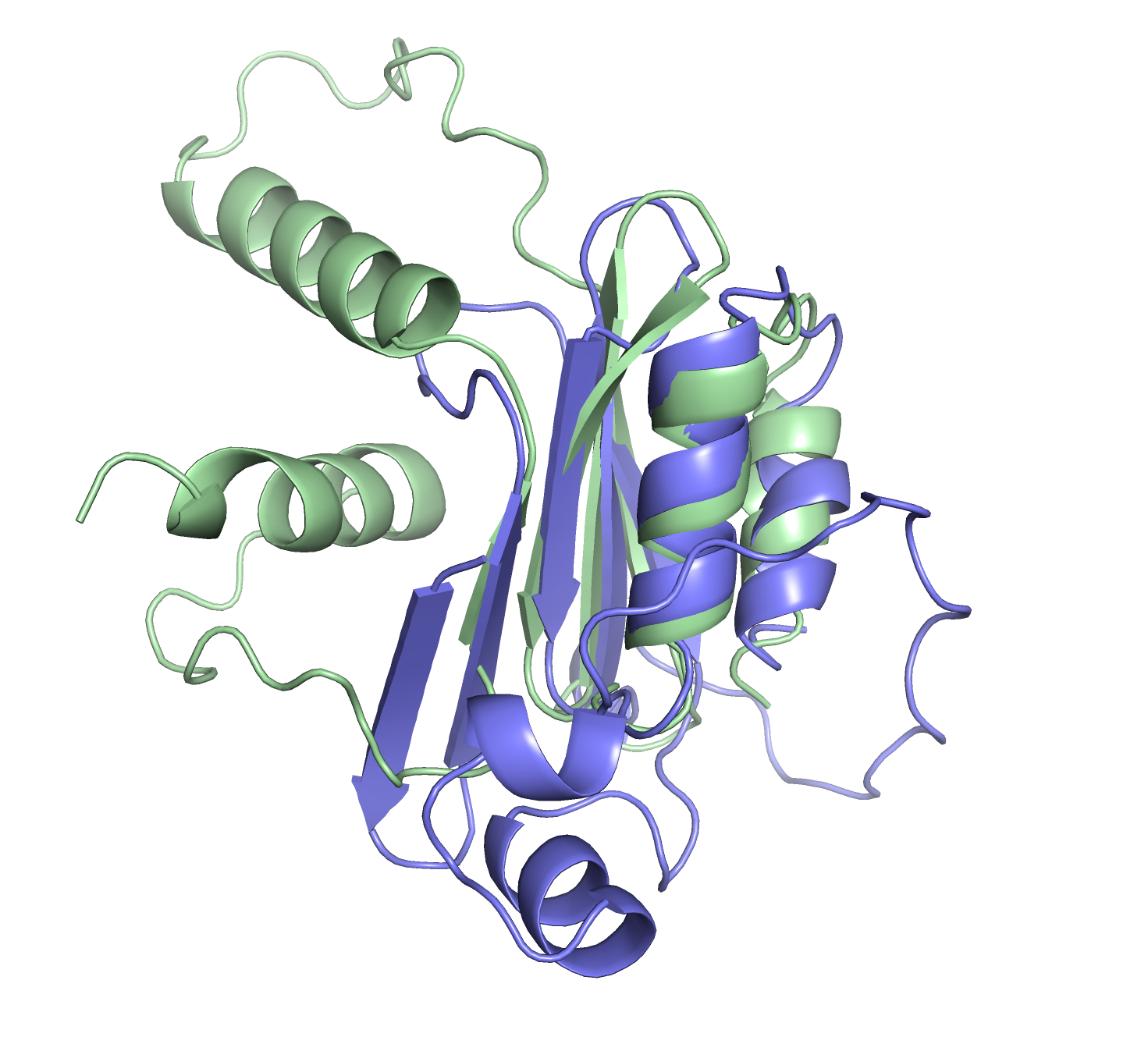}};
        \node[anchor=north,fill=white,font=\large] at (image4.north) {Seq. identity: 30.13\%};
        
        \node[anchor=south west,inner sep=0] (image5) at (0,0)
            {\includegraphics[width=0.35\textwidth]{figures/fold_cond/00739.png}};
        \node[anchor=north,fill=white,font=\large] at (image5.north) {Seq. identity: 36.08\%};
        
        \node[anchor=south west,inner sep=0] (image6) at (0.35\textwidth,0)
            {\includegraphics[width=0.35\textwidth]{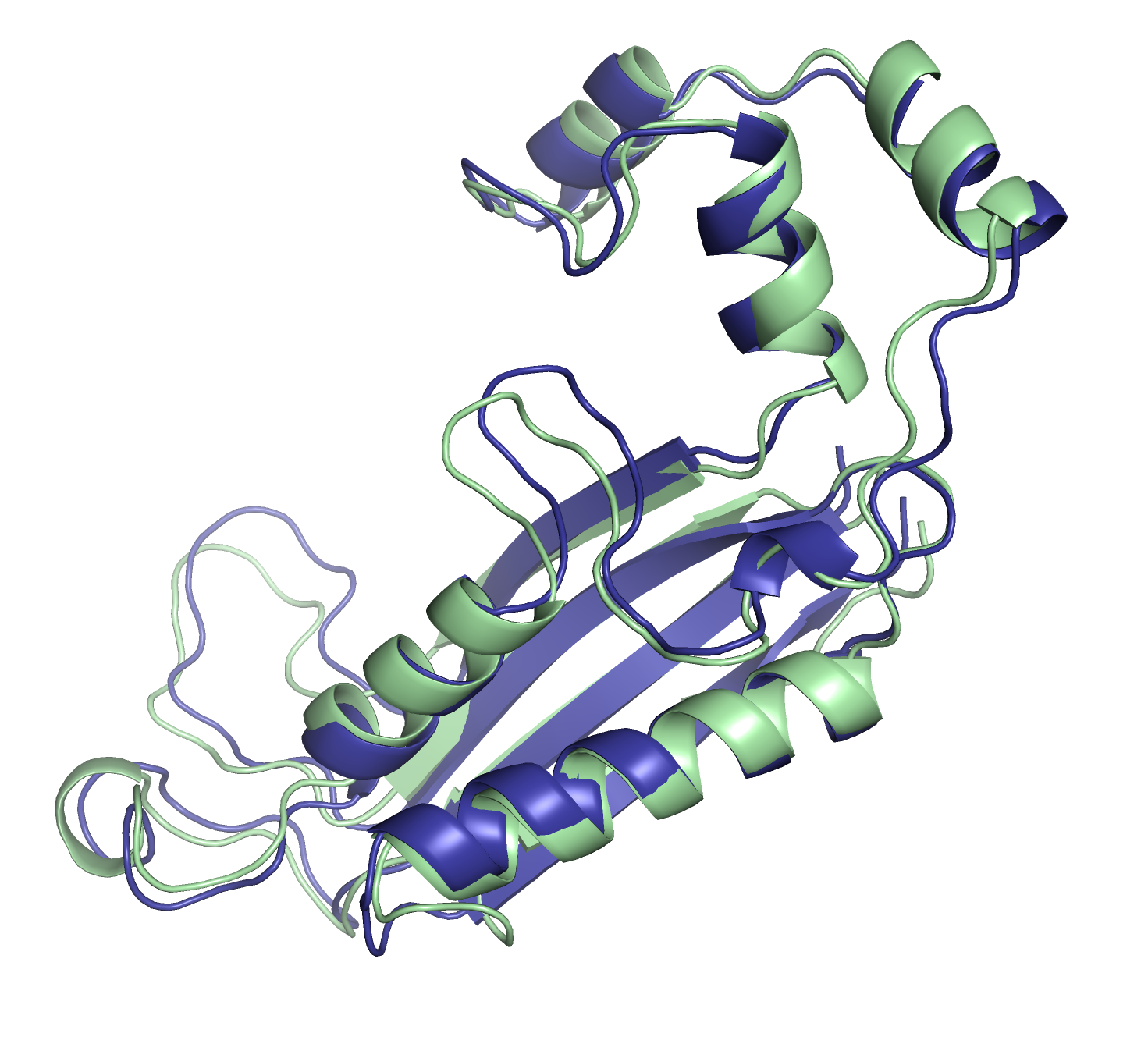}};
        \node[anchor=north,fill=white,font=\large] at (image6.north) {Seq. identity: 38.50\%};
    \end{tikzpicture}
    } 
    \vspace{-1em}
    \caption{Examples of successful proteins generated via fold-conditioning, aligned on the corresponding target proteins. Sequence identity percentage between the target protein and the generated one is reported for each design.}
    \vspace{-5mm}
    \label{fig:fold_cond_figure}
\end{figure}

\subsubsection{Fold-conditioned Generation}
\label{section:fold_cond}

Generation of proteins with specific structural properties represents a distinct challenge from sequence-based tasks, requiring the model to capture complex three-dimensional relationships. We explore DiMA's capabilities in fold-conditioned generation using the CHEAP encoder (shorten=1, dim=1024), which provides access to ESMFold's latent space representation of protein structure.

We finetune DiMA on the CATH S40 non-redundant dataset (~27k proteins) and evaluate performance on a hold-out set of 100 structures. For each structure, we generate 10 proteins and assess their similarity to the target fold using the TM-score. Following the protocol of \cite{RFDiffusion}, we consider generation successful if at least one design achieves TM-score $> 0.5$, indicating significant structural similarity. For comparison, we evaluate against RFDiffusion, a specialized structure generation model capable of fold-conditioning.

DiMA achieves a mean TM-score of 0.93 with a 100\% success rate across the benchmark set, compared to RFDiffusion's mean TM-score of 0.48 and 41\% success rate. The average RMSD between the best attempts and their target folds is 2.6\AA, indicating that DiMA generates structurally similar but non-identical proteins. This performance difference likely stems from DiMA's use of richer structural encodings compared to RFDiffusion's secondary structure and block-adjacency representations. Figure~\ref{fig:fold_cond_figure} illustrates examples of generated proteins alongside their target folds. Complete experimental details are provided in Appendix~\ref{appendix:fold}.

\section{Conclusion}
In this paper, we introduce DiMA, a continuous latent diffusion framework for protein sequence generation that operates effectively across diverse protein language model representations. Through systematic exploration of architectural choices and diffusion components, we establish a robust methodology that generalizes from sequence-only encoders to multimodal representations spanning 8M to 3B parameters.

DiMA demonstrates versatile functionality across multiple protein design scenarios, including unconditional generation, family-specific design, motif scaffolding, sequence infilling, and fold-conditioned generation. The framework achieves competitive performance with significantly larger models while maintaining computational efficiency, as the encoder becomes dispensable during inference.

This work provides both architectural insights and practical applicability for protein sequence generation, offering a unified approach that bridges sequence and structure generation through continuous diffusion. The systematic methodology we establish creates a foundation for future developments in computational protein design, demonstrating how domain-specific adaptations can unlock the potential of diffusion models for biological sequence generation.

\section*{Impact statement}
This work advances machine learning approaches for protein sequence generation, introducing a framework that bridges discrete and continuous methodologies. Beyond the core technical contributions to diffusion models and protein representation learning, our work has potential applications in therapeutic protein design and synthetic biology. While these applications could benefit society through new drug development and biotechnology advances, we acknowledge they require careful consideration of biosafety and ethical implications in deployment. We believe the benefits of advancing protein design capabilities outweigh potential risks when proper safeguards and responsible development practices are followed. Our framework emphasizes transparency through extensive evaluation metrics and validation protocols, promoting reproducible and responsible progress in computational biology.

\bibliography{main}

\bibliographystyle{icml2025}

\newpage
\appendix
\onecolumn
\section*{Appendix}
\addtocontents{toc}{\protect\setcounter{tocdepth}{2}}
\addtocontents{toc}{\string\renewcommand{\string\cftsecfont}{\string\normalfont}}

\tableofcontents
\vspace{1em}  
\hrule        

\newpage
\section{Datasets}\label{subsec:datasets}
SwissProt is a dataset that contains a high-quality, manually annotated subset of the UniProt \citep{uniprot2021uniprot} database. This dataset is small enough and good enough for proof-of-the-concept studies. After filtering out all sequences shorter than 128 and trimming all sequences longer than 254, we ended up with ~470k sequences. MMseqs2 clustering of this dataset (\textgreater50\% sequence identity and \textgreater80\% sequence overlap) reveals the presence of clusters of similar sequences with the maximum number of sequences in a cluster equal to 1570. Each of those clusters comprises sequences that belong to a single protein family. For example, the most populous cluster is 1570 protein sequences of cytochrome b of different species, a very abundant protein involved in electron transport in eukaryotic cells. Around 120k sequences do not form clusters under the conditions used.

Another dataset we use is AFDBv4-90 from \citet{AFDBv4_90}, a subset of the UniRef50 database. The sequences in this dataset obey two conditions: 1. The sequence identity between all members is no more than 50\%, and 2. The average predicted pLDDT by AlphaFold is no less than 90. After filtering out all sequences shorter than 128 and longer than 254, we ended up with ~2.2 million whole sequences of highly diverse proteins of high quality.

\section{Evaluation Metrics}\label{appendix:metrics}
\subsection{Quality}
\label{appendix:metrics:quality}
\textbf{pLDDT}. 
To assess the foldability of our generated sequences, we utilize ESMfold to predict the three-dimensional structure of the given protein sequence. 
For each amino acid within the predicted structure, ESMfold provides a pLDDT score, which represents the confidence of the model in the predicted positions of amino acids in the 3D structure. 
We average these pLDDT scores for all amino acids in the sequence to gauge the overall confidence in the predicted protein structure. 
It is worth noting that, while higher average pLDDT scores indicate a reliable structure prediction, lower scores may not necessarily denote poor prediction. 
In some cases, they can also signify the presence of intrinsically disordered regions in the protein, segments that are inherently flexible and do not conform to a fixed structure but still play vital roles in protein functionality \citep{plddt_IDP, shukla_intrinsic_2023}.


\textbf{ProGen perplexity}
To assess how probable the generated sequences we utilize the ProGen2-base \citep{ProGen_Nature} model of 764M parameters to estimate perplexity.
\begin{equation}
\mathcal{P}_{ProGen}(S) = \exp \left\{ -\frac{1}{|S|} \sum_{i=1}^{|S|} \log p(s_i | S_{< i}, \Theta_{ProGen-base}) \right\}
\end{equation}

\textbf{ESM-2 pseudoperplexity}. 
To assess how probable the original sequence is under the model's distribution, we used pseudoperplexity \citep{MLM_pseudoperplexity} using ESM-2 650M encoder transformer-based language model \citep{ESM-2}. Each token (amino acid) in the sequence was masked and then predicted, considering all other tokens in the sequence.
The final pseudoperplexity value is aggregated using the following equation:

\begin{equation}
\mathcal{P}_{ESM-2}(S) = \exp \left\{ -\frac{1}{|S|} \sum_{i=1}^{|S|} \log p(s_i | S_{\backslash i}, \Theta_{ESM-2}) \right\}
\end{equation}
Here, $\mathcal{P}_{ESM-2}(S)$ represents the pseudoperplexity of sequence $S$, $|S|$ denotes the length of sequence $S$, $s_i$ is the $i$-th token in the sequence, $S_{\backslash i}$ represents the sequence without the $i$-th token, and $\Theta_{ESM-2}$ denotes the parameters of the ESM-2 model.


\textbf{TM-score}. To evaluate the structural relevance of the generated sequences, we turned to the TM-score \citep{TM-score}, a widely recognized metric for evaluating structural similarity between protein pairs.
The TM-score measures the similarity between two protein structures and helps distinguish proteins with a similar fold from those with different folds. 
Unlike many other metrics for 3D-alignment, it does not depend on protein size and always ranges between 0 and 1, where a TM-score above 0.5 indicates a similar fold in structure.
The TM-score is given by:
\begin{equation}
    \text{TM-score} = \frac{1}{L_{\text{target}}} \sum_{i=1}^{L_{\text{query}}} \frac{1}{1 + \left(\frac{d_i}{d_0(L_{\text{target}})}\right)^2}
\end{equation}

Here, $L_{\text{target}}$ is the length of the target protein, $L_{\text{query}}$ is the number of aligned residues between the two proteins, $d_i$ is the distance between the $i$-th aligned residue pairs, and $d_0$ is a scaling factor to normalize the length difference between query and target proteins.
To calculate TM-scores for each sample of generated sequences, we first obtained their 3D structures using ESMFold. 
For each of these structures, we have found the closest natural protein in the SwissProt and AFDBv4-90 datasets from the AlphaFold Database \citep{AlphaFoldDB} using the FoldSeek \citep{FoldSeek}.

\textbf{BLAST Identity.}
For each sequence, we ran BLAST with specific parameters (e-value = 0.05 and BLOSUM62 substitution matrix) to identify similar sequences within the training dataset. The number of matching amino acids between the generated sequence and the most identical sequence found in training data was normalized by sequence length and multiplied by 100 to obtain percentages. The BLAST identity metric is the average over a batch of 2048 sequences.



\begin{wraptable}{R}{0.5\textwidth}
  \centering
   \caption{\sl \small Review of the metrics across modalities
   for evaluating generation quality,
   diversity, novelty, and distribution similarity.}
   \label{tab:table_metrics_dstr}

   \begin{tabular}{lccc}
   \toprule
   & \bf{Sequence}
   & \bf{Structure}
    \\
    \midrule
    
    \makecell{ \textbf{Distributional}\\ \textbf{similarity}}
    & \makecell{FD-seq\\OT-seq\\MMD-seq}
    & \makecell{FD-struct\\OT-struct\\MMD-struct} \\
    
   \midrule

   \multirowcell{4}{\textbf{Quality}}
   & \makecell[t]{ProGen-2 ppl\\ESM-2 pppl\\BLAST\\scPerplexity}
    & \makecell[t]{pLDDT\\TM-score\\scPerplexity} \\
    
    \midrule
    
    \multirowcell{2}{\textbf{Diversity}}
    & \makecell[t]{Rep\\CD} \\

    \midrule
    
    \multirowcell{1}{\textbf{Novelty}}
    & Novelty \\

    \bottomrule
   \end{tabular}
\end{wraptable}

\subsection{Diversity}
\label{appendix:metrics:diversity}

\textbf{Rep}.
Rep quantifies the internal diversity of generated sequences by assessing the prevalence of repeated subsequences, it is calculated as \( \textbf{Rep} (y) = 1 -  \prod_{n\in \{8, 16, 32, 64\}} \frac{|\text{\# of unique n-subseq in } y|}{|\text{\# of n-subseq in } y|} \), where $y$ is a set of generated proteins.  
n-subseq means the subsequence of consecutive amino acids of length n.

\textbf{CD}.
To evaluate the model's capacity to generate distinct protein variants while avoiding redundant outputs we employ the \textbf{clustering density} metric ($CD_t$) at two sequence identity thresholds: $t=\%50$ and $t=95\%$. $CD_t$ represents the ratio of sequence clusters at threshold $t$ to the total number of generated proteins. Therefore, $CD_t$ ranges from 0 to 1, where 1 indicates that all sequences form individual clusters and the sample is diverse. $CD_{0.5}$ is an established metric for assessing broad sequence diversity \citep{uniprot2021uniprot}, analogous to the widely-adopted TM-score threshold of 0.5 used in structure generation \citep{FrameDiff}. We employ MMseqs2 \citep{mmseqs} to perform sequence-based clustering at given thresholds $t$ (coverage = 0.8, cov-mode = 0, cluster-mode = 1.).
While clustering at a moderate threshold ($50\%$) reveals the model's ability to generate diverse proteins, individual clusters may still contain nearly identical sequences—an undesirable characteristic for generative models. Therefore, we complement our analysis with $CD_{0.95}$, which specifically identifies near-duplicate sequences. This dual-threshold approach provides a more comprehensive assessment of sequence diversity compared to single-metric evaluations.

\textbf{PCD and NCD}.
While $CD_t$ can capture mode collapse in a batch of sequences, it also highly rates random sequences. To evaluate the degree of novelty of the generated sequences we perform \textbf{co-clustering analysis} of generated sequences with the dataset sequences using MMseqs2 (identity = 0.5, coverage = 0.8, cov-mode = 0, cluster-mode = 1). This analysis yields two metrics: $PCD_{0.5}$ and $NCD_{0.5}$, representing the ratios of "positive" clusters (PC, containing both generated and dataset sequences) and "negative" clusters (NC, containing only generated sequences) to the total number of sequences, respectively. The desired values of $PCD_{0.5}$ and $NCD_{0.5}$ should be close to reference ones. 

Notably, that generation out of distribution is also very important, so we evaluate the quality of generated sequences from other ("negative") clusters. We found that the average pLDDT of these sequences from DiMA (SwissProt and AFDBv4-90: 65 \textpm 14 and 63 \textpm 12, respectively), which is significantly higher than that of other models (nanoGPT: SwissProt and AFDBv4-90 43 \textpm 12 and 52 \textpm 16). This indicates that the model generalizes beyond the training data.

\begin{figure}[h]    
\centering    
\includegraphics[width=0.9\textwidth]{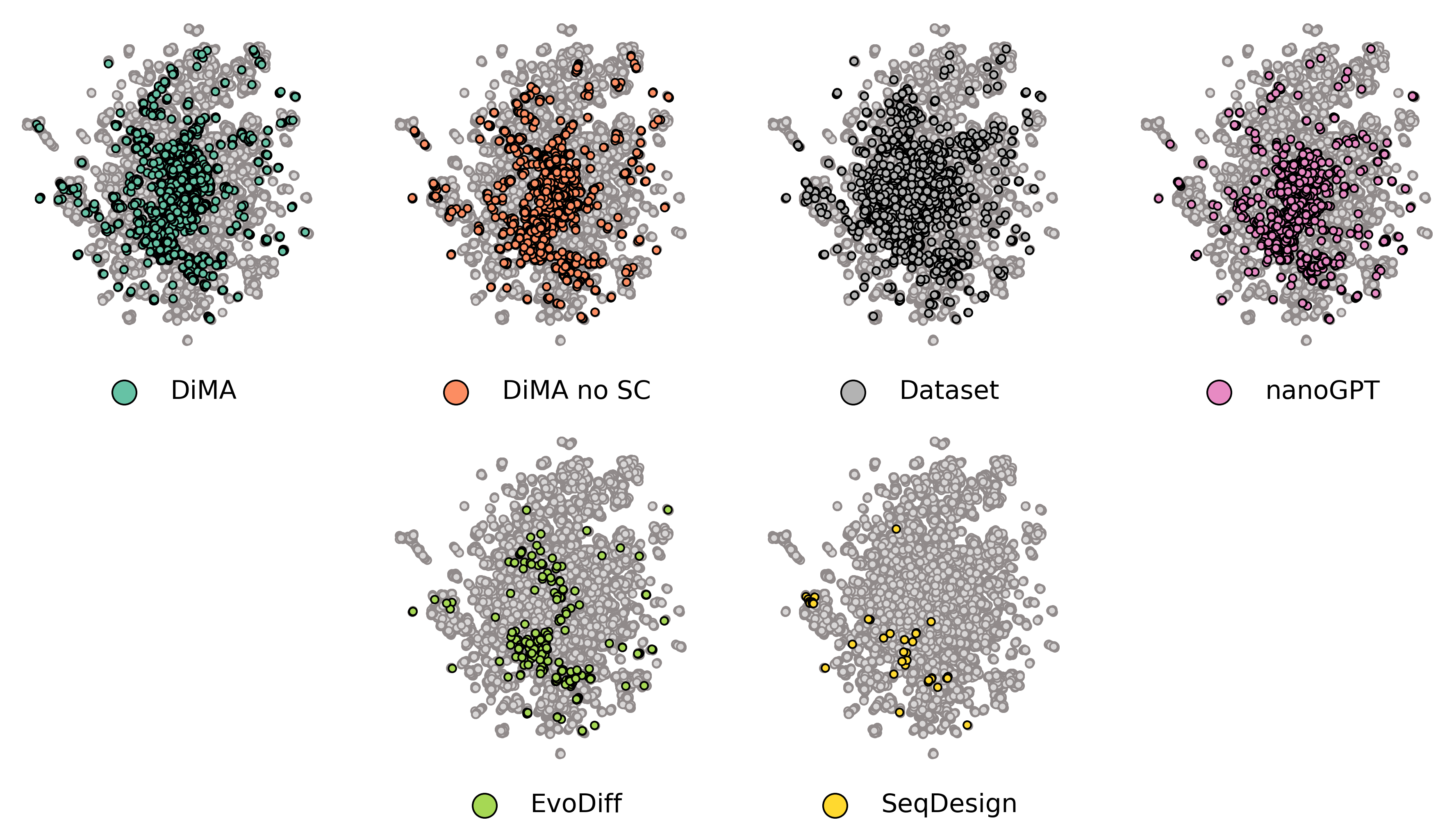}
\caption{UMAP projection of sequences from PC. Training dataset - SwissProt. Grey background points - dataset sequences from PC. }
\label{fig:umap_swissprot}
\centering    
\includegraphics[width=0.9\textwidth]{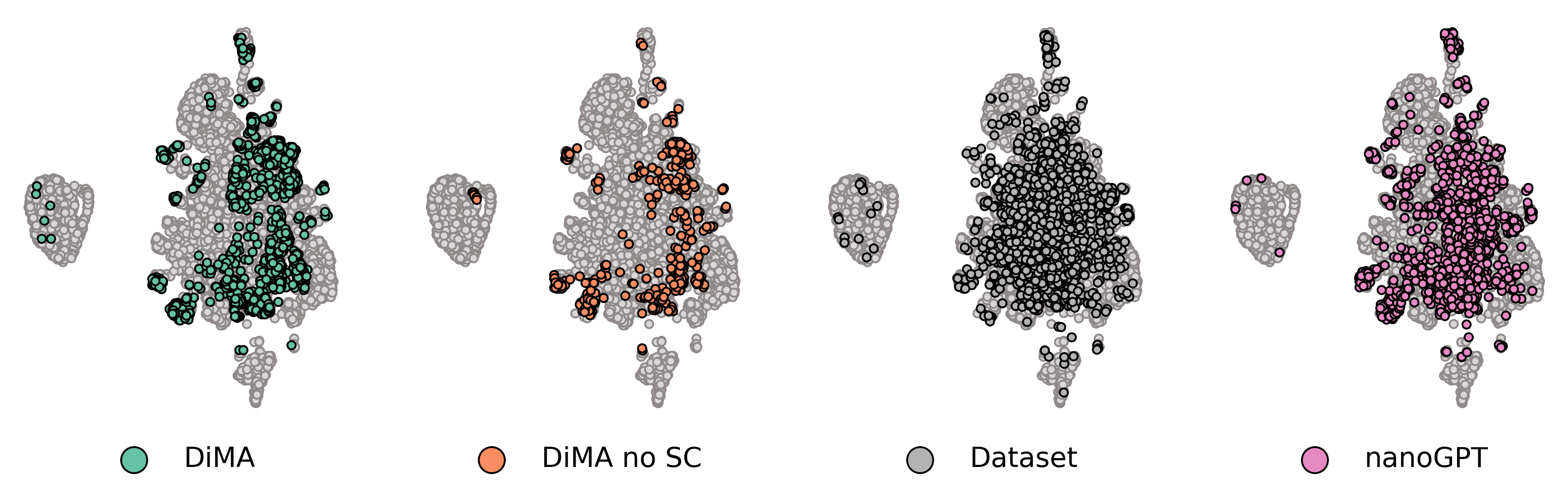}
\caption{UMAP projection of sequences from PC. Training dataset - AFDBv4-90. Grey background points - dataset sequences from PC. }
\label{fig:umap_afdb}
\end{figure}

\textbf{UMAP}.
To visually represent the distribution of generated sequences across PC, we trained UMAP on all sequences from PC for all models (parameters - n\_neighbors - 25 and min\_dict - 0.5).
The UMAP plots in Figures ~\ref{fig:umap_swissprot} and ~\ref{fig:umap_afdb} show that despite the fact that the diversity metrisc of the DIMA w/o self-conditioning are higher, DIMA with self-condition has the same coverage on the SwissProt (and even more coverage on AFDBv4-90). This and the fact that DIMA is closer to the dataset in terms of distribution learning metrics shows that DIMA w/o self condition achieved better diversity by generating sequences that greatly differ from those from the dataset. 

\subsection{Distribution similarity}
\label{appendix:metrics:dist_sim}

\textbf{Fréchet ProtT5 Distance (FD-seq) and Fréchet ProteinMPNN Distance (FD-struct).}
The Fréchet distance, also known as the 2-Wasserstein distance, quantifies the dissimilarity between two samples drawn from multivariate Gaussian distributions, denoted as \(X_1 \sim \mathcal{N}(\mu_1, \Sigma_1)\) and \(X_2 \sim \mathcal{N}(\mu_2, \Sigma_2)\), and is defined as follows:
\begin{equation}
    d(X_1, X_2)^2 = ||\mu_1 - \mu_2||^2 + \text{Tr}(\Sigma_1 + \Sigma_2 - 2\sqrt{\Sigma_1 \Sigma_2})
\end{equation}

\textbf{Maximum mean discrepancy (MMD).}
The idea behind MMD involves assessing the distance between two samples by measuring the difference in mean values resulting from applying a smooth function to the samples. A biased empirical estimate of MMD between two samples $X = \{x_1,...,x_n\}$ and $Y = \{y_1,...,y_n\}$ using kernel $k$ is defined as follows:
\begin{align}
{}&MMD_{k}^2(X, Y) = \frac{1}{n^2}\sum_{i=1}^{n}\sum_{j=1}^{n}\left ( k({x}_{i}, {x}_{j}) +
            k({y}_{i}, {y}_{j}) - 2k({x}_{i}, {y}_{j})\right )
\end{align}

As a kernel function, we used the radial basis function kernel (RBF). We evaluated the distance between batches of sequences, each of size $n$ equal to 2048, sampled from the dataset and generated by the respective models. Following the methodology proposed for 3D structures in \citep{Evaluation_Metrics3d}, we utilized ProtT5 sequence representations to calculate MMD.

\textbf{1-Wasserstein optimal transport (OT)}.
The BLAST identity metric effectively evaluates the similarity between generated and natural sequences. However, its limitation lies in assessing the model's capability to produce diverse sequences, as it may identify the same dataset sequence as the closest match for every generated sequence. To overcome this limitation, we employ transportation theory to establish optimal pairs between generated sequences and the dataset.

Optimal transport theory, initially devised for solving economic problems, has found applications in various fields, including physics, biology, and tomography. To implement our approach, we calculate pairwise Levenshtein distances and use them as transportation costs. Subsequently, we determine optimal sequence pairs using the Earth Mover Distance (EMD) solver with a uniform distribution of the samples. We use the average distance between these optimal pairs, measuring both the diversity and proximity of generated samples to the dataset.

\begin{figure} 
    \centering
    \includegraphics[width=\linewidth]{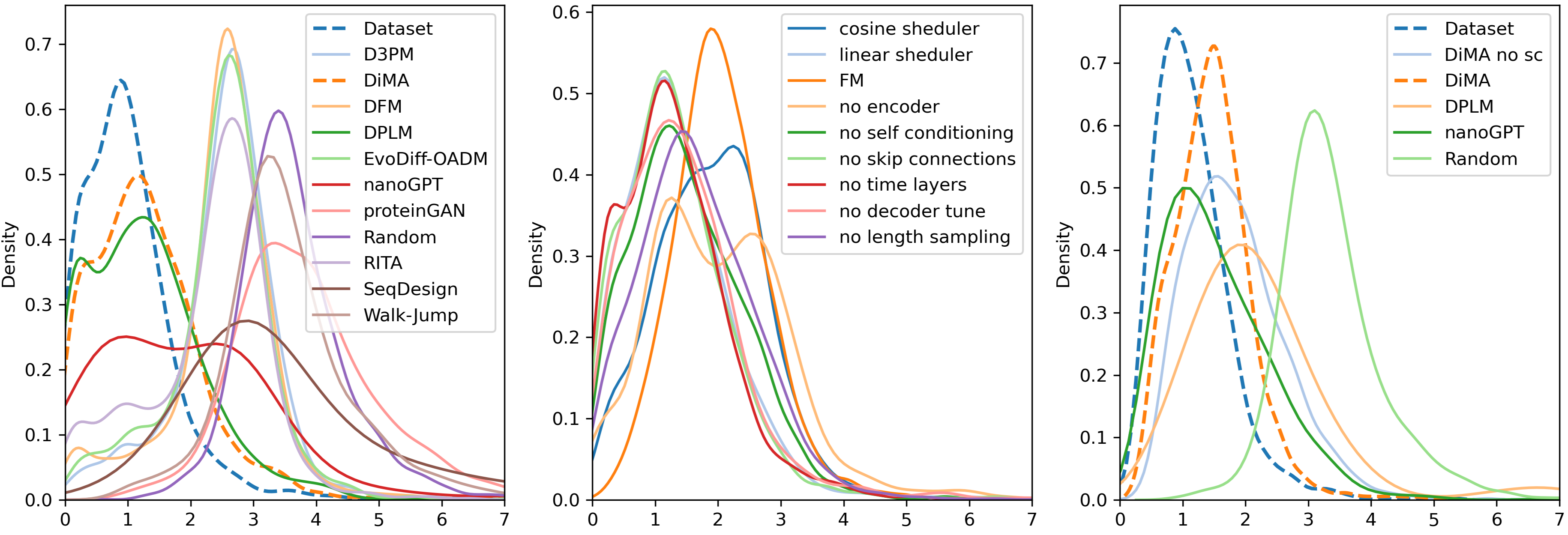}
    \caption{
    The distribution of optimal transport distances between pairs of generated and dataset sequences. For each model, we compute pairwise Levenshtein distances between generated sequences and dataset sequences, then find optimal matching pairs using Earth Mover Distance with uniform distribution of samples. \textbf{Left}: Comparison of DiMA against baseline models on SwissProt dataset. \textbf{Center}: Analysis of DiMA's architectural components through ablation studies. \textbf{Right}: Performance comparison on the larger AFDBv4-90 dataset. The dashed blue line represents the reference distribution obtained by matching samples within the dataset (optimal transport distance to itself), while the dashed orange line shows DiMA's distribution.}
\label{fig:fig_OT_swprt}
\end{figure}

The inherent diversity of the dataset, i.e., when a sample from the dataset pairs with itself, gives zero distances $(OT(dataset) = 0)$. In contrast, random sequences form optimal pairs with the highest mean distances, as illustrated in Figure \ref{fig:fig_OT_swprt}. The optimal transport distance distributions reveal differences in how models capture the protein sequence space. Most ablation studies (Figure \ref{fig:fig_OT_swprt}, center) show distributions similar to the reference, except for flow matching, cosine scheduler, and no encoder variants, indicating these components are most critical for DiMA's performance. Several baseline models (D3PM, DFM, EvoDiff-OADM, RITA) cluster around a similar mode between random and reference distributions, suggesting they mainly learn basic patterns like amino acid frequencies while capturing only a limited set of protein families, as evidenced by their left tail behavior (Figure \ref{fig:fig_OT_swprt}, left). 

DiMA's distribution (dashed orange) most closely matches the dataset reference (dashed blue) across both datasets. On SwissProt, DPLM shows a sharp, concentrated peak indicating high-quality but limited diversity, while other baselines show broader, right-shifted distributions indicating greater deviation from natural sequences. On the larger AFDBv4-90 dataset, while nanoGPT's distribution mode is closer to the reference, DiMA generates fewer distant proteins (smaller right tail) and better maintains the overall distribution shape, demonstrating robust performance even with increased dataset complexity (Figure \ref{fig:fig_OT_swprt} right).

Although our OT implementation offers advantages over BLAST, it has a special feature: the EMD solver identifies an exact pair for each sequence. This poses a challenge when dealing with two query sequences that are similar to one dataset sequence but distant from others, resulting in one close pair and one distant pair. However, we employ EMD precisely to penalize such cases, reinforcing the generation of diverse rather than similar sequences.

\textbf{Structural analogues}.
To measure structural distribution similarity, we calculate analogous FD, MMD, OT metrics using structural encoder ProteinMPNN.
ProteinMPNN is a powerful graph neural network (GNN) model pretrained on a massive dataset of protein structures.

\subsection{Novelty}
\label{appendix:metrics:novelty}

To directly evaluate the potential memorization of the training data, we measure \textbf{novelty} by calculating the mean sequence identity between each generated sequence and its nearest neighbor in the training dataset.

We assume that novel proteins should be far from the train dataset, so for each generated sequence, we computed distance to the nearest train sequence. The golden standard for pairwise distance measure between amino acid sequences is an alignment score using  Needleman–Wunsch (NW) algorithm. However, due to $O(N^2)$ calculation cost we use BLAST to find the nearest sequence in the training set and only then we align these sequences using NW. (We employ BLAST and NW with the following parameters: evalue = 15.05, matrix = BLOSUM62, word\_size= 2;  matrix= BLOSUM62, gap\_open= -10, gap\_extend\=-0.5). 
 The novelty value of a batch of generated sequences is defined as  \( \textbf{Novelty} (y) = {\frac {1}{s}}\sum _{i=1}^{s} 1- \frac{|\text{\# of same letters in alignment}|}{\text{alignment length}} \), where $y$ is a set of generated proteins $s$

\newpage
\section{Model Details}



\subsection{Model Architecture}
\label{appendix:architecture}

\begin{wrapfigure}{R}{0.5\textwidth}
    \centering
    \includegraphics[width=\linewidth]{"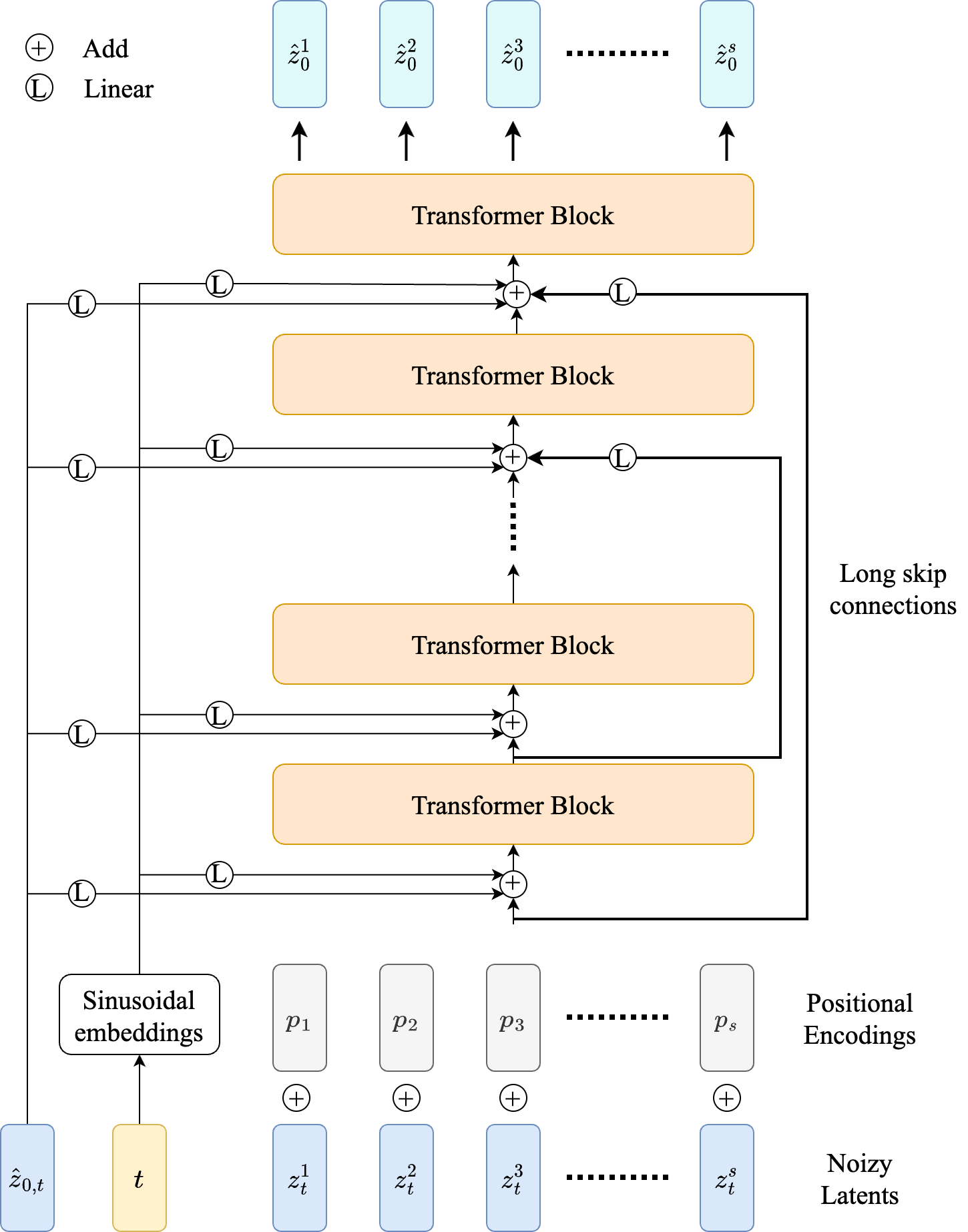"}
    \caption{The architecture of the denoising model.}
    \label{architecture}
    \vspace{-10mm}
\end{wrapfigure}

We employ a 12-layer Transformer model with 16 attention heads and a hidden size of 320 as the backbone for our diffusion model, incorporating several modifications specifically designed to optimize the denoising diffusion process in the context of protein-related data.
To enhance the model's performance, we first introduce trainable positional encodings to the noisy protein latents, allowing the model to better capture the sequential nature of the data. The input for each transformer block is constructed as a sum of the output from the previous block, time embeddings, and self-condition predictions, which are projected through linear layers. This approach facilitates the integration of temporal information and improves the model's ability to learn complex patterns.
Additionally, we implement long skip connections, recognizing that for time steps close to zero, the model's output closely resembles the input. This modification is crucial as it aids in learning an identity transformation, thereby stabilizing the training process and enhancing the model's overall efficacy.
The architecture of our model is illustrated in Figure~\ref{architecture}.

\subsection{Training Details}
\label{appendix:resources}

All models were trained with a batch size of $512$ and a learning rate of $1e^{-4}$ to convergence. 
We clip our gradient norm to $2$ and have a linear warmup schedule for the first $5000$ iterations.
We also use a 0.9999 EMA.

The experiments were conducted using $4$ A100 80GB GPUs.
Each training session lasts approximately 10 days

\subsection{Length Determination}
\label{appendix:length}

\begin{figure}[h]
    \centering
    \includegraphics[width=\linewidth]{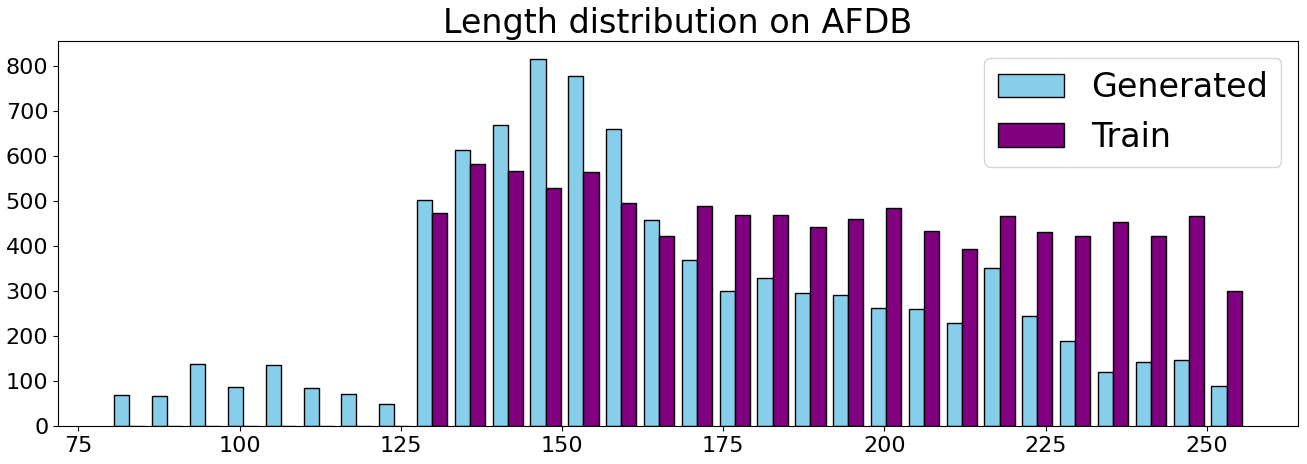}
    \caption{
    The distribution of lengths in the training and generated sets for models trained on AFDB.}
\label{fig:length-hists}
\end{figure}

During inference, the model must determine the length of the generated sequence. We explore two strategies to address this challenge: training diffusion models with and without padding masks.

\begin{itemize}
\item Diffusion with padding masks: During training, we provide the attention mask for pad tokens along with the corrupted latents and exclude pad tokens when computing the diffusion loss. At inference time, the sequence length is sampled from the empirical length distribution observed in the training set.
\item Diffusion without padding masks: In this setting, no explicit information about padding is provided during training, and the loss is computed over the entire sequence, including pad tokens. Consequently, during inference, the model must implicitly infer the appropriate sequence length without external guidance.
\end{itemize}

Figure~\ref{fig:length-hists} compares the length distributions of sequences from the training set and those generated using the second approach. Notably, the generated sequences exhibit a distribution shift, deviating from the expected length profile of the training data.

To mitigate this mismatch, we adopt the attention-masked approach, incorporating explicit length information during training and sampling lengths during inference to ensure closer alignment with the training distribution.

\subsection{Parameters for Navigating the Quality-Diversity Trade-off}
\label{appendix:knobs}

\begin{figure}[h]
    \centering
    \begin{minipage}[t]{0.49\textwidth}
        \centering
        \includegraphics[width=\linewidth]{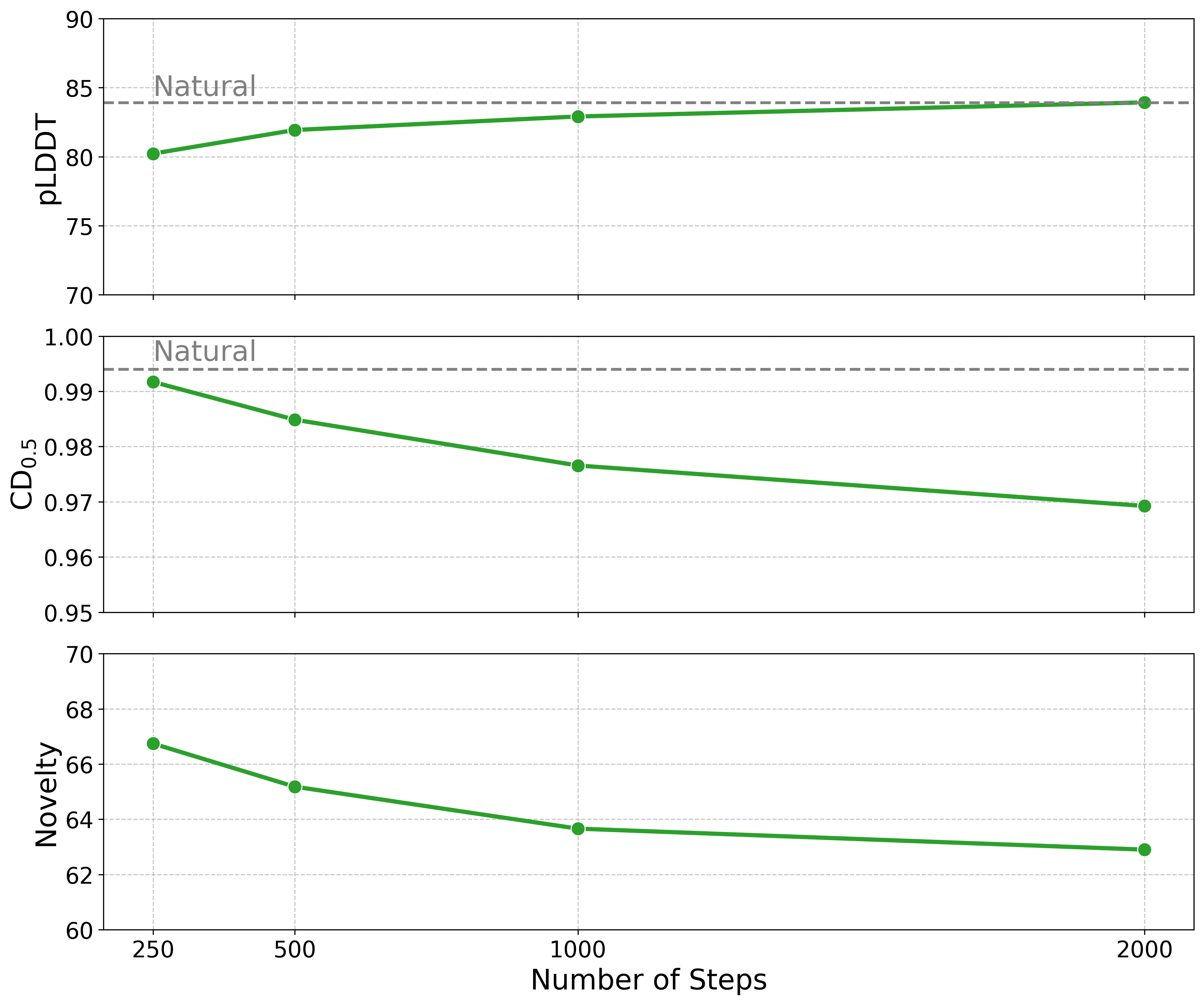}
        \caption{Effect of the number of sampling steps on DiMA-35M generation quality. Increasing the number of diffusion steps from 250 to 2000 shows a trade-off between quality and diversity. \textbf{Top}: structural quality (pLDDT) improves with more sampling steps, approaching natural protein level (gray dashed line). \textbf{Middle}: sequence diversity ($CD_{0.5}$) slightly decreases with more steps but remains consistently high ($>$0.97). \textbf{Bottom}: novelty decreases with increased sampling steps, indicating convergence toward more conservative protein designs. Each point represents evaluation on 2048 protein sequences generated with the specified number of steps.}
        \label{fig:sample_steps}
    \end{minipage}
    \hfill
    \begin{minipage}[t]{0.49\textwidth}
        \centering
        \includegraphics[width=\linewidth]{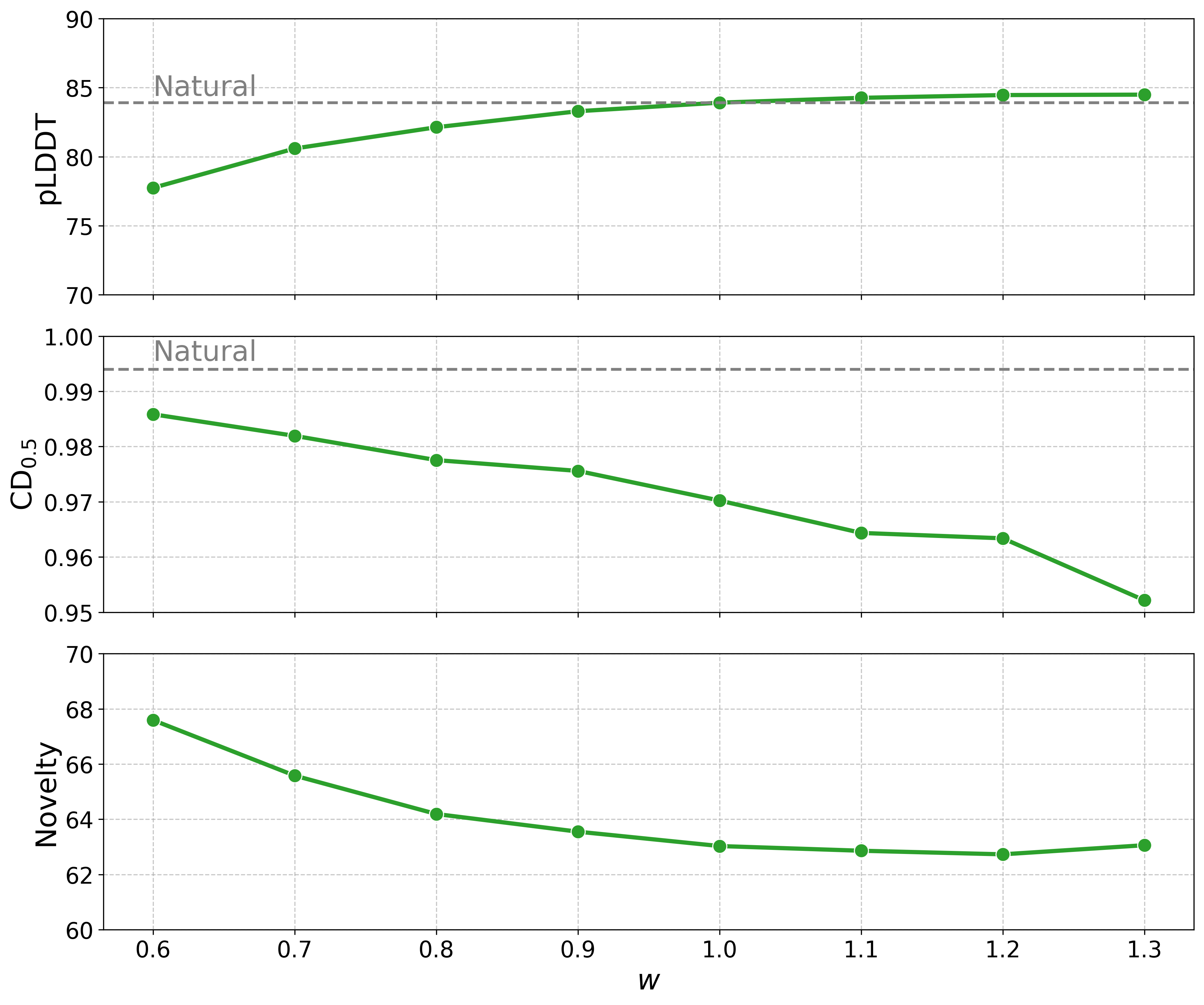}
        \caption{Effect of self-conditioning rate parameter ($w$) on DiMA-35M performance metrics. In self-conditioned generation $x_0 = \hat{x}_0 + w * (x_0 - \hat{x}_0)$, the parameter $w$ controls how strongly the model relies on its own previous predictions during the diffusion process. \textbf{Top}: structural quality (pLDDT) increases with higher $w$ values, approaching natural protein levels (gray dashed line) at $w\geq$~0.9. \textbf{Middle}: sequence diversity ($CD_{0.5}$) decreases as $w$ increases, with noticeable reduction at $w>$~1.0. \textbf{Bottom}: novelty follows a similar trend, showing higher values at lower $w$. Each point represents evaluation on 2048 protein sequences generated with 2000 sampling steps and the specified $w$ value.}
        \label{fig:self-cond_rate}
    \end{minipage}
\end{figure}

\subsection{The Rationale Behind the tan-10 Schedule}
\label{appendix:schedule}

The key issue with standard noise schedules (linear, cosine) is that they corrupt data very gradually at small timesteps (Figure \ref{fig:schedulers}). This means the model spends significant training resources on nearly trivial denoising tasks where input and target are almost identical. Our approach ensures the difficulty of the denoising task increases steadily with each timestep. By designing a noise schedule where reconstruction loss grows linearly with diffusion time, we create learning problems of incrementally increasing difficulty, allowing the model to make consistent progress throughout training rather than facing an abrupt jump in task complexity.

\begin{figure}[h]
    \centering
    \begin{minipage}[t]{0.49\textwidth}
        \centering
        \includegraphics[width=\linewidth]{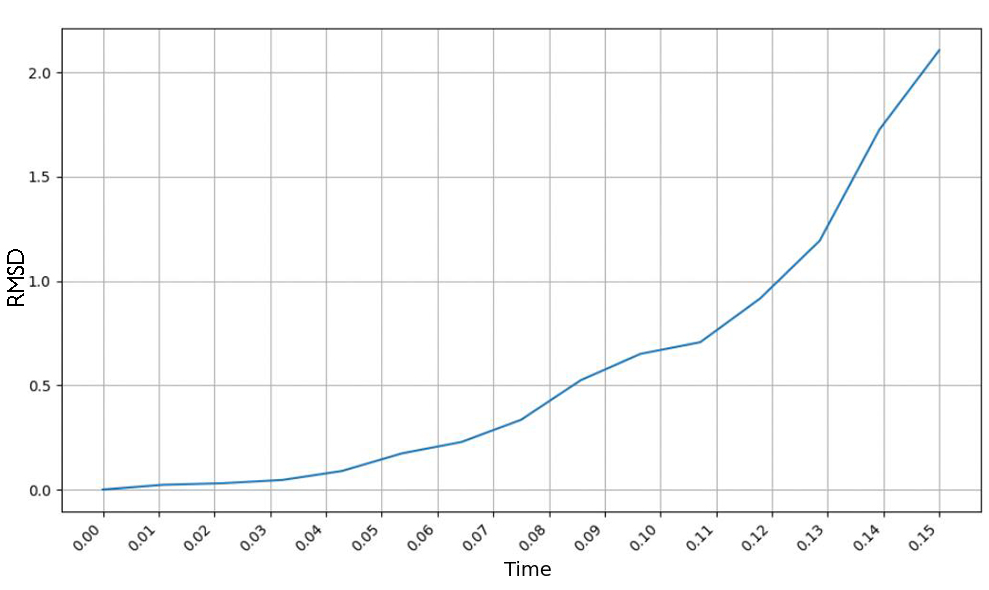}
        \caption{Structural stability of protein representations under noise. Root Mean Square Deviation (RMSD) between original protein structure and structure reconstructed from noisy CHEAP representations at different diffusion timesteps.}
        \label{fig:rmsd_vs_time}
    \end{minipage}
    \hfill
    \begin{minipage}[t]{0.49\textwidth}
        \centering
        \includegraphics[width=\linewidth]{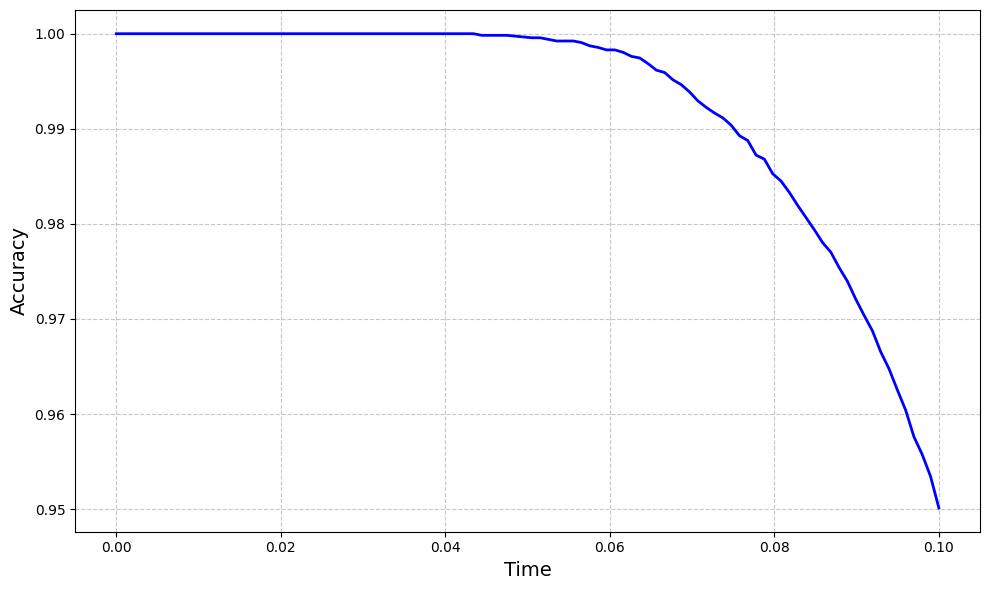}
        \caption{Sequence reconstruction accuracy under noise. Percentage of amino acids correctly reconstructed from noisy CHEAP representations at different diffusion timesteps.}
        \label{fig:accuracy_vs_time}
    \end{minipage}
\end{figure}

When working with protein encodings, reconstruction at small noise levels turns out to be quite robust. For example, testing DiMA with CHEAP representations (which enables dual-decoding into both sequence and structure) we observe that at $t=$~0.05, sequence reconstruction accuracy remains 100\%, and structural RMSD stays below 0.2$\AA$ (Figures \ref{fig:rmsd_vs_time} and \ref{fig:accuracy_vs_time}). This shows that unlike for example pixel-based image diffusion models, DiMA does not have to spend much effort at small noise steps to learn the fine-grained denoising capability. Our schedule leverages this robustness by allocating more training to challenging noise levels rather than the nearly lossless stages with $t\to$~0.

\newpage
\section{Related Work}

Diffusion generative models, introduced by \citet{sohl2015deep}, have gained attention for their remarkable results in image \citep{ho2020denoising, song2020score, song2020denoising}, and speech generation \citep{chen2020wavegrad, popov2021grad}.
Due to their impressive generative quality, some studies have extended the application of diffusion models to the text domain. \citet{hoogeboom2021argmax} and \citet{austin2021structured} proposed multinomial diffusion for discrete data corruption. 
Subsequently, other works \citep{li2022diffusion, lin2023text, gulrajani2023likelihood, han2022ssd, strudel2022self, gao2022difformer, shabalin2025tencdm} adapted Gaussian diffusion to sequence learning by embedding discrete data into continuous space.
\citet{yuan2022seqdiffuseq} extended the text diffusion model to the sequence-to-sequence setting.
\citet{ye2023dinoiser} conducted a study on the discrepancy of the text embedding space, demonstrating that the diffusion task at small noise scales is trivial. 
\citet{zhang2023planner} implemented latent text diffusion inside a VAE with an autoregressive decoder. 
\citet{lovelace2022latent} utilized diffusion models to generate a fixed-length latent representation, mapped into a high-dimensional space with the reconstruction network before being fed into an autoregressive decoder to generate text.

In protein science, deep learning has emerged as a transformative tool. Pre-trained on extensive protein sequence datasets,  it provides representations widely employed in various tasks \citep{ProtT5, ESM-2, lu2024tokenized}.
Generative models for protein sequences, exemplified by recent advancements, enhance predictions of proteins with improved properties and functions \citep{gen_seq, gen_struct}.
Simultaneously, progress in the sequence-to-structure domain, as seen in models like AlphaFold \citep{AlphaFold} and ESMFold \citep{ESM-2}, enables the prediction of 3D protein conformation from amino acid sequences.
Models such as ProteinMPNN \citep{ProteinMPNN} or ESM-IF1 \citep{ESM-IF1} predict an amino acid sequence given a specific 3D structure, effectively reverse engineering the process.

In the realm of protein generation, a diverse array of autoregressive models has been developed, establishing a sophisticated baseline for subsequent model classes \citep{ProGen_Nature, protgpt2, SeqDesign, lv2024prollama, hesslow2022rita, SeqDesign}.
Beyond autoregressive approaches, both categorical and continuous diffusion methods have emerged as promising techniques for sequence generation \citep{EvoDiff, wang2024diffusion, lee_score-based_2023, zhang2023pro}. Additionally, three-dimensional diffusion models have been successfully utilized for the generation of protein structures \citep{RFDiffusion, FoldingDiff, AlQuraishi_diffusion, fu2024latent}. Notably, there are models that facilitate the simultaneous generation of both sequence and structure, providing a more integrated approach to protein design \citep{campbell2024generative, ingraham2023illuminating}.
Furthermore, the field has seen the introduction of energy-based models \citep{frey2023protein} and generative adversarial networks (GANs) \citep{ProteinGAN}, which offer alternative frameworks for protein generation.

\newpage
\section{Additional results}

\subsection{Component Analysis and Architectural Optimization on SwissProt Dataset}\label{appendix:ablation}

\begin{wrapfigure}{r}{0.5\textwidth}
    \centering
    \includegraphics[width=\linewidth]{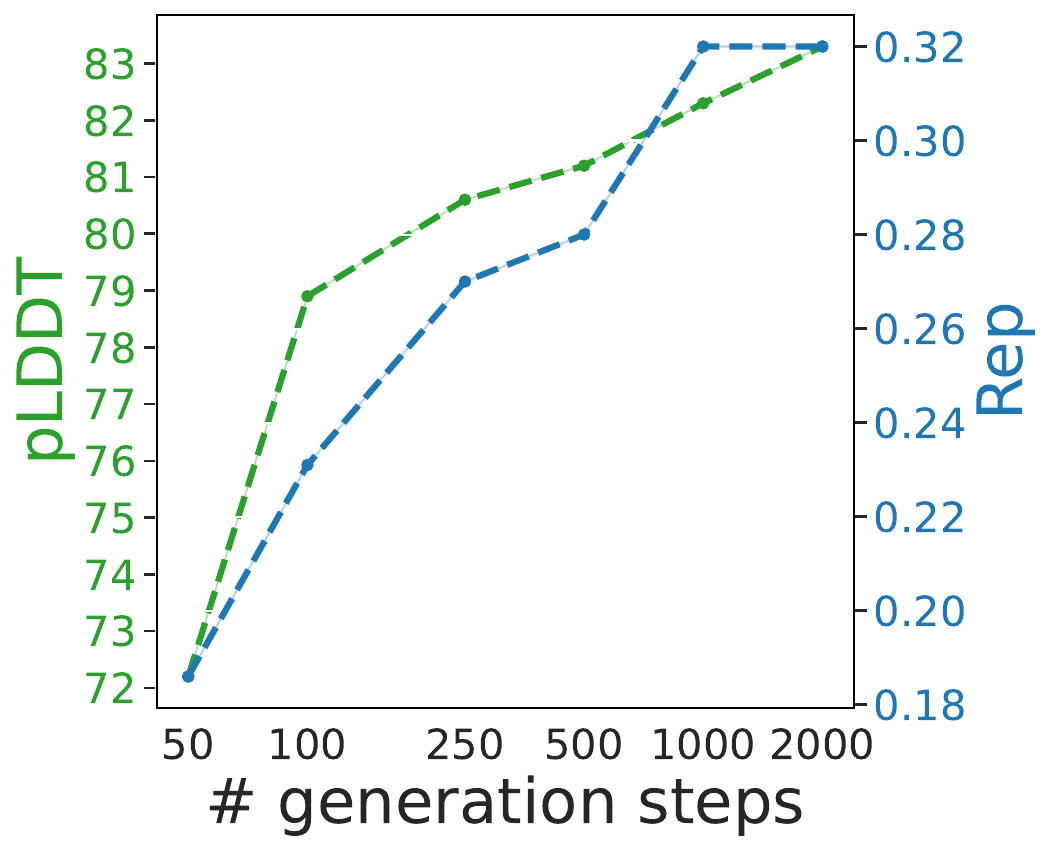}
    \caption{
    The dependence between the structural plausibility of generation, the degree of repetition, and the number of generation steps.}
\label{fig:generation_steps}
\end{wrapfigure}

In this section, we conduct a detailed ablation study to assess the impact of key architectural components in DiMA, trained on the SwissProt dataset with the lightweight ESM-8M encoder. Our goal is to evaluate how different design choices influence generation quality, distribution alignment, novelty, and diversity.

To achieve this, we compare multiple model variants across a comprehensive set of metrics, ensuring a rigorous analysis of their ability to generate high-quality, diverse, and novel proteins while maintaining fidelity to real protein distributions. 
All models were evaluated using 250 diffusion steps to ensure consistency in generation.
The results are summarized in Table \ref{tab:ablation:total}. 
This study ultimately aims to refine DiMA's capacity for robust and biologically meaningful protein generation.

The proposed model also establishes a trade-off between structural plausibility and diversity as illustrated in Figure \ref{fig:generation_steps}. 
Increasing the number of diffusion generation steps in protein production leads to higher protein quality; however, this improvement is accompanied by a slight decrease in protein diversity. This trade-off between quality and diversity provides flexibility during generation, allowing for the selection of proteins based on desired characteristics.

\begin{table}[p]
   \centering
   \small
   \vspace{-2.5mm}
   \caption{\sl Ablation study evaluating key components of DiMA trained on the SwissProt dataset using the ESM-8M encoder. The study examines the impact of architectural choices from multiple perspectives, employing a comprehensive set of metrics to assess generation quality, distribution alignment, novelty, and diversity of the produced proteins. All models were evaluated using 250 diffusion steps during generation.}
   \vspace{1.5pt}
   \label{tab:ablation:total}

  \resizebox{\linewidth}{!}{ %
   \begin{tabular}{llcccccc}
   \toprule
   & \bf Model 
   & \bf{pLDDT} ($\uparrow$) 
   & \bf{Progen ppl} ($\downarrow$) 
   & \bf{ESM-2 pppl} ($\downarrow$) 
   & \bf{scPpl} ($\downarrow$)
   & \bf{TM-score} ($\uparrow$)
   & \bf{BLAST} ($\uparrow$)
   \\
   \cmidrule[1pt](lr){2-8}

    \multirow{14}{*}{\rotatebox[origin=c]{90}{\textbf{Quality}}} 

   & Dataset                                & 80.7               & 6.03               & 5.35               & 1.88               & 0.80              & 100 \\
   & Random sequences                       & 24.8               & 21.91              & 21.53              & 2.77               & 0.33              & 0   \\
   \cmidrule[0.5pt](lr){2-8}
   & \chl DiMA                              & \chl \textbf{83.3} & \chl \textbf{5.07} & \chl \textbf{4.68} & \chl \textbf{1.17} & \chl \textbf{0.87} & \chl \textbf{68} \\
   & \hspace{3pt} w/o skip connections      & 77.3               & 6.79               & 5.84               & 1.87               & 0.82               & 61 \\
   & \hspace{3pt} w/o time layers           & 79.4               & 6.42               & 5.49               & 1.83               & 0.85               & 66 \\
   & \hspace{3pt} w/o ESM encoder           & 62.7               & 10.42              & 9.22               & 2.09               & 0.71               & 48 \\
   & \hspace{3pt} w/o self-conditioning     & 68.2               & 10.45              & 9.18               & 2.08               & 0.74               & 46 \\
   & \hspace{3pt} w/o finetuned decoder	    & 80.1	             & 6.66	              & 5.59	           & 1.78               & 0.85	             & 65 \\
   & \hspace{3pt} w/o length sampling       & 65.0              & 11.36              & 9.84               & 2.12               & 0.72                 & 44\\
   & \hspace{3pt} w linear schedule         & 77.0               & 7.66               & 6.29               & 1.89               & 0.82               & 58 \\
   & \hspace{3pt} w cosine schedule         & 54.1               & 13.11               & 10.86              & 2.16               & 0.60               & 34 \\
   & \hspace{3pt} w flow-matching	        & 63.4	             & 11.44              &	8.97	           & 2.08	            & 0.68	             & 40 \\

   \toprule
   \vspace{2\baselineskip}\\
   &  
   & \bf{FD-seq} ($\downarrow$) 
   & \bf{MMD-seq} ($\downarrow$) 
   & \bf{OT-seq} ($\downarrow$) 
   & \bf{FD-struct} ($\downarrow$)
   & \bf{MMD-struct} ($\downarrow$)
   & \bf{OT-struct} ($\downarrow$)
   \\
   \cmidrule[1pt](lr){2-8}
    \multirow{14}{*}{\rotatebox[origin=c]{90}{\textbf{Distributional Similarity}}} 

   & Dataset                                & 0.13               & 0.000               & 1.08               & 0.000               & 0.000              & 0.053 \\
   & Random sequences                       & 3.97               & 0.200               & 3.88               & 1.231               & 0.412              & 1.313 \\
   \cmidrule[0.5pt](lr){2-8}
   & \chl DiMA                              & \chl \textbf{0.34} & \chl \textbf{0.020}& \chl \textbf{1.26}     & \chl 0.030     & \chl 0.004     & \chl 0.090 \\
   & \hspace{3pt} w/o skip connections      & 0.45               & 0.021              & 1.36                   & \textbf{0.029}               & \textbf{0.002}              & \textbf{0.081}\\
   & \hspace{3pt} w/o time layers           & 0.41               & 0.022              & 1.29                   & 0.035              & 0.004              & 0.097\\
   & \hspace{3pt} w/o ESM encoder           & 1.07               & 0.068              & 2.04                   & 0.069              & 0.010              & 0.153\\
   & \hspace{3pt} w/o self-conditioning     & 0.55               & 0.047              & 1.51                   & 0.031              & 0.005              & 0.093\\
   & \hspace{3pt} w/o finetuned decoder	    & 0.54               & 0.031              & 1.44                   & 0.042              & 0.004              & 0.108\\
   & \hspace{3pt} w/o length sampling       & 0.67               & 0.058              & 1.67                   & 0.048              & 0.007              & 0.139\\
   & \hspace{3pt} w linear schedule         & 0.47               & 0.026              & 1.37                   & 0.031              & 0.003              & 0.092\\
   & \hspace{3pt} w cosine schedule         & 0.94               & 0.091              & 1.90                   & 0.122              & 0.019              & 0.215\\
   & \hspace{3pt} w flow-matching	        & 0.71               & 0.063              & 1.75                   & 0.049              & 0.008              & 0.130\\

    \toprule
    \vspace{2\baselineskip}\\
   & 
   & \bf{Rep} ($\downarrow$) 
   & \bf{CD$_{0.5}$} ($\uparrow$) 
   & \bf{CD$_{0.95}$} ($\uparrow$) 
   & \bf{PCD$_{0.5}$}
   & \bf{NCD$_{0.5}$}
   & \bf{Novelty} ($\uparrow$) 
   \\
   \cmidrule[0.5pt](lr){2-8}
    \multirow{14}{*}{\rotatebox[origin=c]{90}{\textbf{Diversity and Novelty}}} 

   & Dataset                                & 0.045              & 1.000                  & 0.943              & 0.990              & 0.304    & 25.3           \\
   & Random sequences                       & 0.000              & 1.000                  & 1.000                & 1.000                & 0.000 & 85.1                \\
   \cmidrule[0.5pt](lr){2-8}
   & \chl DiMA                              & \chl 0.320& \chl 0.611 & \chl 0.992& \chl 0.246& \chl 0.392  & \chl 35.7\\
   & \hspace{3pt} w/o skip connections      & 0.274              & 0.619              & 0.990              & 0.187                & 0.439 & 43.1 \\
   & \hspace{3pt} w/o time layers           & 0.256              & 0.550              & 1.000                & 0.246              & 0.347 & 38.4\\
   & \hspace{3pt} w/o ESM encoder           & 0.346              & 0.619              & 1.000                & 0.107              & 0.507 & 50.0\\
   & \hspace{3pt} w/o self-conditioning     & 0.043              & 0.929              & 1.000                & 0.146              & 0.779 & 56.6\\
   & \hspace{3pt} w/o finetuned decoder	    & 0.266              & 0.589              & 0.996              & \textbf{0.255}       & 0.357 & 38.4\\
   & \hspace{3pt} w/o length sampling       & 0.050              & 0.880              & 1.000                & 0.089              & 0.726 & 58.4\\
   & \hspace{3pt} w linear schedule         & 0.208              & 0.611              & 1.000                & 0.181              & 0.431 & 45.8\\
   & \hspace{3pt} w cosine schedule         & 0.046              & 0.878              & 1.000                & 0.017              & 0.798 & 67.0\\
   & \hspace{3pt} w flow-matching	        & \textbf{0.041}     & \textbf{0.960}        & 1.000                & 0.214              & 0.945 & \textbf{72.2}\\
  
   \bottomrule

   \end{tabular}
   }
\end{table}

\subsection{Comparison Across Generative Paradigms on SwissProt Dataset}
\label{appendix:comparison}

In this section, we present a comprehensive evaluation of diverse generative paradigms for protein sequence design, including latent Gaussian diffusion (DiMA), autoregressive models, score-based models, GANs, and discrete diffusion models. All models are trained on the SwissProt dataset to ensure consistency and comparability.

\begin{figure}[p]
    \centering
    \includegraphics[width=0.7\linewidth]{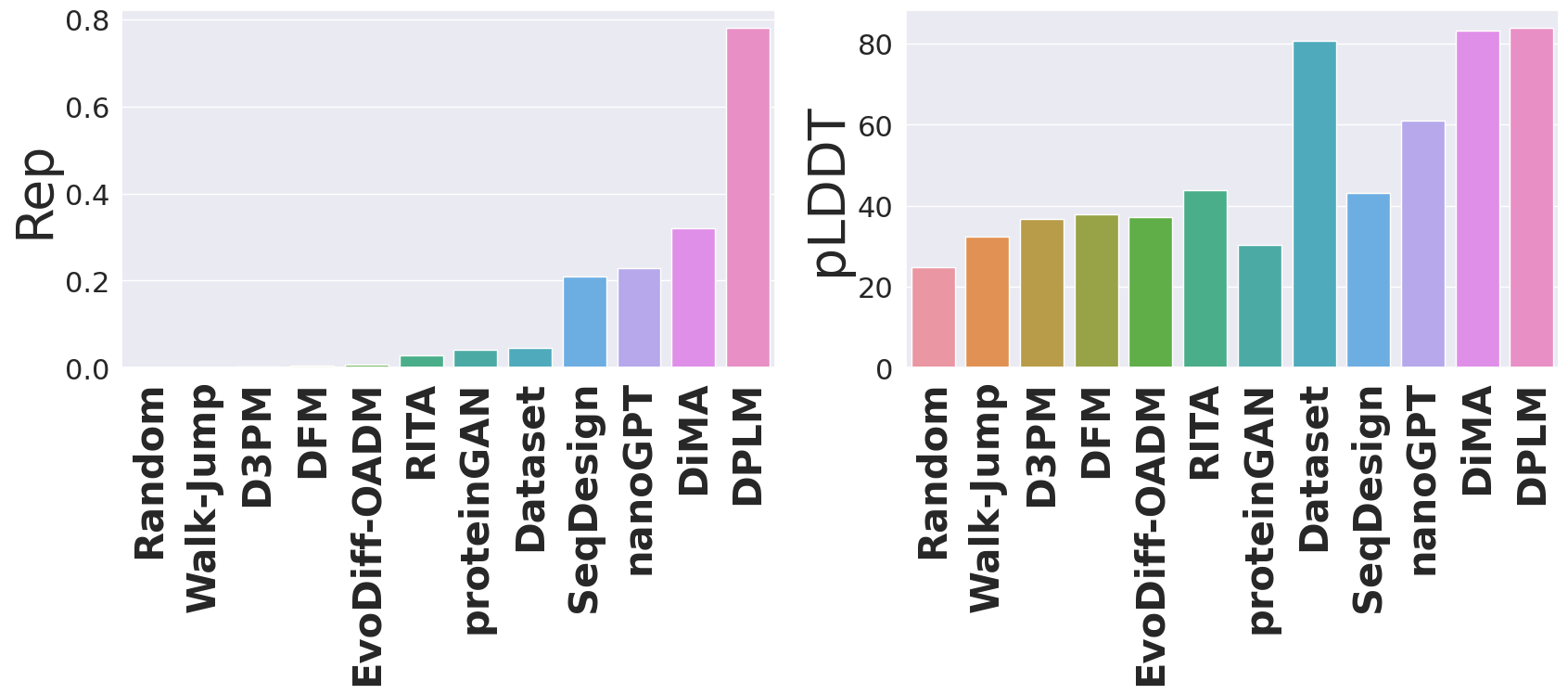}
    \caption{ \small
    Comparison of Rep (diversity) and pLDDT (structural quality) values for different protein generation models trained on the SwissProt dataset.}
\label{fig:comparison}
\end{figure}

Our primary objective is to investigate how these generative approaches impact key aspects of protein generation, such as quality, distribution alignment, novelty, and diversity. By analyzing these dimensions, we aim to uncover the strengths and limitations of each paradigm in capturing the complexity of protein sequences.

To achieve this, we employ a robust set of evaluation metrics, enabling a rigorous and multidimensional comparison of the models' performance.
The results of this evaluation are summarized in Table \ref{tab:comparison:total}.

\begin{table}[p]
   \centering
   \vspace{-2.5mm}
   \caption{\sl Performance comparison of different generative paradigms for protein sequence generation, each trained from scratch with approximately 35M parameters on the SwissProt dataset. The comparison assesses generation quality, distribution alignment, novelty, and diversity.}
   \vspace{1.5pt}
   \label{tab:comparison:total}

  \resizebox{\linewidth}{!}{ %
   \begin{tabular}{llcccccc}
   \toprule
   & \bf Model 
   & \bf{pLDDT} ($\uparrow$) 
   & \bf{Progen ppl} ($\downarrow$) 
   & \bf{ESM-2 pppl} ($\downarrow$) 
   & \bf{scPpl} ($\downarrow$)
   & \bf{TM-score} ($\uparrow$)
   & \bf{BLAST} ($\uparrow$)
   \\
   \cmidrule[1pt](lr){2-8}

    \multirow{13}{*}{\rotatebox[origin=c]{90}{\textbf{Quality}}} 

   & Dataset                                & 80.7               & 6.03               & 5.35               & 1.88               & 0.80              & 100 \\
   & Random sequences                       & 24.8               & 21.91              & 21.53              & 2.77               & 0.33              & 0   \\
   \cmidrule[0.5pt](lr){2-8}
   & Walk-Jump                           & 32.4               & 15.47              & 14.72               & 2.41              & 0.35               & 1  \\
   & RITA                                & 43.9               & 14.99              & 13.77              & 2.36               & 0.48               & 28  \\
   & proteinGAN                          & 30.4               & 17.58              & 16.48              & 2.57               & 0.00               & 0\\
   & SeqDesign                           & 43.1               & 12.78              & 11.89              & 2.35               & 0.41               & 17  \\
   & EvoDiff-OADM                        & 37.1               & 16.42              & 15.77              & 2.44               & 0.42               & 12  \\
   & D3PM                                & 36.7               & 16.83              & 16.52              & 2.36               & 0.48               & 9   \\
   & DFM                                 & 37.8               & 16.48              & 15.25              & 2.44               & 0.40               & 9   \\
   & DPLM                                & \textbf{84.0}                & \textbf{3.57}               & \textbf{3.50}               & 1.68              & \textbf{0.93}               & \textbf{88}  \\
   & nanoGPT                             & 61.0                & 8.87               & 8.18               & 2.04              & 0.63               & 43  \\
   & \chl DiMA                         	   & \chl 83.3                & \chl 5.07               & \chl 4.68              & \chl \textbf{1.17}               & \chl  0.87               & \chl 68  \\

   \toprule
   \vspace{2\baselineskip}\\
   &  
   & \bf{FD-seq} ($\downarrow$) 
   & \bf{MMD-seq} ($\downarrow$) 
   & \bf{OT-seq} ($\downarrow$) 
   & \bf{FD-struct} ($\downarrow$)
   & \bf{MMD-struct} ($\downarrow$)
   & \bf{OT-struct} ($\downarrow$)
   \\
   \cmidrule[1pt](lr){2-8}
    \multirow{13}{*}{\rotatebox[origin=c]{90}{\textbf{Distributional Similarity}}} 

   & Dataset                                 & 0.13               & 0.00               & 1.08               & 0.00               & 0.00              & 0.05 \\
   & Random sequences                       & 3.97               & 0.20               & 3.88               & 1.23               & 0.41              & 1.31 \\
   \cmidrule[0.5pt](lr){2-8}
   & Walk-Jump                              & 2.63               & 0.33              & 3.56                & 0.61               & 0.05               & 0.69     \\
   & RITA                                   & 1.19               & 0.14              & 2.28                & 0.37               & 0.03               & 0.52     \\
   & proteinGAN                             & 2.94               & 0.17              & 3.98               & 0.93               & 0.34               & 1.02     \\
   & SeqDesign                              & 3.53               & 0.19              & 5.12               & 0.95               & 0.25               & 1.11     \\
   & EvoDiff-OADM                           & 1.49               & 0.11              & 2.63               & 0.52               & 0.20               & 0.66     \\
   & D3PM                                   & 1.50               & 0.19              & 2.56                  & 0.57               & 0.05               & 0.72     \\
   & DFM                                    & 1.46               & 0.19              & 2.49                  & 0.52               & 0.04               & 0.68     \\
   & DPLM                                   & 0.50               & \textbf{0.02}              & 3.50               & 1.68               & 0.93               & 0.88  \\
   & nanoGPT                                & 1.24               & 0.06              & 2.53               & 0.15               & 0.04               & 0.26     \\
   & \chl DiMA              	                  & \chl \textbf{0.34}               & \chl \textbf{0.02}              & \chl \textbf{1.26}               & \chl \textbf{0.03}               & \chl \textbf{0.01}             & \chl \textbf{0.09}     \\

    \toprule
    \vspace{2\baselineskip}\\
   & 
   & \bf{Rep} ($\downarrow$) 
   & \bf{CD$_{0.5}$} ($\uparrow$) 
   & \bf{CD$_{0.95}$} ($\uparrow$) 
   & \bf{PCD$_{0.5}$}
   & \bf{NCD$_{0.5}$}
   & \bf{Novelty} ($\uparrow$)
   \\
   \cmidrule[1pt](lr){2-8}
    \multirow{13}{*}{\rotatebox[origin=c]{90}{\textbf{Diversity and Novelty}}} 

   & Dataset                                & 0.045              & 1.000                  & 0.943              & 0.990              & 0.304  & 25.3             \\
   & Random sequences                       & 0.000              & 1.000                  & 1.000                & 1.000                & 0.000   & 85.1             \\
   \cmidrule[0.5pt](lr){2-8}
   & Walk-Jump                              & \textbf{0.001}              & \textbf{1.000}                & 1.000                & 0.000                & 1.000     & 82.2             \\
   & RITA                                   & 0.028              & 0.988              & 0.998              & 0.125              & 0.861   & 60.4             \\
   & proteinGAN                             & 0.042              & 0.955              & 1.000               & 0.000                & 0.955   & \textbf{83.5}             \\
   & SeqDesign                              & 0.210              & 0.929              & 1.000               & 0.009              & 0.929   & 81.2             \\
   & EvoDiff-OADM                           & 0.006              & 0.986              & 1.000               & 0.058              & 0.929   & 77.6             \\
   & D3PM                              	  & 0.003              & 0.994              & 1.000                & 0.025              & 0.968  & 78.4              \\
   & DFM                                    & 0.004              & 0.996              & 1.000                & 0.048              & 0.947     & 77.2           \\
   & DPLM                                   & 0.781              & 0.494              & 0.812              & 0.267              & 0.236     & 11.5           \\
   & nanoGPT                                & 0.228              & 0.900              & 0.994              & 0.226              & 0.679     & 53.7           \\
   & \chl DiMA                         	        & \chl 0.320              & \chl  0.611              & \chl 0.992              & \chl 0.246              & \chl 0.392      & \chl 35.7          \\
   \bottomrule

   \end{tabular}
   }
\end{table}

\subsection{Representation Space Scaling}
\label{appendix:encoders}

In this section, we explore the impact of encoder choice on the performance of latent diffusion models for protein sequence generation. All models in this analysis are configured with approximately 35 million parameters and are trained on the AFDBv4-90 dataset to ensure consistency and comparability.
The results of this analysis, including detailed performance metrics and comparisons, are presented in Table \ref{tab:encoders:total}, providing valuable insights into the role of encoder design in latent diffusion models.

\paragraph{CHEAP Architecture}

CHEAP~\citep{lu2024tokenized} is a hierarchical representation learning framework for proteins, composed of six key components:
\begin{enumerate}
    \item ESM-3B Encoder – Aggregates outputs from multiple transformer blocks into a sequence of 1024-dimensional embeddings, maintaining the length of the amino acid sequence.
    \item Normalizer – Performs per-component normalization, removing high-activation components to enhance stability.
    \item CHEAP Encoder – Compresses the sequence embeddings into a lower-dimensional latent space, reducing both dimensionality and sequence length by a predefined factor.
    \item CHEAP Decoder – Maps the CHEAP latent representations back to ESMFold embeddings.
    \item Sequence Decoder – Reconstructs the amino acid sequence from the ESMFold embeddings.
    \item Structure Decoder – Recovers the protein backbone from the ESMFold embeddings.
\end{enumerate}

Our diffusion model operates on CHEAP latents. In the CHEAP\_shorten\_1\_dim\_1024 configuration, the CHEAP encoder and decoder are omitted, meaning diffusion is trained directly on the normalized aggregated embeddings from ESM-3B. 

\begin{table}[p]
   \centering
   \small
   \caption{\sl Perfomance of protein sequence generation using DiMA and different encoders on AFDBv4-90. The comparison assesses generation quality, distribution alignment, novelty, and diversity from different perspectives.}
   \vspace{1.5pt}
   \label{tab:encoders:total}

  \begin{tabular}{llccccc}
   \toprule
   & \bf Encoder 
   & \bf{pLDDT} ($\uparrow$) 
   & \bf{Progen ppl} ($\downarrow$) 
   & \bf{ESM-2 pppl} ($\downarrow$) 
   & \bf{scPpl} ($\downarrow$)
   & \bf{BLAST} ($\uparrow$)
   \\
    \cmidrule[1pt](lr){2-7}
    \multirow{8}{*}{\rotatebox[origin=c]{90}{\textbf{Quality}}} 

    & ESM2-8M           & 74.3          & 10.71            & 8,19            & 1.89                 & \textbf{50} \\
    & ESM2-35M          & 75.7          & 10.16            & 8.00            & 1.95                 & 44 \\
    & ESM2-150M         & 80.1          & 9.92             & 7.01            & 1.86                 & 48 \\
    & ESM2-650M         & 82.5          & \textbf{9.00}    & 5.90            & 1.79                 & \textbf{50} \\
    & ESM2-3B           & \textbf{83.4} & \textbf{9.00}    & \textbf{5.80}   & \textbf{1.78} & \textbf{50} \\
    & ESMC-300M         & 82.7          & 9.41             & 6.05            & \textbf{1.78} & 49 \\
\cmidrule[0.5pt](lr){2-7}
    & CHEAP\_shorten\_1\_dim\_1024 & 81.92         & 9.25             & 6.92               &  1.81          & \textbf{50}  \\
    & CHEAP\_shorten\_2\_dim\_1024 & 78.81         & 9.77             & 7.81           & 1.86                  & 47 \\
\cmidrule[0.5pt](lr){2-7}
    & SaProt-35M        & 82.23         & 9.93             & 6.90           & 1.82                     & \textbf{50} \\
    & SaProt-650M       & 83.01         & 9.57             & 6.65           & 1.79                     & \textbf{50} \\

\toprule
\vspace{2\baselineskip}\\
   & 
   & \bf{FD-seq} ($\downarrow$) 
   & \bf{MMD-seq} ($\downarrow$) 
   & \bf{OT-seq} ($\downarrow$) 
    & \bf{FD-struct} ($\downarrow$) 
   & \bf{MMD-struct} ($\downarrow$) 
   \\
\cmidrule[1pt](lr){2-7}
\multirow{8}{*}{\rotatebox[origin=c]{90}{\parbox{1cm}{\textbf{Distributional} \\ \textbf{Similarity}}}}
    & ESM2-8M               & 0.560           & 0.0389          &   1.488          & 0.027 & 0.100 \\ 
    & ESM2-35M              & 0.340           & 0.0204          &   1.294          & 0.029 & 0.098 \\
    & ESM2-150M             & 0.323           & 0.0152          & \textbf{1.232}   & 0.018 & 0.081 \\ 
    & ESM2-650M             & 0.318           & 0.0151          &   1.247          & 0.018 & 0.081 \\
    & ESM2-3B             & \textbf{0.314}    & \textbf{0.0143} &   1.246          & 0.022 & 0.077 \\ 
    & ESMC-300M               & 0.326         & 0.0155          &   1.259          & 0.020 & 0.086 \\ 
\cmidrule[0.5pt](lr){2-7}                                                   
    & CHEAP\_shorten\_1\_dim\_1024     & 0.346           & 0.0178          &   1.289          & \textbf{0.013} & \textbf{0.067} \\ 
    & CHEAP\_shorten\_2\_dim\_1024     & 0.340           & 0.0201          &   1.328          & 0.014 & 0.092 \\ 
\cmidrule[0.5pt](lr){2-7}
    & SaProt-35M            & 0.366           & 0.0188          &   1.269          & 0.021 & 0.071 \\ 
    & SaProt-650M           & 0.411           & 0.0207          &   1.326         & 0.017 & 0.078 \\
\toprule
\vspace{2\baselineskip}\\
   & 
   & \bf{Rep} ($\downarrow$) 
   & \bf{CD$_{0.5}$} ($\uparrow$) 
   & \bf{CD$_{0.95}$} ($\uparrow$) 
   & \bf{Novelty} ($\uparrow$)
   \\
\cmidrule[1pt](lr){2-6}
\multirow{8}{*}{\rotatebox[origin=c]{90}{\textbf{Diversity and Novelty}}}
    & ESM2-8M           & 0.029            & 0.981             & 1.0          & 68.0     &   \\
    & ESM2-35M          & 0.019            & 0.986             & 1.0          & \textbf{69.1}     &    \\
    & ESM2-150M          & 0.015            & \textbf{0.988}   & 1.0          & 65.6     &   \\
    & ESM2-650M          & \textbf{0.01}    & 0.986             & 1.0          & 64.1     &  \\
    & ESM2-3B            & \textbf{0.01}    & 0.969             & 1.0          & 63.0     &   \\
    & ESMC-300M          & 0.012            & 0.863             & 1.0          & 64.2     &   \\
\cmidrule[0.5pt](lr){2-6}       
    & CHEAP\_shorten\_1\_dim\_1024  & 0.017            & 0.951             & 1.0          & 64.6     &  \\
    & CHEAP\_shorten\_2\_dim\_1024  & 0.012            & 0.946             & 1.0          & 66.2     &   \\
\cmidrule[0.5pt](lr){2-6}
    & SaProt-35M         &  0.018           & 0.976             & 1.0          & 65.5     &  \\
    & SaProt-650M        & \textbf{0.010}            & 0.980             & 1.0          & 65.7     &  \\
            
  \bottomrule
    \end{tabular}
\end{table}

\subsection{Comparison with Pre-trained Protein Models}
\label{appendix:pretrain}

In this section, we compare DiMA with existing large protein models, including 
RITA~\citep{hesslow2022rita}, 
ProtGPT2~\citep{protgpt2}, 
ProGen2~\citep{ProGen_Nature}, 
EvoDiff~\citep{EvoDiff}, 
ProLLAMA~\citep{lv2024prollama}, 
DPLM~\citep{wang2024diffusion}, 
Chroma~\citep{ingraham2023illuminating}, 
Multiflow~\citep{campbell2024generative}, 
RFDiffusion~\citep{RFDiffusion} in different configurations.
For all models, we adhere to the sampling parameters recommended by the authors. 
This experiment specifically focuses on methods that provide publicly accessible pre-trained weights, ensuring transparency and reproducibility in our evaluation.

\begin{table}[th!]
   \centering
   \caption{\sl Comparison of the DiMA model with established pre-trained large protein models.}
   \vspace{1.5pt}
   \label{tab:appendix:pretrain}

   \resizebox{\linewidth}{!}{ %
  \begin{tabular}{llccccccc}
   \toprule
   & \bf Encoder 
   & \bf{pLDDT} ($\uparrow$) 
   & \bf{Progen ppl} ($\downarrow$) 
   & \bf{ESM-2 pppl} ($\downarrow$) 
   & \bf{scPpl} ($\downarrow$)
   & \bf{Rep} ($\downarrow$)
   & \bf{CD$_{0.5}$} ($\uparrow$)
   & \bf{CD$_{0.95}$} ($\uparrow$)
   \\

\midrule                                             
    & Multiflow-21M             & 82.8               & 8.67               & 4.87              & \textbf{1.00}& 0.181   & 0.990      & 1.000  \\
    & Chroma-33M                & 66.8               & 12.09              & 7.64              & 1.55        & 0.022   & 1.000      & 1.000  \\
    & RFDiffusion-80M           & 76.7               & 12.07              & 8.05              & 1.25       & 0.018   & 1.000      & 1.000  \\
    & ProtGPT2-738M	            & 63.0               & 7.79               & 5.70              & 2.20        & 0.096   & 0.998    & 1.000  \\
    & ProGen2-151M              & 46.2               & 12.78              & 11.33             & 2.39        & 0.084   & 0.998    & 1.000  \\
    & ProGen2-764M              & 50.3               & 12.05              & 10.94             & 2.37       & 0.066   & 0.996    & 0.996\\
    & ProGen2-2.7B              & 52.3               & 11.78              & 10.57             & 2.35        & 0.044   & 0.992    & 0.994\\
    & ProGen2-6.4B              & 57.2               & 9.71               & 8.67              & 2.26        & 0.087   & 0.976    & 1.000  \\
    & EvoDiff-38M               & 40.2               & 17.46              & 15.61             & 2.53        & 0.005   & 1.000      & 1.000  \\
    & EvoDiff-640M              & 40.5               & 17.35              & 15.38             & 2.52        & 0.000   & 1.000      & 1.000  \\
    & ProLLAMA-7B               & 53.1               & 10.50              & 7.46              & 2.26        & 0.133   & 0.982    & 1.000 \\
    & RITA-85M                  & 40.3               & 18.34              & 16.16             & 2.55        & 0.000   & 1.000      & 1.000  \\
    & RITA-300M                 & 41.5               & 19.10              & 15.73             & 2.57        & 0.000   & 0.990    & 0.990\\
    & RITA-680M                 & 42.5               & 20.48              & 15.31             & 2.63        & 0.000   & 0.958    & 0.958\\
    & RITA-1.2B                 & 42.6               & 19.39              & 15.22             & 2.64        & 0.000   & 0.966    & 0.966\\
    & DPLM-150M                 & 81.8               & \textbf{3.90}      & 2.82              & 1.60         & 0.658   & 0.654    & 0.917\\
    & DPLM-650M                 & 81.8               & 4.36               & \textbf{2.41}     & 1.60         & 0.533   & 0.746    & 0.943\\
    & DPLM-3B                   & 83.1               & 4.16               & 2.75              & 1.57         & 0.911   & 0.568    & 0.732\\
    & PLAID-100M                   & 53.5               & 14.98               & 	13.46              & 2.29         & 0.001   & 1.000    & 1.000 \\
\midrule
    & \chl DiMA-35M             & \chl \textbf{83.4} & \chl 9.00          & \chl 5.80         & \chl 1.78    & \chl 0.010   & \chl 0.969    &\chl 1.000 \\
   \bottomrule
   \end{tabular}
   }

\end{table}

The majority of models were pre-trained on distinct versions of the UniProt \citep{uniprot2021uniprot} dataset.
As a result, the application of distributional similarity metrics in the current experiment is rendered unfeasible.
Consequently, we focused solely on evaluating quality and diversity metrics.
Given that RFDiffusion generates protein structures, we employed ProteinMPNN, a neural network trained to predict amino acid sequences from 3D protein structures, to infer sequences from the generated structures. 
The authors of RFDiffusion ran ProteinMPNN multiple times for each generated structure and selected the sequence with the lowest perplexity as the final prediction.  
In contrast, we performed a single ProteinMPNN prediction for each generated protein, using the output of the first run to represent the inferred sequence. 
This approach was chosen to accelerate the inference process of the model and to ensure that the final perplexity metric is not artificially inflated.  

We conduct a comprehensive comparison of DiMA with a suite of existing pre-trained models for generating proteins of varying sizes. Due to the absence of a reference sample in this experiment, we focus on evaluating protein quality and diversity.
The results are presented in the Table~\ref{tab:appendix:pretrain}.
DiMA, DPLM, ProtGPT2, and RFdiffusion models demonstrated the strongest performance in protein structural plausibility and foldability assessment.
The remaining baselines exhibit significantly lower quality in terms of perplexity and structural plausibility.

Notably, RFDiffusion, trained on structural representations of proteins, exhibits a high degree of protein structural plausibility, potentially attributed to its structural bias. 
However, RF-Diffusion exhibits a high perplexity value, suggesting a low quality of predicted amino acid sequences and potentially indicating limitations in the performance of ProteinMPNN. 
While the combined use of these models for protein sequence generation yields promising results, it does not achieve state-of-the-art performance. 

The family of DPLM models demonstrates high protein foldability quality with low perplexity scores, indicating the generation of high-quality proteins. 
However, DPLM models exhibit a significant drawback in terms of diversity (see Table \ref{tab:appendix:pretrain}).
A substantial portion of subsequences are repeated across multiple generated proteins, negatively impacting the representativeness of the generated proteins.
Notably, DPLM-3B generates 27$\%$ duplicate sequences, highlighting the challenge of balancing quality with diversity in this model.

DiMA is capable of generating high-quality and diverse protein sequences with reasonable predicted structures.
Using $100$ times fewer parameters, it achieves comparable quality to other models, like DPLM models, while surpassing them in the diversity of proteins generated. 
These findings highlight DiMA's potential as a promising approach for protein sequence generation. It balances computational efficiency with the generation of diverse and high-quality proteins.

\subsection{Protein Family Controllable Generation}
\label{appendix:family}

In this section, we evaluate the model's capability to generate novel, high-quality, and diverse proteins by leveraging protein family labels. Specifically, we focus on eight distinct protein families:
\begin{compactitem}
    \item CRISPR-associated proteins (CRISPR),
    \item Calmodulins,
    \item Glycosyl hydrolase family 12 (GH12),
    \item LexA DNA-binding domain proteins (LEXA),
    \item Phage lysozymes (Lysozyme),
    \item Transcriptional repressor NrdR proteins (NRDR),
    \item Phosphosugar-binding domain proteins (PHI), and
    \item Phosphoribosylaminoimidazole carboxylase (PurE).
\end{compactitem}

We filter the proteins, removing any duplicate entries.
The sizes of the datasets for these protein families are detailed in Table \ref{tab:appendix:family:datasets}.

\begin{table}[!htbp]
   \centering
   \caption{\sl \small The dataset sizes for the protein families used in controllable generation.}
   \vspace{1.5pt}
   \label{tab:appendix:family:datasets}

   \begin{tabular}{cccccccc}
   \toprule
     \bf{CRISPR} 
   & \bf{Calmodulins}
   & \bf{GH12}
   & \bf{LexA}
   & \bf{Lysozyme}
   & \bf{NrdR} 
   & \bf{PHI} 
   & \bf{PurE}    
\\
 
   \midrule

 2818          & 551          & 2061          & 4516          & 11395          & 4515          & 23359 & 30817          \\
  
   \bottomrule
   \end{tabular}
\end{table}

We investigate two approaches for conditional protein generation: (1) classifier guidance, which avoids additional diffusion model training, and (2) conditional fine-tuning of the diffusion model across all protein families simultaneously.

\textbf{Classifier Guidance.} 
In the classifier guidance approach, we train a classifier that takes as input a noisy, normalized encoding of ESM2-650M along with a time label and predicts logits for $9$ classes. Label $0$ represents the absence of a class, where a protein is randomly sampled from the AFDB in $50\%$ of cases, ensuring it does not belong to any of the specified protein families. The remaining labels correspond to specific protein families.
The classifier architecture consists of three transformer blocks with latent attention~\citep{lee2024nv}, followed by a linear layer. Training is performed with a batch size of $1024$ and a learning rate of $2e-4$ over $300$ iterations.

\textbf{Conditional Fine-tuning.}
In the conditional fine-tuning approach, we also utilize $9$ labels and randomly sample a protein from the AFDB in $50\%$ of cases for the absence of a specific family. To condition the diffusion model on family labels, we introduce an embedding layer. The label embeddings, after passing through a linear layer, are incorporated into the input of each transformer block within the diffusion model.
The fine-tuning process is conducted with a batch size of 1024 over 2500 iterations, using a learning rate of $5e-5$. 

In both approaches, we utilize DiMA 35M, trained on ESM2-650M encodings, as the backbone for protein representations on AFDB dataset.

\textbf{Baselines.}
We compare DiMA against two groups of baselines that enables controllable generation based on family tag:
\begin{itemize}
    \item Autoregressive models: ProLLAMA~\citep{lv2024prollama} (7B), ProGen2~\citep{ProGen_Nature} (151 M), NanoGPT~\citep{nanoGPT} (35M).
    \item Discrete diffusion model: EvoDiff~\citep{EvoDiff} (640M)
\end{itemize}
For training the baseline models, we primarily adhere to the authors' recommendations, training each model until convergence while monitoring the loss on a validation set.

For training ProLLAMA, we initialize the model with pre-trained weights available at \url{https://huggingface.co/GreatCaptainNemo/ProLLaMA}. Fine-tuning is performed using LoRA~\citep{hu2021lora} on an annotated dataset comprising eight protein families, following the format recommended by the original authors.
During our experiments, we observed that the generation parameters suggested by the authors were suboptimal for producing family-specific protein data. Most generated proteins did not correspond to the specified family. To address this issue, we adjusted the parameters and identified a set of values that yielded the highest generation fidelity. The optimized parameters are: $top_k=40$, $top_p=0.9$, repetition penalty$=1.2$, temperature$=0.5$. 
The most significant modification was in the temperature parameter, where we replaced the originally suggested value of $0.2$ with $0.5$, resulting in improved family-specific protein generation.

For fine-tuning ProGen2, we use the code provided in the repository \url{https://github.com/hugohrban/ProGen2-finetuning} and initialize the model with pre-trained weights from \url{https://github.com/salesforce/progen/tree/main/progen2}. To condition the model on the family label, we prepend the family tag at the beginning of each sequence. For sampling, we use the following parameters: temperature $= 0.5$ and $top_k = 40$.

For NanoGPT fine-tuning, we apply the same conditioning and sampling schemes as used with ProGen2. The model is initialized with weights pre-trained on the AFDB dataset.

For EvoDiff, we initialize the model using the pre-trained weights available at \url{https://github.com/microsoft/evodiff?tab=readme-ov-file}, specifically the tag \texttt{d3pm\_uniform\_640M}. The family tag is added to the beginning of each sequence for conditioning, and the model is then fine-tuned. During sampling, we generate sequences starting with an empty sequence that includes the family label at the beginning.

\textbf{Evaluation Metrics.}
We use the following metrics to evaluate the generated protein sequences.
First, we use InterProScan~\citep{blum2025interpro} to determine whether the generated proteins belong to the specified family. The proportion of correctly classified proteins among all generated sequences is referred to as Fidelity.
Next, we analyze only the generated proteins that are confirmed to belong to the target family. For these proteins, we compute the following metrics: quality (pLDDT), novelty (novelty), and diversity (CD$_{0.5}$). 
The detailed calculation of the metrics is presented in Appendix~\ref{appendix:metrics}
The aggregated values, calculated as the arithmetic mean for each family and each model, are presented in Table \ref{tab:appendix:family:global}.

\begin{table}[t]
   \centering
   \caption{\sl The performance of DiMA protein family generation against baseline methods. DiMA$_{cond}$ represents conditional fine-tuning, while DiMA$_{cg}$ corresponds to classifier guidance. The best-performing scores are highlighted in \textbf{bold}, and the second-best scores are indicated with \underline{underlined} formatting.}
   \vspace{1.5pt}
   \label{tab:appendix:family:global}

   \begin{tabular}{lcccccccc}
   \toprule
   \bf{Model}
   & \bf{CRISPR} 
   & \bf{Calmodulins}
   & \bf{GH12}
   & \bf{LexA}
   & \bf{Lysozyme}
   & \bf{NrdR} 
   & \bf{PHI} 
   & \bf{PurE}    
\\
 
   \midrule
ProLLAMA      & 0.591             & 0.531             & 0.611             & 0.583             & 0.706             & \underline{0.543} & 0.642             & 0.576 \\
ProGen2       & 0.622             & 0.088             & 0.545             & 0.531             & 0.578             & 0.520             & 0.561             & 0.514 \\
NanoGPT       & 0.520             & 0.501             & \underline{0.640} & \underline{0.612} & 0.679             & 0.528             & \underline{0.652} & 0.542 \\
Evodiff       & 0.591             & \textbf{0.648}    & 0.535             & 0.547             & 0.591             & 0.515             & 0.649             & 0.572 \\
\midrule   
DiMA$_{cond}$ & \textbf{0.765}    & 0.614             & \textbf{0.674}    & \textbf{0.679}    & \textbf{0.736}    & \textbf{0.552}    & \textbf{0.658}    & \textbf{0.590} \\
DiMA$_{cg}$   & \underline{0.689} & \underline{0.629} & 0.604             & 0.584             & \underline{0.720} & 0.536             & 0.634             & \underline{0.578} \\ 
   \bottomrule
   \end{tabular}
\end{table}

\begin{figure}[H]    
\centering    
\includegraphics[width=0.33\textwidth]{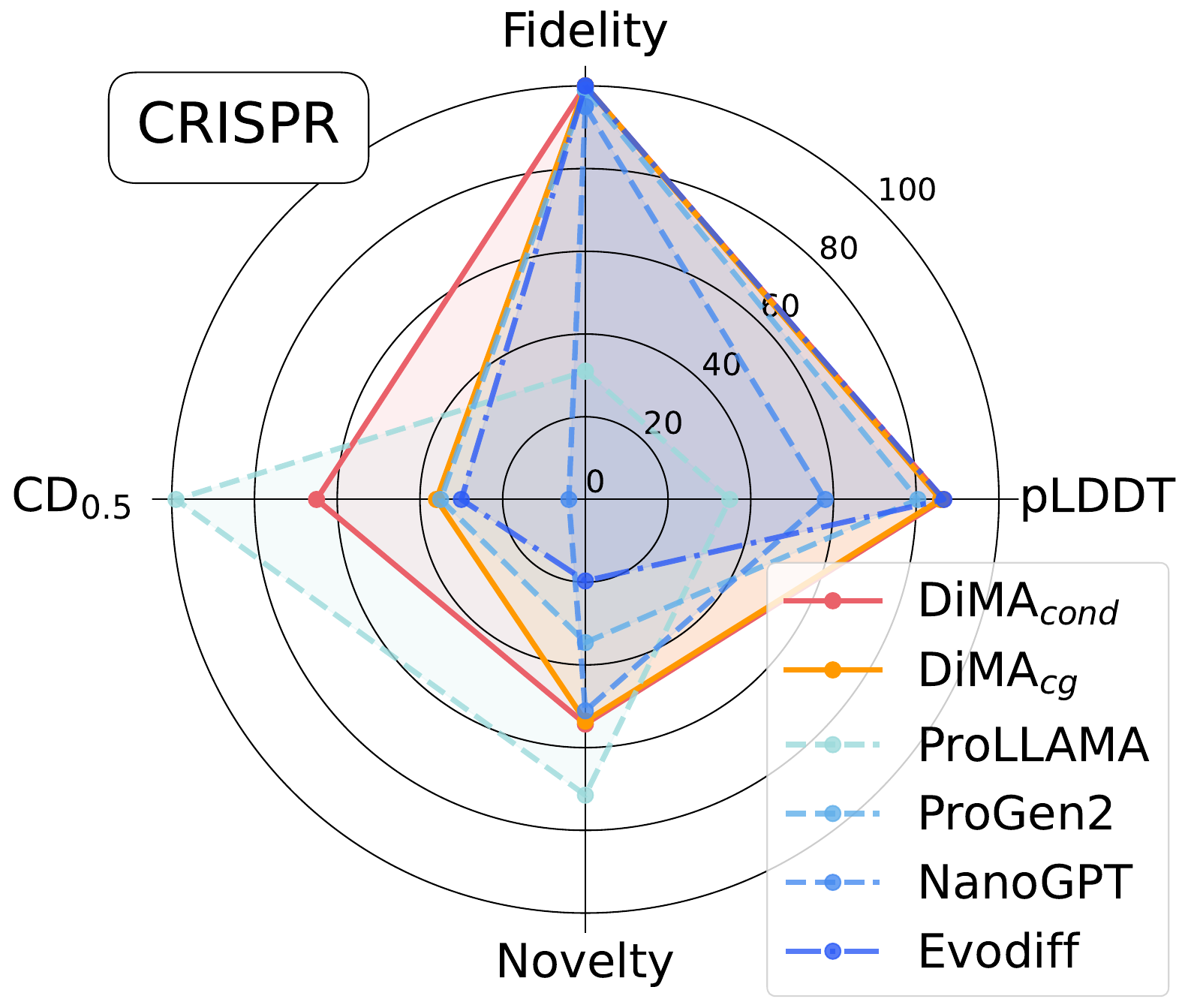} \hfill
\includegraphics[width=0.33\textwidth]{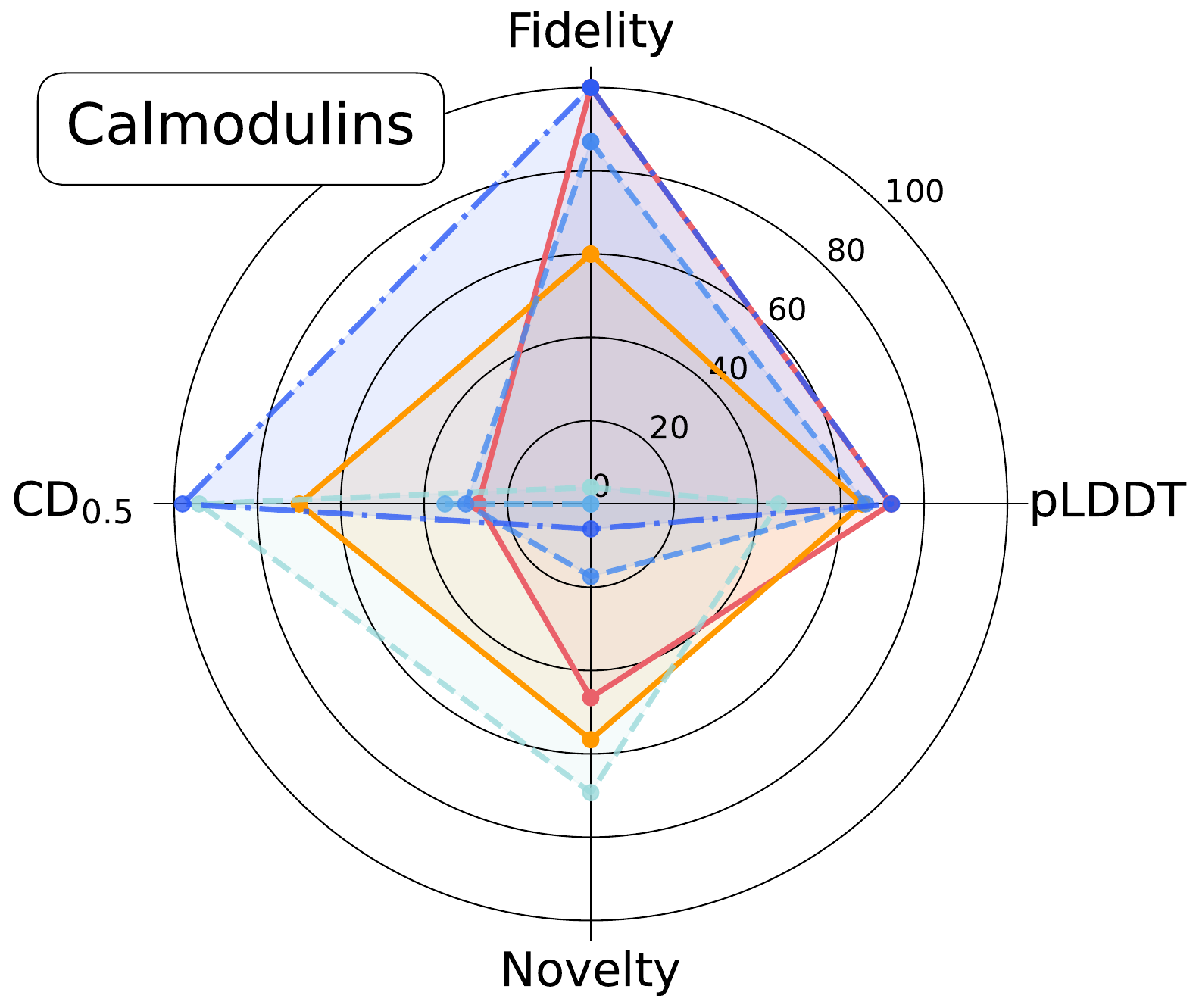} \hfill
\includegraphics[width=0.33\textwidth]{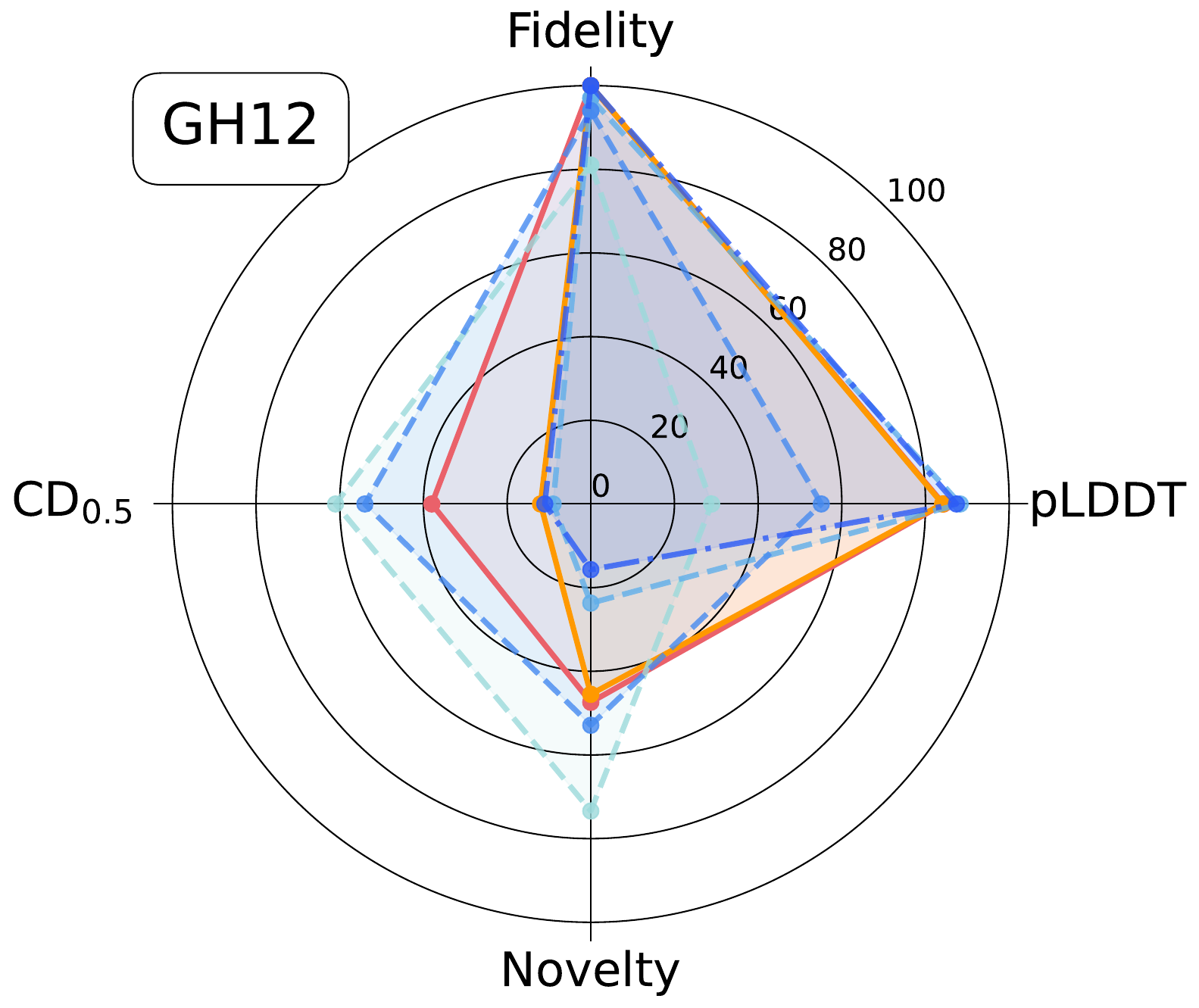} 
\\
\vspace{0.5cm}
\includegraphics[width=0.33\textwidth]{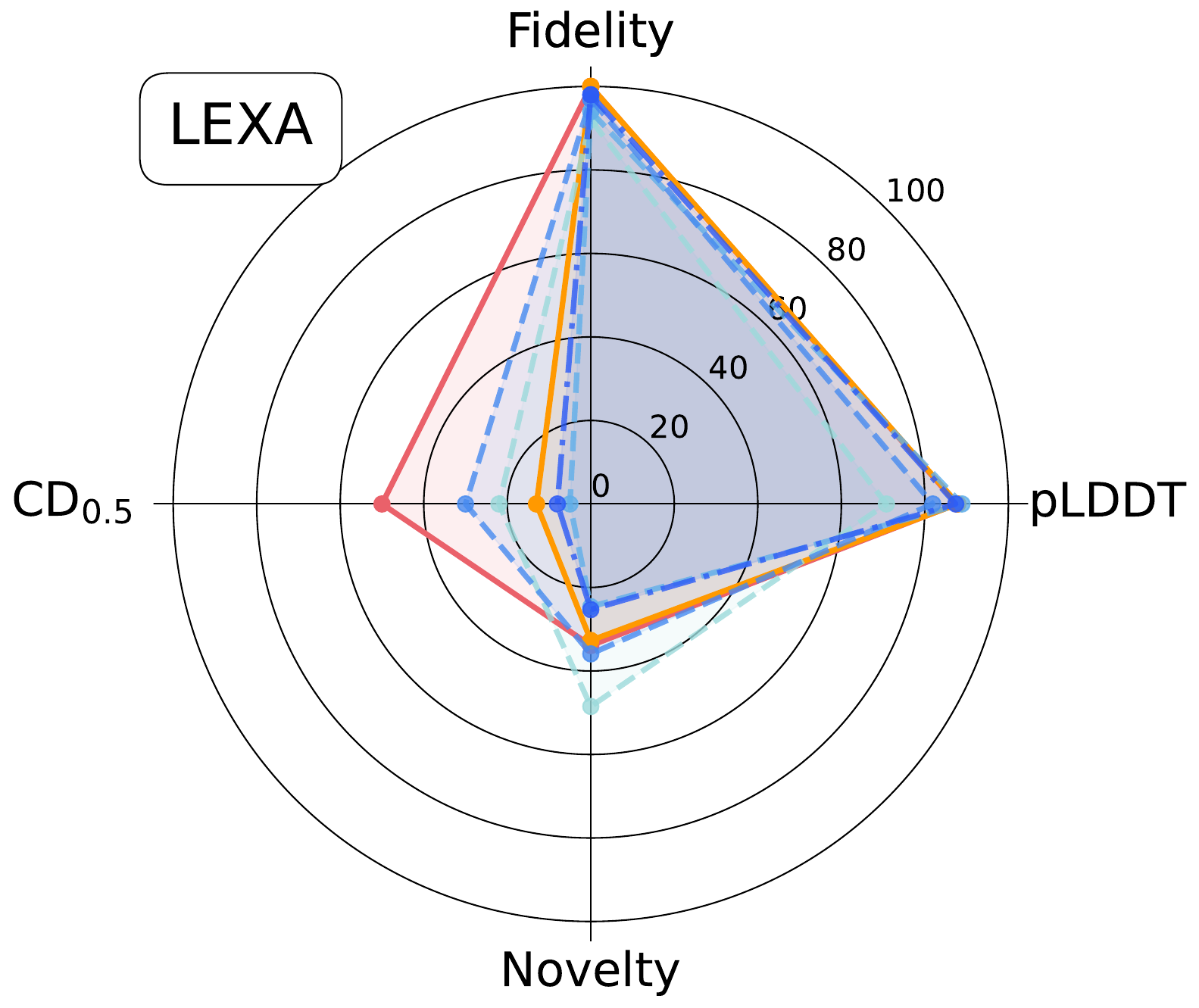} \hfill
\includegraphics[width=0.33\textwidth]{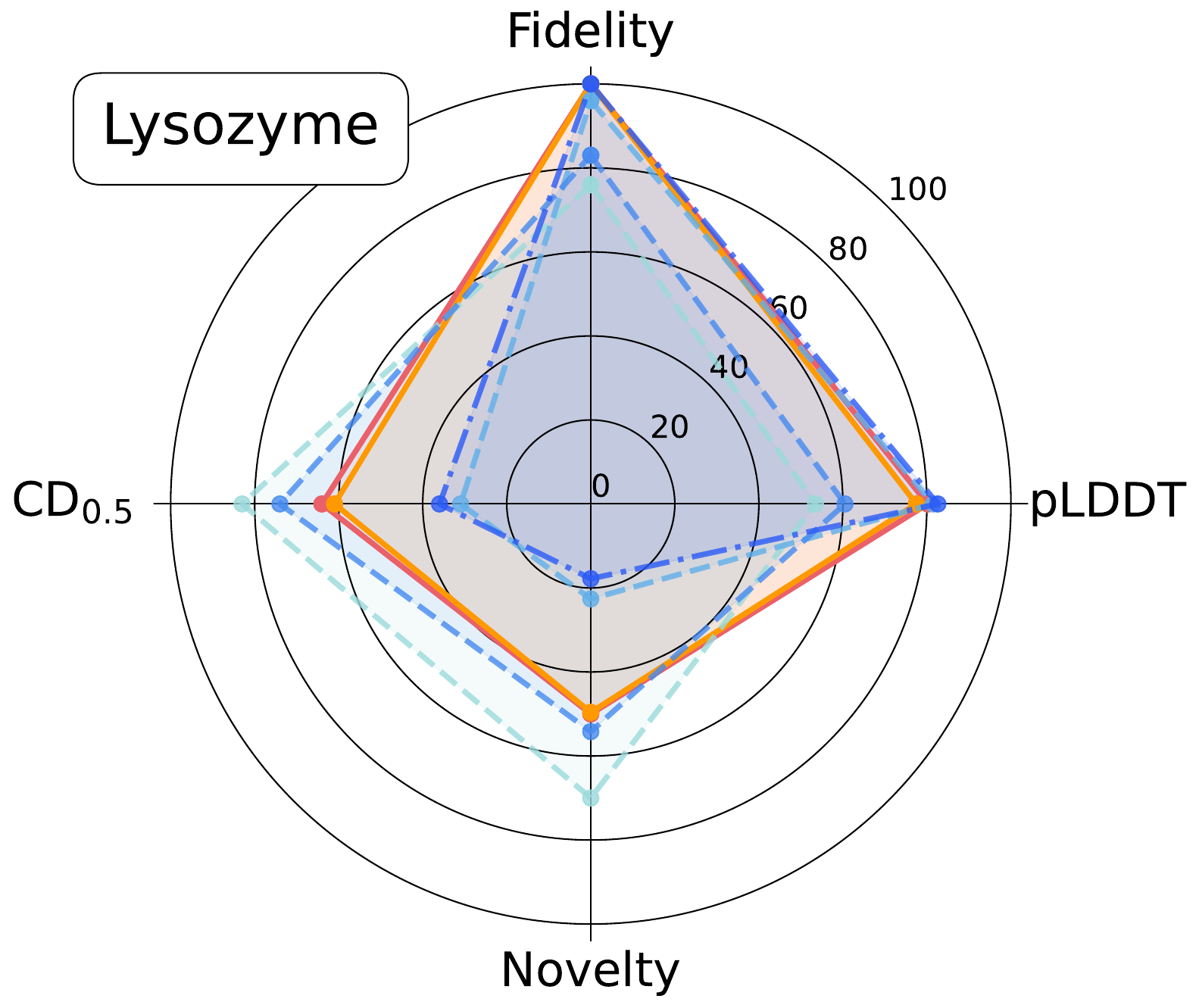} \hfill
\includegraphics[width=0.33\textwidth]{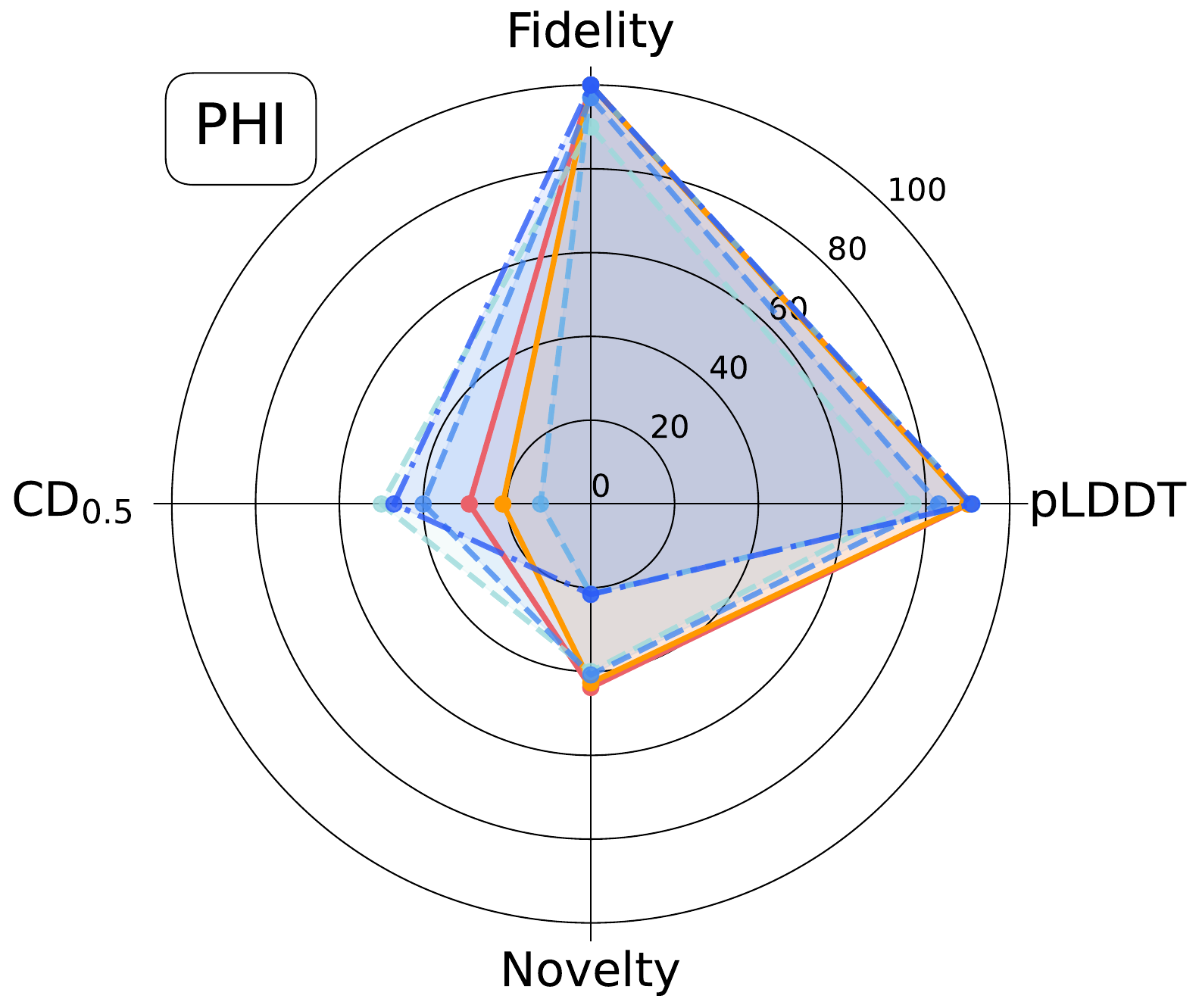}
\\
\vspace{0.5cm}


\caption{The comparison of the characteristics of generated proteins in the protein family controllable generation task. Solid lines represent DiMA models, including conditional fine-tuning (DiMA$_{cond}$) and classifier guidance (DiMA$_{cg}$). Dashed lines correspond to autoregressive baselines, while the dash-dot line indicates the performance of the discrete diffusion approach.}
\label{fig:families}
\end{figure}

\textbf{Results.}
The results presented in Figure~\ref{fig:families} highlight the remarkable ability of DiMA to generate novel, high-quality, and diverse proteins guided by family class labels. 
Unlike autoregressive baselines, which frequently deviate from the specified class labels, DiMA demonstrates a strong adherence to the intended protein family. 
A significant drawback of the autoregressive models is that a substantial proportion of their generated proteins fail to belong to the desired class. 
Even when they do, the quality of these proteins is notably lower, with an average pLDDT score hovering around 60 across many families. 
Despite this limitation, autoregressive models excel in producing a diverse range of proteins, showcasing their strength in variability.

The discrete diffusion model, on the other hand, achieves comparable performance to DiMA in terms of fidelity and protein quality. However, it struggles to match DiMA's ability to generate truly novel proteins while maintaining diversity. This limitation underscores the unique advantage of DiMA in balancing quality, novelty, and diversity in protein generation.

Interestingly, the classifier-guidance approach (DiMA$_{cg}$) emerges as a compelling alternative. While it does not outperform the fine-tuned model (DiMA$_{cond}$) it still delivers robust results, as evidenced by Table~\ref{tab:appendix:family:global}.
Notably, DiMA$_{cg}$ has an undeniable advantage over all other approaches, it achieves competitive performance without requiring additional fine-tuning of the generative model.

\subsection{Joint Sequence-Structure Generation with DiMA}
\label{app:structure}

In this section, we evaluate the DiMA model trained on the latent representations of the CHEAP encoder~\citep{lu2024tokenized}. The CHEAP framework has demonstrated that vector representations derived from input sequences can be successfully decoded into both the amino acid sequence and the corresponding protein structure. This capability enables the construction of a latent diffusion model that jointly generates structurally consistent amino acid sequences and backbone conformations.

To assess the self-consistency of the generated proteins, we follow the methodology of \citet{yim2023se} and compute two key metrics:
\begin{itemize}
\item Self-consistent Root Mean Square Deviation (scRMSD): Measures the deviation between the predicted backbone structure and the backbone reconstructed from the predicted sequence.
\item Self-consistent TM-score (scTM): Quantifies the structural similarity between the predicted and reconstructed backbones.
\end{itemize}
A generated protein is considered designable if it satisfies the criterion scRMSD $< 2 \mathring{A}$ and scTM $> 0.5$. 
Following established literature~\citep{yim2023se, lin2024out, campbell2024generative, bose2023se}, we also report the proportion of designed proteins that meet this self-consistency threshold. 

\begin{figure}[!htb]    
\centering    
\includegraphics[width=0.33\textwidth]{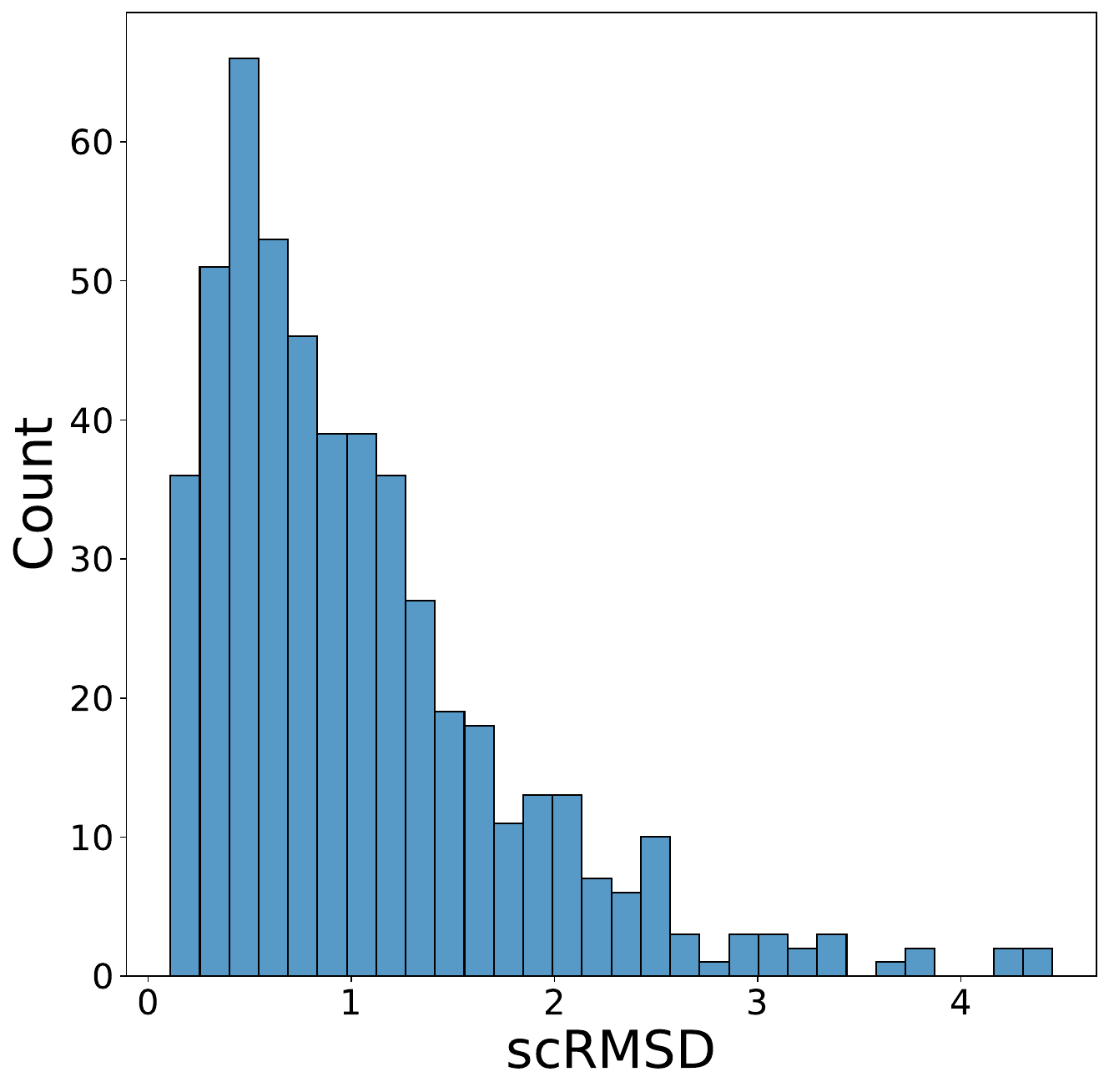} 
\includegraphics[width=0.33\textwidth]{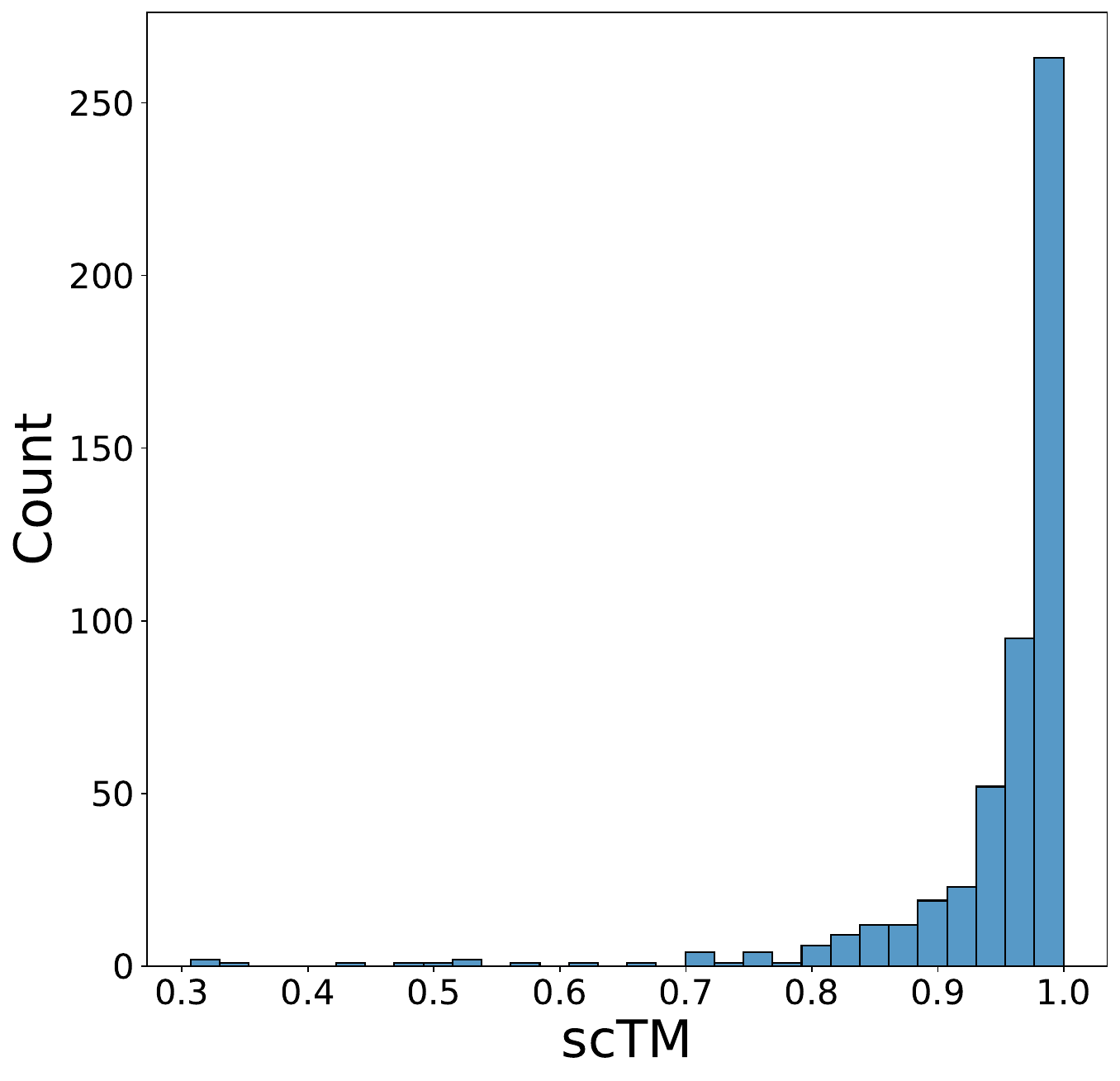}

\caption{Distributions of the self-consistency metrics between generated structure and structure reconstructed from the generated sequence.}
\label{fig:structure_generation}
\end{figure}

We present the distributions of self-consistency metrics in Fig.~\ref{fig:structure_generation}, demonstrating the high designability of the generated sequences and structures. The results indicate that the predicted amino acid sequences and backbone conformations are highly consistent, with $89\%$ of proteins achieving scRMSD $<$ $2 \mathring{A}$ and $99\%$ exhibiting scTM $>$ 0.5.
To further assess the designability of DiMA, we compare its performance against structure-based generative models \citep{RFDiffusion, yim2023se, lin2024out, bose2023se, ingraham2023illuminating} in Table~\ref{tab:structure_generation}. DiMA's performance is evaluated using two complementary approaches:
\begin{itemize}
\item Co-design evaluation: The mean scRMSD and the fraction of designable structures are computed by comparing the generated protein backbone with the structure reconstructed from its predicted sequence. This corresponds to the DiMA [co-design] approach.
\item Structure-only evaluation: The generated structures are assessed independently using ProteinMPNN, which generates eight candidate sequences for each structure. The sequence yielding the best RMSD is selected, following the standard evaluation protocol~\citep{yim2023se}. This corresponds to the DiMA [structure-only] approach.
\end{itemize}

\begin{table}[ht!]
   \centering
   \caption{\sl The performance of DiMA in structure generation compared to structure-based protein models. DiMA's quality is assessed through two approaches. In the co-design evaluation, the mean scRMSD and the fraction of designable structures are measured by comparing the RMSD between the generated structure and the structure reconstructed from the generated sequence. This corresponds to the DiMA [co-design] approach. In the DiMA [structure-only] case we perform standard structure evaluation approach, the generated structure is assessed independently by using ProteinMPNN to produce eight plausible sequences for each design, selecting the sequence with the best RMSD.}
   \vspace{1.5pt}
   \label{tab:structure_generation}

   \begin{tabular}{lcc}
   \toprule
   \bf{Model}
   & \bf{Fraction ($\uparrow$) } 
   & \bf{scRMSD ($\downarrow$) }
\\
 
   \midrule
RFDiffusion          & 0.969             & 0.650   \\
Genie 2              & 0.960             & ---            \\
FoldFlow-OT          & 0.820             & 1.806   \\
Chroma               & 0.700             & --- \\
Genie                & 0.581             & 2.968            \\
FrameDiff            & 0.555             & 2.929  \\

\midrule   
DiMA$_{\textbf{CHEAP\_shorten\_1\_dim\_1024}}$ [structure only]                & 0.923             & 1.091   \\
DiMA$_{\textbf{CHEAP\_shorten\_1\_dim\_1024}}$ [co design]                 & 0.888             & 1.043  \\ 
\midrule   
DiMA$_{\textbf{CHEAP\_shorten\_2\_dim\_1024}}$ [structure only]                & 0.845             & 1.211   \\
DiMA$_{\textbf{CHEAP\_shorten\_2\_dim\_1024}}$ [co design]                 & 0.782             & 1.496  \\ 
   \bottomrule
   \end{tabular}
\end{table}

Tables \ref{tab:structure_generation}, \ref{tab:encoders:total}  demonstrate that DiMA generates high-quality, diverse, novel, and designable protein structures despite being trained exclusively on sequence data. 
Notably, DiMA outperforms Chroma, Genie, and FrameDiff by a substantial margin, while achieving results slightly worse to state-of-the-art models like RFDiffusion and Genie2.
Although DiMA slightly lags behind the best 3D protein models, its strong performance—despite lacking direct training on structural data—highlights its ability to leverage sequence-based representations for structure generation. This reinforces the power of sequence-driven generative models, which benefit from the extensive coverage of the protein sequence landscape, even in the absence of explicit structural supervision.

\subsection{Sequence Infilling}\label{appendix:infilling}

\begin{table}[!bt]
   \centering
   \caption{\sl \small Performance comparison of DiMA and DPLM on the protein infilling task, measured by three metrics: success rate, average quality (Region pLDDT), and average novelty (Region identity). }
   \label{tab:condition:inpainting}

  \begin{tabular}{lcccc}
   \toprule
   & \bf Model 
   & \bf{Success rate, \%} ($\uparrow$) 
   & \bf{Region pLDDT} ($\uparrow$) 
   & \bf{Region Novelty} ($\uparrow$) 
   \\

\midrule                                             
    & DiMA           & 42.2     & 66.9     & 80 \\
    & DPLM	       & 40.0       & 59.3      & 75 \\
    & Random         & 21.1     & 50.9         & 92 \\

   \bottomrule
   
   \end{tabular}

\end{table}

Sequence infilling requires generating a missing segment of amino acids that maintains structural and functional coherence with the surrounding sequence context. This task serves as a test of conditional generation capabilities, demanding the model to generate novel sequence segments that integrate naturally with the given context while maintaining both global structural properties and local features at the boundaries of infilled regions.

We construct our evaluation set from SwissProt sequences with strict filtering criteria to ensure a challenging benchmark. The final test set comprises 180 sequences, each having at most 50\% sequence identity to the training set and containing no ambiguous amino acids or non-standard residues. For each sequence, we mask a region of random length, uniformly sampled between 8 and 50 amino acids, at a random position. This variation in mask length tests the model's ability to handle different infilling scenarios.

We augment the base DiMA architecture (35M parameters) with a conditional adapter for infilling. The adapter consists of 3 transformer blocks with architecture matching the base model. It processes concatenated ESM2-650M encodings of unmasked regions and mask tokens, with its outputs added to all diffusion transformer blocks via residual connections. The adapter is trained on our unconditional training set using AdamW optimizer ($\beta_1=0.9$, $\beta_2=0.999$) with a batch size of 1024 and learning rate of 5e-5 with cosine decay. Random masking is applied during the training process, which continues for 10,000 steps.

\begin{wrapfigure}{R}{0.49\textwidth}
    \includegraphics[width=0.49\textwidth]{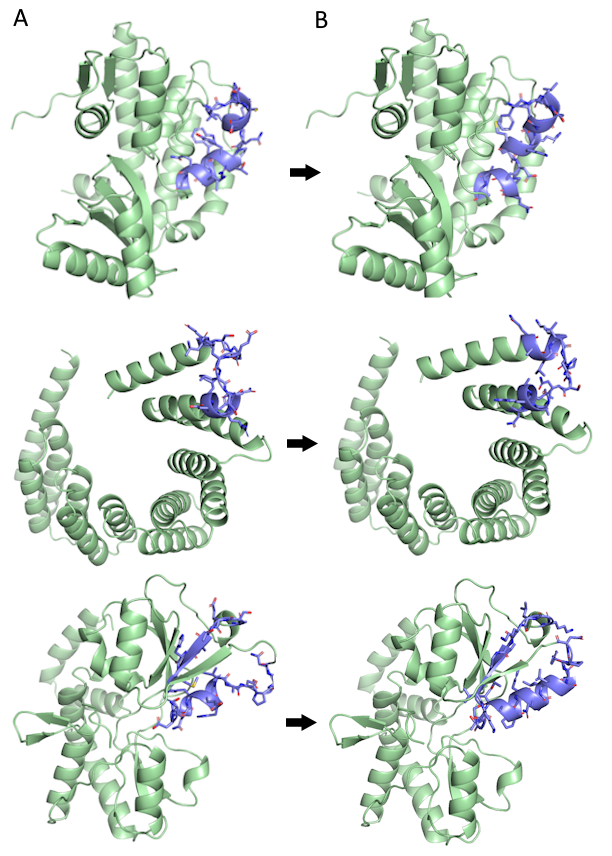}
    \caption{Examples of successfully solved infilling cases. A- reference proteins, B- DiMA generated proteins. DiMA produces a novel region (purple), being conditioned on its surroundings (green).}
    \label{fig:fig_inpainting}
\end{wrapfigure}

We define a success metric based on three criteria that must be simultaneously satisfied: (1) the complete sequence must achieve pLDDT $\geq$ 80, (2) the infilled region must maintain pLDDT $\geq$ 80, and (3) the unmasked regions must preserve their structure with RMSD $\leq$ 1\AA{} compared to reference. For each test sequence, we generate 10 infilling attempts to account for generation stochasticity, considering success if any attempt satisfies all the criteria. We use ESMFold for pLDDT calculation and structure prediction. We compare DiMA against a random baseline (randomly sampled amino acids for the masked region) and DPLM.

The quantitative evaluation demonstrates DiMA's strong performance, achieving a 42.2\% success rate compared to 40.0\% for DPLM and 21.1\% for the random baseline. DiMA generates infilled regions with higher average quality (pLDDT 66.9 vs 59.3 for DPLM) while maintaining high sequence novelty (mean 80.0). Table~\ref{tab:condition:inpainting} presents the complete results across all evaluation metrics.

Several key observations emerge from our experiments. DiMA maintains high structural quality while achieving greater sequence novelty compared to baselines. The success rate correlates positively with the length of the infilled region up to approximately 30 amino acids. Failed cases often involve regions near critical structural elements or domain boundaries. The adapter-based approach proves effective for conditioning without requiring full model retraining. 

Figure~\ref{fig:fig_inpainting} shows selected examples of sequence infilling where the generated regions maintain structural coherence with their surroundings. These results suggest that DiMA's diffusion-based architecture can be extended to handle conditional generation tasks while maintaining its core strengths in protein sequence modeling.

\subsection{Functional-motif Scaffolding}
\label{appendix:scaffolding}

Functional motif scaffolding requires generating protein sequences that incorporate predefined structural motifs while maintaining their spatial configuration. Following the RFDiffusion benchmark protocol, we evaluate performance on 24 distinct scaffolding problems. Each problem consists of a structural motif that must be preserved within a newly designed protein structure.

For this task, we use DiMA with the SaProt-650M encoder and augment it with a conditional encoder. The conditional encoder processes motif information and passes it to each transformer layer in the diffusion model through additive conditioning. We fine-tune the model on randomly masked proteins, where the masking pattern mimics the scaffolding task setup. During inference, we sample 100 designs per problem.

We employ two complementary success metrics:
\begin{itemize}
    \item \textbf{Success Rate (SR)}. A design is considered successful if it achieves pLDDT $\geq$ 70 (indicating overall structural quality) and maintains motif RMSD $\leq$ 1\AA{} (indicating preservation of the functional motif). The SR for each problem is the fraction of successful designs among 100 samples.
    \item \textbf{Unique Success Rate (USR)}. To account for potential mode collapse, we measure CD\textsubscript{0.5} diversity by cluster successful designs using the 50\% sequence identity threshold and count only cluster centroids. This metric reflects the diversity of viable solutions.
\end{itemize}

As shown in Table~\ref{tab:condition:scaffolding_sr}, DiMA successfully solves 19 out of 24 benchmark problems, positioning it among the top-performing methods. Detailed analysis reveals several key findings:

\begin{itemize}
    \item \textbf{Performance by Problem Type}. DiMA shows particularly strong performance on 6E6R variants (SR: 0.66-0.79) but struggles with certain challenging cases like 4JHW and 5YUI where most methods fail.
    
    \item \textbf{Solution Diversity}. DiMA achieves a mean unique SR of 0.1, matching ESM-3 and outperforming RFDiffusion (0.06), indicating greater diversity in successful designs.
    
    \item \textbf{Model Size Efficiency}. DiMA (35M parameters) achieves competitive performance with significantly larger models like ESM-3 and RFDiffusion, as visualized in Figure~\ref{fig:scaffolding_main}.
\end{itemize}

\begin{figure}{}
    \centering
    \includegraphics[width=0.7\textwidth]{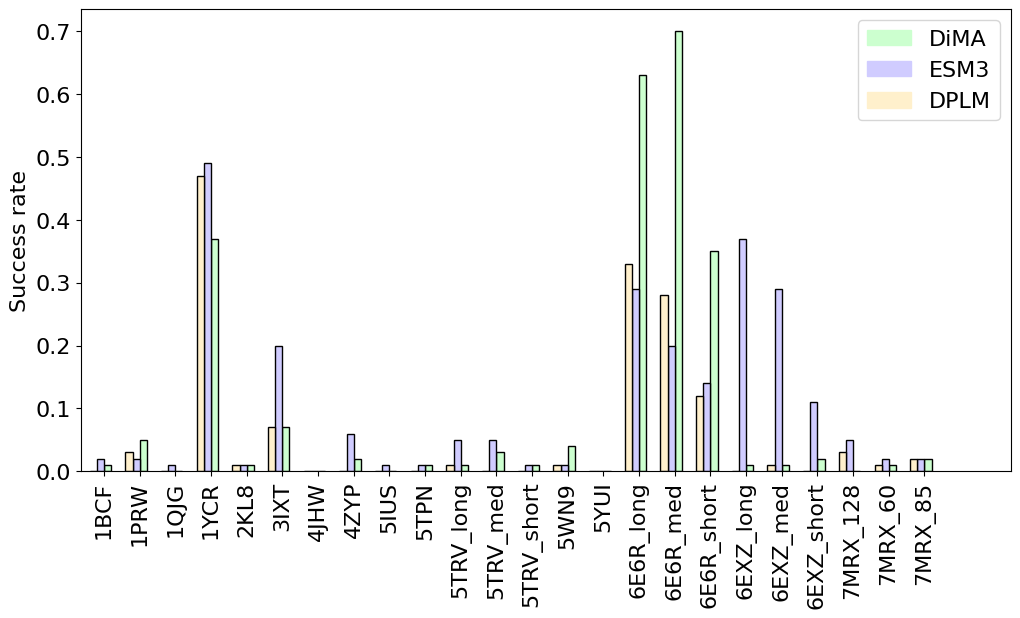}
    \caption{ \small Motif-scaffolding: models performances across 24 benchmark problems, measured by unique success rates. DPLM model ia sequence based, ESM3 and DiMA- are co-generation models variants.}
    \label{fig:scaffolding_unique}
\end{figure}

Figure~\ref{fig:scaffolding_unique} presents a detailed comparison of unique success rates across methods, highlighting DiMA's ability to generate diverse successful scaffolds. DiMA and ESM-3 show different strengths across problem types, with DiMA performing better on 6E6R variants while ESM-3B excels at 6EXZ problems (Table~\ref{tab:condition:scaffolding_sr}).

\begin{table}[t]
\caption{\sl \small Detailed results on scaffolding benchmark for each task separately.}
   \label{tab:condition:scaffolding_sr}
\begin{tabular}{>{\bfseries}l *{9}{c}}
\toprule
 & \multicolumn{4}{c}{\bfseries sequence} & \multicolumn{2}{c}{\bfseries structure} & \multicolumn{3}{c}{\bfseries cogeneration} \\
\cmidrule(lr){2-5} \cmidrule(lr){6-7} \cmidrule(lr){8-10}
 & EvoDiff & DPLM & ESM3 & DPLM 2 & RFDiffusion & DPLM2 & ESM3 & DPLM 2 & DIMA \\
\midrule
1BCF & 0 & 0 & 0.89 & 0.01 & 1 & 0.07 & 0.23 & 0.01 & 0.01 \\
1PRW & 0.61 & 0.83 & 0.96 & 0.86 & 0.08 & 0.96 & 0.54 & 0.84 & 0.7 \\
1QJG & 0 & 0 & 0.02 & 0.03 & 0 & 0 & 0.03 & 0.02 & 0 \\
1YCR & 0.02 & 0.38 & 0.41 & 0.77 & 0.74 & 0.93 & 0.18 & 0.53 & 0.37 \\
2KL8 & 0.04 & 0.08 & 0.11 & 0.47 & 0.88 & 0.94 & 0.11 & 0.57 & 0.03 \\
3IXT & 0.06 & 0.17 & 0.18 & 0.67 & 0.25 & 0.77 & 0.02 & 0.41 & 0.44 \\
4JHW & 0 & 0 & 0 & 0 & 0 & 0 & 0 & 0 & 0 \\
4ZYP & 0 & 0 & 0.03 & 0.16 & 0.4 & 0.51 & 0.08 & 0.1 & 0.02 \\
5IUS & 0 & 0 & 0 & 0 & 0.02 & 0 & 0 & 0 & 0 \\
5TPN & 0 & 0 & 0.03 & 0 & 0.61 & 0.06 & 0.01 & 0 & 0.01 \\
5TRV\_long & 0 & 0 & 0.19 & 0 & 0.37 & 0.08 & 0.19 & 0 & 0.05 \\
5TRV\_med & 0 & 0 & 0.16 & 0.03 & 0.24 & 0.07 & 0.16 & 0.02 & 0.05 \\
5TRV\_short & 0 & 0 & 0.01 & 0.07 & 0.04 & 0.1 & 0.01 & 0.03 & 0.01 \\
5WN9 & 0 & 0 & 0.02 & 0 & 0 & 0.2 & 0 & 0 & 0.1 \\
5YUI & 0 & 0 & 0 & 0 & 0.02 & 0 & 0 & 0 & 0 \\
6E6R\_long & 0.01 & 0.65 & 0.07 & 0.91 & 0.86 & 0.92 & 0.04 & 0.78 & 0.67 \\
6E6R\_med & 0.03 & 0.94 & 0.24 & 0.93 & 0.89 & 0.88 & 0.14 & 0.77 & 0.79 \\
6E6R\_short & 0.07 & 0.87 & 0.09 & 0.86 & 0.39 & 0.78 & 0.06 & 0.64 & 0.66 \\
6EXZ\_long & 0 & 0.01 & 0.32 & 0.61 & 0.76 & 0.63 & 0.13 & 0.44 & 0.01 \\
6EXZ\_med & 0 & 0 & 0.31 & 0.66 & 0.49 & 0.63 & 0.31 & 0.55 & 0.01 \\
6EXZ\_short & 0 & 0 & 0.31 & 0.66 & 0.39 & 0.41 & 0.28 & 0.58 & 0.02 \\
7MRX\_128 & 0 & 0.02 & 0.36 & 0.23 & 0.09 & 0.32 & 0.37 & 0.2 & 0 \\
7MRX\_60 & 0 & 0.31 & 0.65 & 0.28 & 0.11 & 0.31 & 0.59 & 0.22 & 0.01 \\
7MRX\_85 & 0 & 0.34 & 0.68 & 0.26 & 0.02 & 0.41 & 0.74 & 0.24 & 0.02 \\
\midrule
\bfseries Total Success & 7 & 11 & 21 & 18 & 21 & 20 & 20 & 18 & 19 \\
\bfseries mean SR & 0.04 & 0.19 & 0.25 & 0.35 & 0.36 & 0.42 & 0.18 & 0.29 & 0.17 \\
\bfseries mean unique SR & - & 0.06 & 0.06 & - & 0.07* & - & 0.1 & - & 0.1 \\
\bottomrule
\multicolumn{10}{l}{*Based on the analysis from \cite{scaffold_frameflow}} \\
\end{tabular}
\end{table}

\subsection{Fold-Conditioned Generation}\label{appendix:fold}

Fold-conditioned protein generation aims to produce novel protein sequences that adopt specific three-dimensional structural folds. Unlike inverse folding, where the protein backbone is fixed, or fold-class conditioning via discrete labels, fold-conditioned generation allows exploration of the structural neighborhood of a target fold through generation of diverse but structurally similar proteins. This capability is particularly valuable for protein engineering applications that require structural similarity while maintaining sequence flexibility.

We utilize DiMA (35M parameters) with the CHEAP encoder (shorten=1, dim=1024 configuration), which provides access to ESMFold's latent space representations. This choice enables direct encoding of structural information into the continuous latent space used by our diffusion model. The model architecture remains unchanged from the base configuration, with structural information incorporated through the same conditioning mechanisms developed for sequence tasks.

The model is finetuned on the CATH non-redundant S40 dataset, comprising approximately 27,000 non-redundant protein structures. We employ the following training parameters:
training duration of 20,000 steps with batch size of 256 and learning rate of 1e-4.

We conduct evaluations on two benchmarks. The first consists of 100 diverse protein structures from a CATH S40 hold-out set. The second uses specific protein folds previously studied in the literature \cite{RFDiffusion}: a de novo designed TIM barrel fold (PDB: 6WVS) and NTF2 fold (PDB: 1GY6). For each target structure, we generate 10 protein designs and assess their structural similarity using TM-score and RMSD metrics.

Success criteria are defined as follows:
\begin{compactitem}
   \item \textbf{Per-protein Success}. At least one generated design achieves TM-score $\geq$ 0.5
   \item \textbf{Overall Success Rate}. Percentage of target proteins meeting per-protein success criterion
   \item \textbf{Mean Maximum TM-score}. Average of best TM-scores across 10 attempts for each target
\end{compactitem}

We compare DiMA against RFDiffusion, a structure generation model that conditions on secondary structure and block-adjacency information. The comparison uses identical evaluation protocols and success criteria for both models.

\begin{table}[h]
\centering
\caption{Performance comparison on fold-conditioned generation tasks.}
\label{tab:fold_cond}
\begin{tabular}{lcc}
\toprule
\textbf{Model / Benchmark} & \textbf{Mean max TM-score} & \textbf{Success rate (\%)} \\
\midrule
DiMA (35M) & 0.93 & 100 \\
RFDiffusion (90M) & 0.48 & 41 \\
\bottomrule
\end{tabular}
\end{table}

On the CATH benchmark, DiMA achieves mean TM-score of 0.93 with 100\% success rate, compared to RFDiffusion's mean TM-score of 0.48 and 41\% success rate. For TIM barrel and NTF2 folds from the RFDiffusion paper DiMA maintains consistent performance with 100\% success rate (Table \ref{tab:fold_cond}). The high TM-scores achieved by DiMA indicate significant structural similarity between generated proteins and their target folds. However, the non-zero RMSD values (mean 2.6\AA) demonstrate that the model generates structurally similar but non-identical proteins, confirming true fold-conditioned generation rather than simple structure copying. Figure~\ref{fig:fold_cond_figure} presents representative examples of generated structures alongside their target folds.

The performance difference between DiMA and RFDiffusion can be attributed to their distinct approaches to structural conditioning. DiMA leverages rich structural encodings from the CHEAP encoder, whereas RFDiffusion relies on more abstract representations through secondary structure and block-adjacency information. This difference in conditioning information likely contributes to DiMA's higher success rates despite its smaller parameter count.

\section{Biological Relevance Analysis}
\label{appendix:biol_relevance}
\begin{figure*}[!htb]
  \vspace{-2.5mm}
  \includegraphics[width=\textwidth]{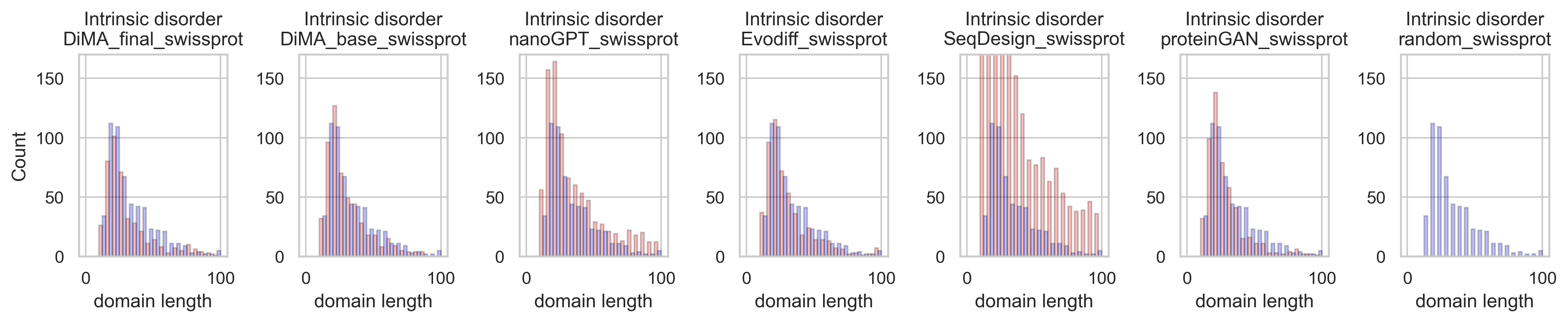}
  \includegraphics[width=\textwidth]{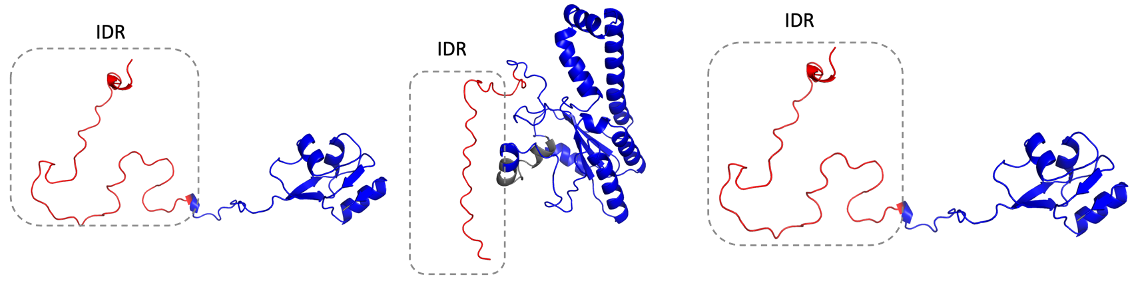}
  \caption{Prediction of intrinsic disorder regions (IDR) using the MobiDBLite tool \citep{piovesan2018mobidb}. (A) Histogram depicting the lengths of intrinsic disorder regions. The blue color represents the dataset, while the red color represents the generated sequences. No hits were found for random sequences. (B) Representative examples of proteins generated by DiMA, highlighting intrinsic disorder regions in red and folded structural domains in blue.}
  \label{fig:fig_annotation_IDR}
\end{figure*}

\begin{figure}[!htb]
    \centering
    \includegraphics[width=0.5\textwidth]{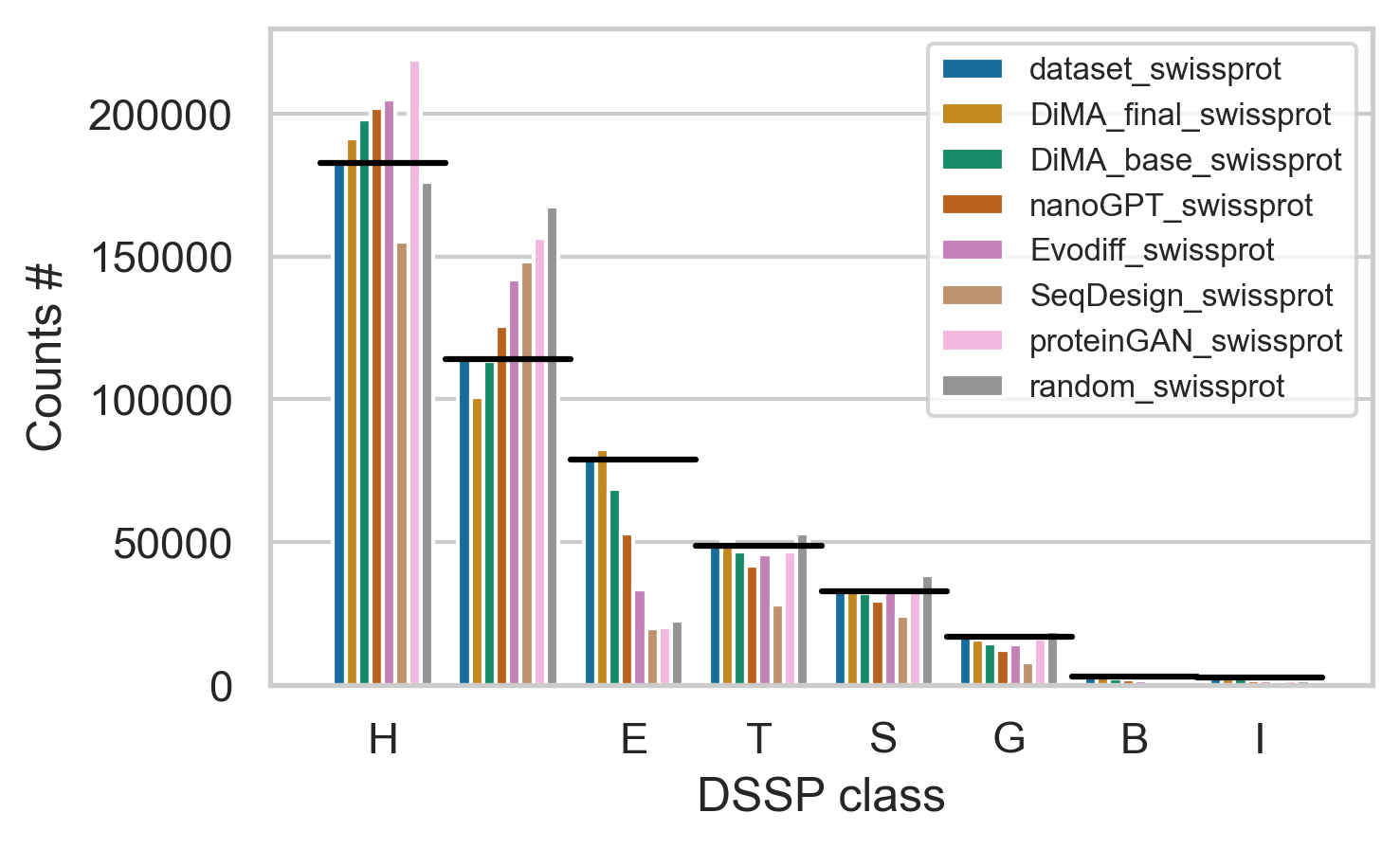}
    \caption{The number of secondary structure elements calculated per residue from ESMFold predicted structures using DSSP \citep{kabsch1983dictionary} software. H = $\alpha$-helix; B = residue in isolated $\beta$-bridge; E = extended strand, participates in $\beta$ ladder; G = 3-helix (310 helix); I = 5 helix ($\pi$-helix); T = hydrogen bonded turn; S = bend.}
    \label{fig:fig_annotation_DSSP}
\end{figure}

\textbf{Superfamily Annotation}.
For proteins annotation we utilized the established protein annotation tool InterProScan \citep{paysan2023interpro,jones2014interproscan}.
InterProScan includes a set of pre-trained models based on hidden Markov models (HMMs), which allow for assigning potential folds and functions. This analysis involves annotating the generated protein sequences using the SUPERFAMILY HMM library \citep{oates2015superfamily}, which provides sequence homology to SCOP structural domains \citep{murzin1995scop}. 
\textbf{IDR Exploration}.
Natural proteins encompass both structured regions and IDRs that lack regular structure but still play functional roles \citep{uversky2015functional} (Figure \ref{fig:fig_annotation_IDR}). To annotate these regions, we employ the MobiDB model within the InterProScan tool, which predicts IDRs in protein sequences using multiple classifiers \citep{piovesan2018mobidb}. Sequences generated by DiMA exhibit a natural-like profile of IDR length distribution (Figure \ref{fig:fig_annotation_IDR}). Generation of both folded and unfolded structural regions provides a distinct advantage for sequence diffusion models over models exclusively trained on folded protein domains. 
\textbf{Secondary Structure Exploration}.
Finally, we calculate the frequency of secondary structure elements within the folded regions using the DSSP tool \citep{kabsch1983dictionary} against protein structures predicted via ESMFold. DiMA mirrors the amount of secondary structural elements of natural proteins. DiMA generates seuqnces with number of secondary elements close to relevant number in validation dataset (Figure \ref{fig:fig_annotation_DSSP}).

\begin{figure}
    \centering
    \includegraphics[width=0.5\textwidth]{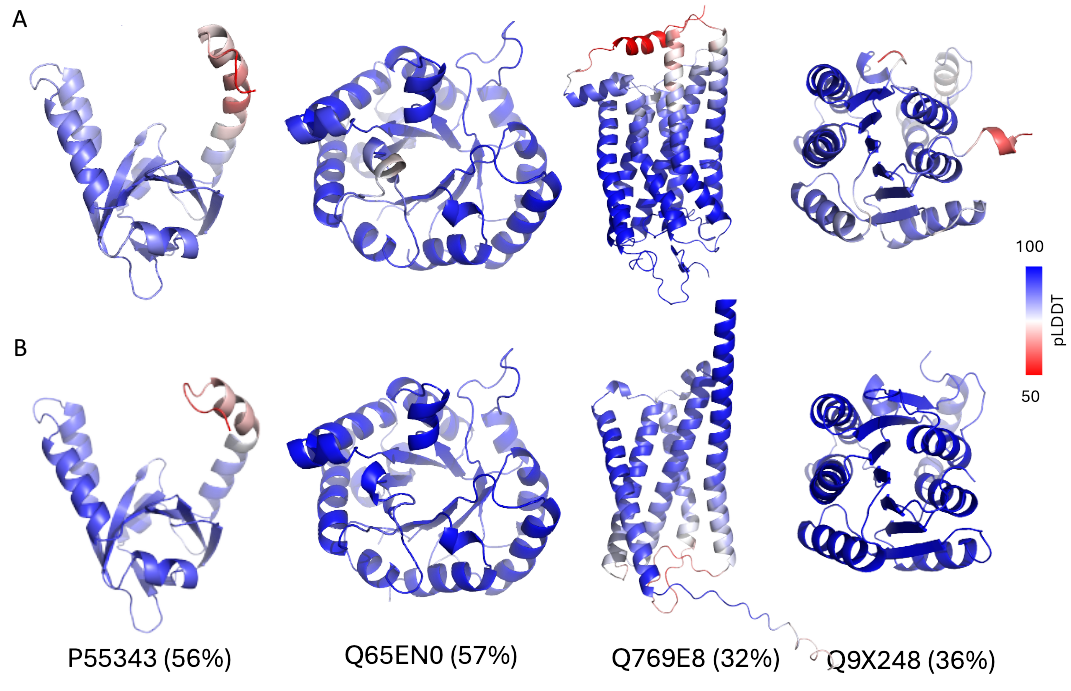}
    \caption{ \small ESMFold predicted representative examples of proteins generated by DiMA (A) and the closest hit SwissProt (B) with UniProt IDs and the homology \%, colored by pLDDT.}
    \label{fig:generation_example}
\end{figure}

\begin{figure}[!htb]
    \centering
    \includegraphics[width=1\textwidth]{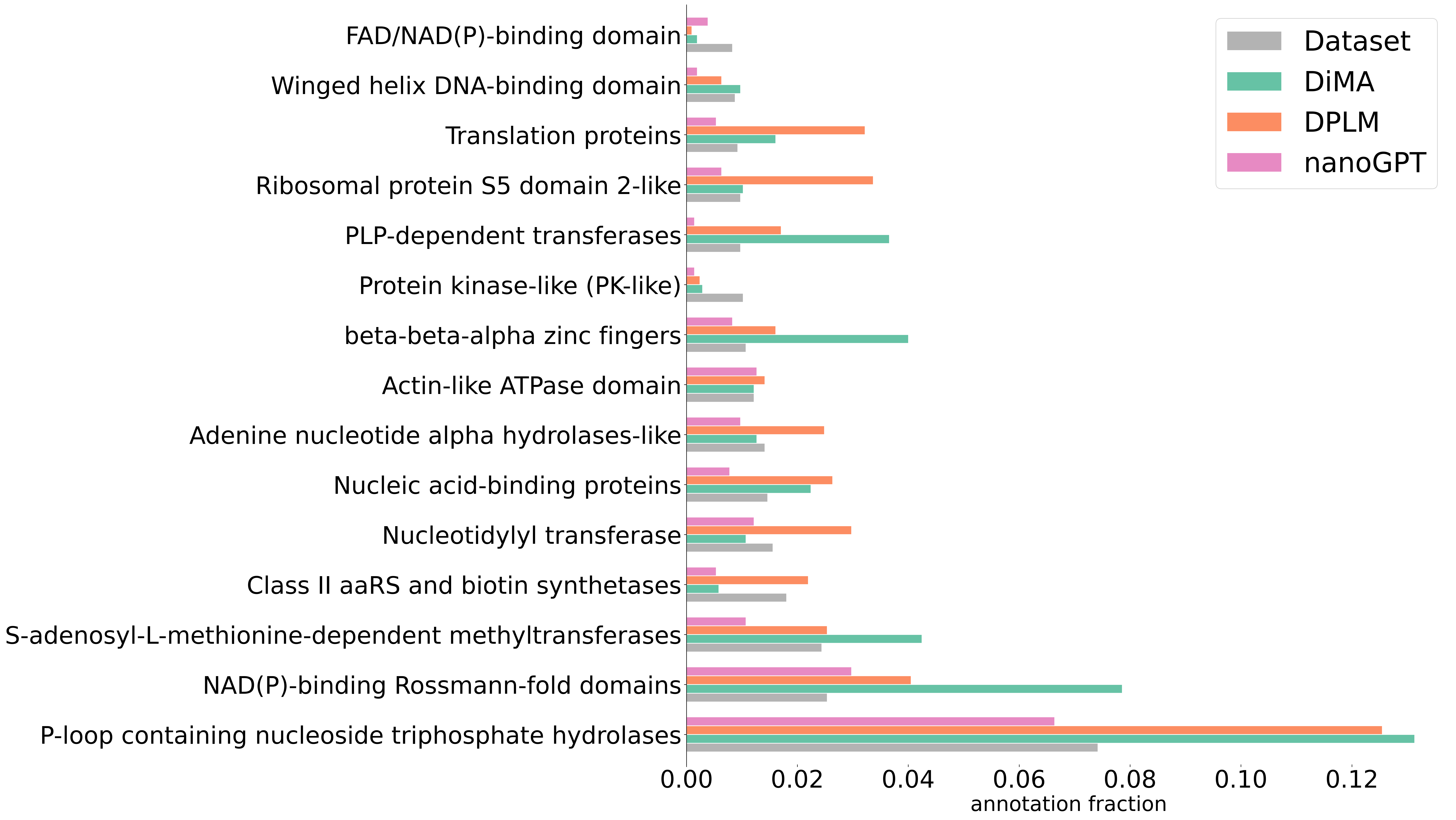}
    \caption{ Histogram depicting the occurrence of the top 15 most frequent SUPERFAMILY domains in the SwissProt dataset pool. \citep{oates2015superfamily,jones2014interproscan}. x- Fraction of each annotation per model.}
\label{fig:fig_annotation_InterPro_class}
\end{figure}

\begin{figure*}[!htb]
    \centering
    \includegraphics[width=\textwidth]{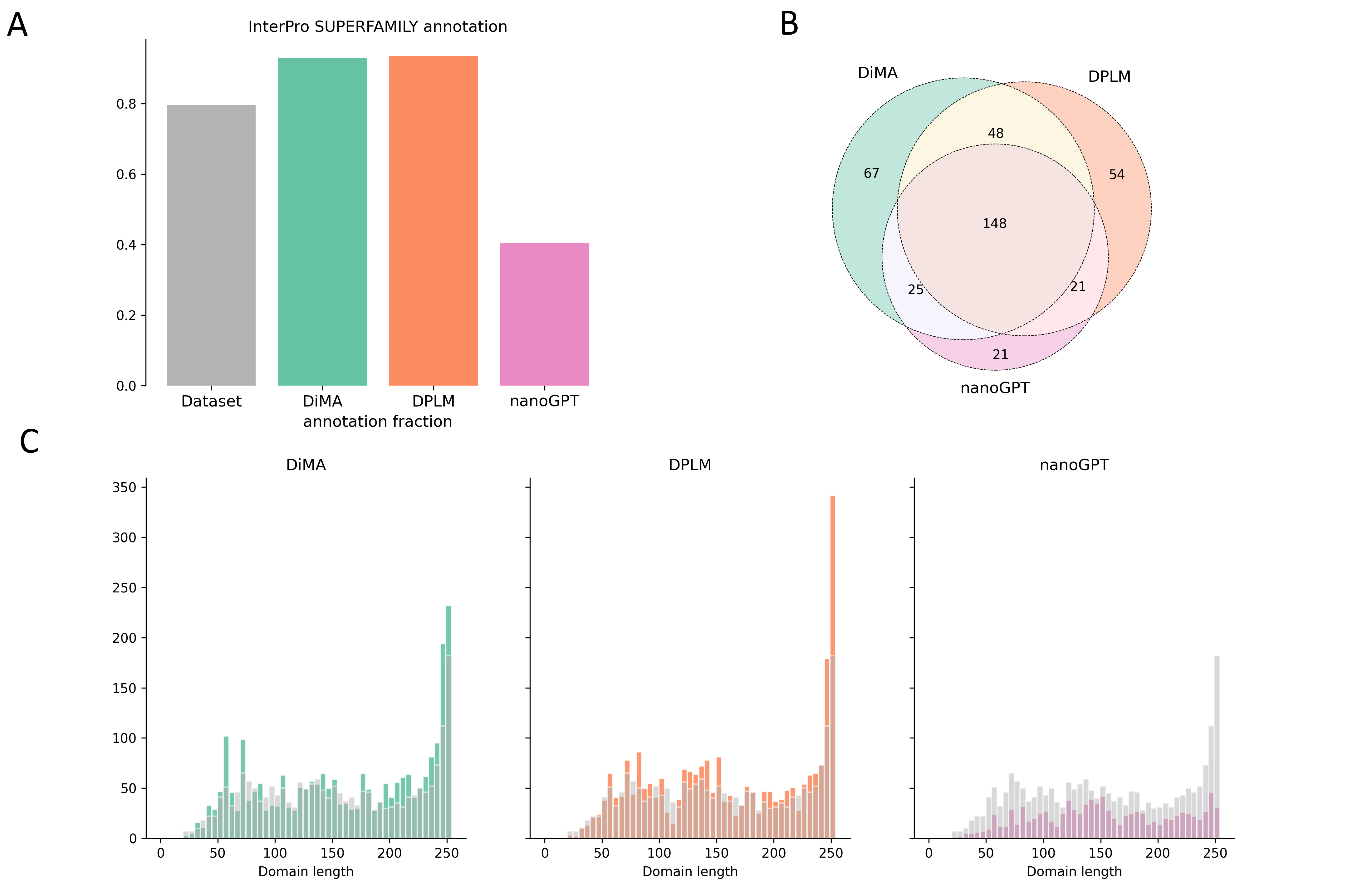}
    \caption{Analysis of protein domain characteristics in generated sequences. 
    (A)~Fraction of sequences containing InterPro SUPERFAMILY domains shows comparable annotation rates between models and natural proteins. 
    (B)~Venn diagram revealing DiMA's superior domain coverage, generating both more unique domains (67) and higher total domain count compared to DPLM and nanoGPT. 
    (C)~Domain length distributions demonstrating that DiMA accurately preserves natural domain length patterns, while DPLM shows bias toward full-sequence domains (240--250~aa), and nanoGPT exhibits non-native uniform distribution. 
    These results indicate DiMA's capacity to capture diverse and biologically relevant domain architectures.}
    \label{fig:fig_annotation_InterPro_distrib}
\end{figure*}

\begin{figure}[!htb]
    \centering
    \includegraphics[width=0.5\textwidth]{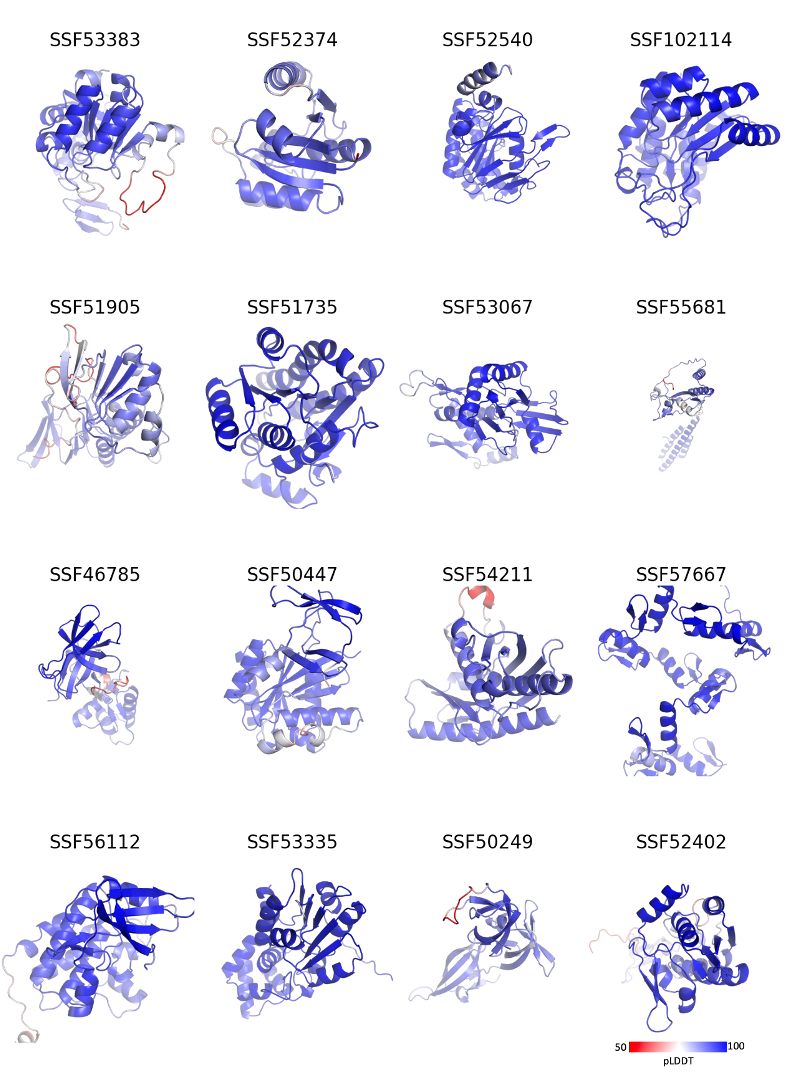}
    \caption{
    Sequence annotation into known structural domains using SUPERFAMILY tool within InterProScan \citep{oates2015superfamily,jones2014interproscan}. ESMFold-predicted structures of representative SUPERFAMILY domains generated by DiMA.}
    \label{fig:fig_annotation_InterPro}
\end{figure}

\end{document}